\pdfoutput=1
\documentclass[11pt]{article}

\usepackage[final]{acl}

\hypersetup{
           breaklinks=true,   
           colorlinks=true,   
           pdfusetitle=true,  
           citecolor=darkblue, 
           linkcolor=darkblue, 
           urlcolor=darkblue
        }

\PassOptionsToPackage{table}{xcolor}
\usepackage{xcolor}
\usepackage{colortbl}

\usepackage{times}
\usepackage[T1]{fontenc}
\usepackage[utf8]{inputenc}
\usepackage{microtype}
\usepackage{inconsolata}
\newlanguage\ttnohyph
\makeatletter
\g@addto@macro\ttfamily{\hyphenchar\font=`\-\relax\language=\ttnohyph\relax}
\makeatother
\usepackage{graphicx}
\usepackage{booktabs}
\usepackage{adjustbox}
\usepackage{placeins}
\usepackage{makecell}
\usepackage{tabularx}
\usepackage{multicol}
\usepackage[inline]{enumitem}
\usepackage{multirow}
\usepackage{xspace}
\usepackage{tikz}
\usepackage{pifont}%
\usepackage{tcolorbox}
\usepackage{listings}
\lstdefinestyle{judgeprompt}{basicstyle=\ttfamily\scriptsize, breaklines=true, breakatwhitespace=true, breakindent=0pt,
  columns=fullflexible, keepspaces=true, upquote=true, aboveskip=0pt, belowskip=0pt}
\newcommand{\ds}[1]{\textsc{#1}\xspace}
\newcommand{\dataset}{\ds{Primeape}}

\newcommand{\relevance}{%
  \adjustbox{valign=t}{\includegraphics[height=0.7\baselineskip]{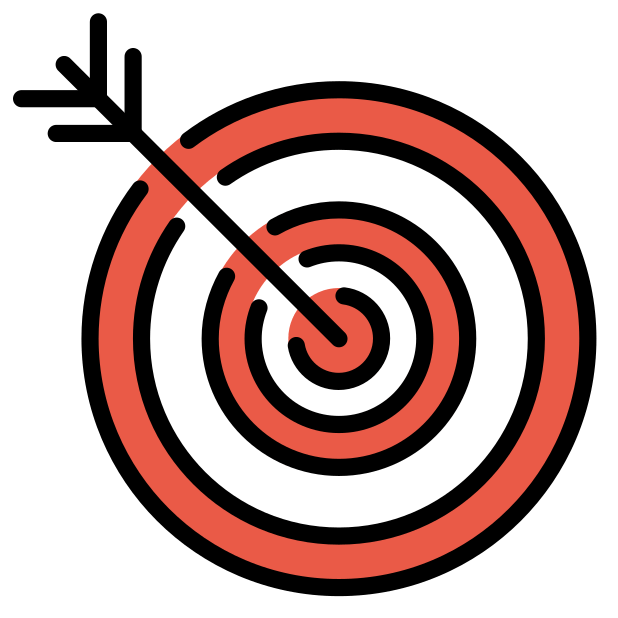}}~\textsc{Relevance}\xspace%
}

\newcommand{\naturalness}{%
  \adjustbox{valign=t}{\includegraphics[height=0.7\baselineskip]{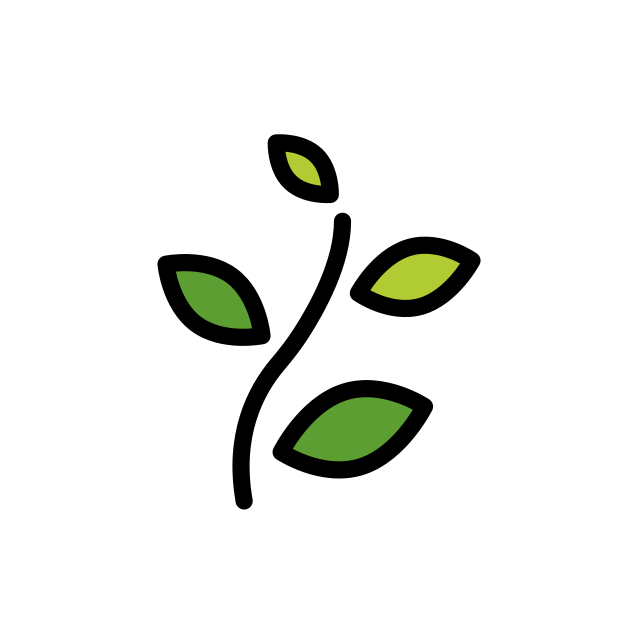}}~\textsc{Naturalness}\xspace%
}

\newcommand{\truthfulness}{%
  \adjustbox{valign=t}{\includegraphics[height=0.7\baselineskip]{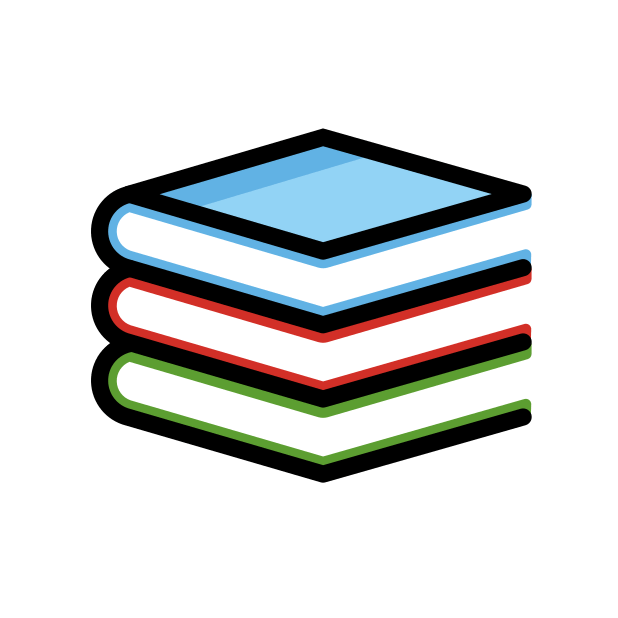}}~\textsc{Truthfulness}\xspace%
}

\newcommand{\safety}{%
  \adjustbox{valign=t}{\includegraphics[height=0.7\baselineskip]{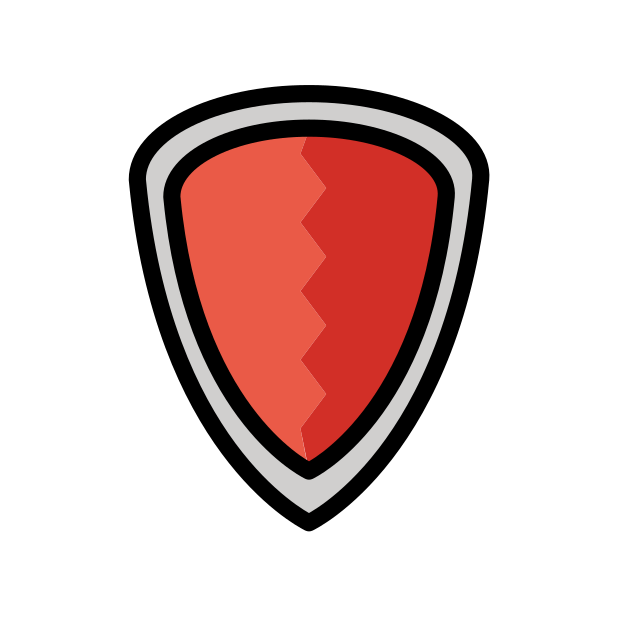}}~\textsc{Safety}\xspace%
}

\newcommand{\overall}{%
  \adjustbox{valign=t}{\includegraphics[height=0.7\baselineskip]{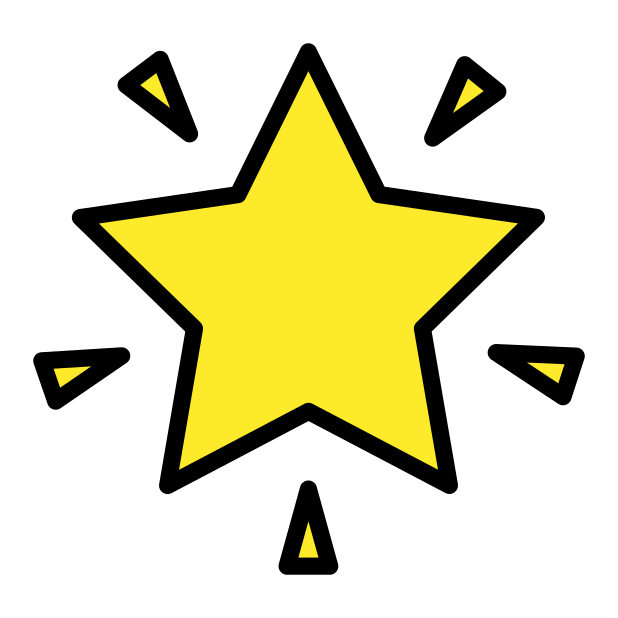}}~\textsc{Overall~quality}\xspace%
}

\makeatletter
\newcommand*{\radiobutton}{%
  \@ifstar{\@radiobutton0}{\@radiobutton1}%
}
\newcommand*{\@radiobutton}[1]{%
  \begin{tikzpicture}
    \pgfmathsetlengthmacro\radius{height("X")/3}
    \draw[radius=\radius] circle;
    \ifcase#1 \fill[radius=.6*\radius] circle;\fi
  \end{tikzpicture}%
}
\makeatother

\newcommand{\logo}[0]{\raisebox{-.25\height}{\includegraphics[width=.06\textwidth]{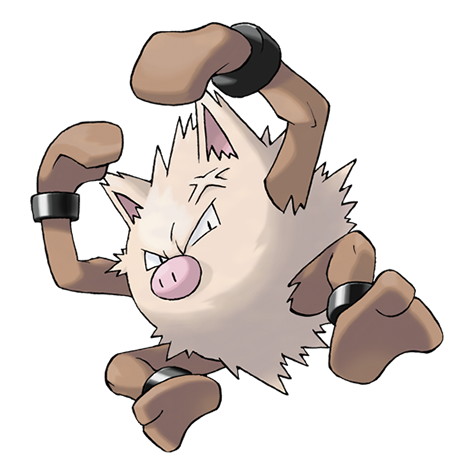}}}

\newcommand{\hf}[2]{\href{https://huggingface.co/#1}{\texttt{#2}}}

\newcommand{\cjoint}{\textsc{joint}\xspace}
\newcommand{\cjointr}{\textsc{joint+r}\xspace}
\newcommand{\ccrit}{\textsc{criterion}\xspace}
\newcommand{\ccritr}{\textsc{criterion+r}\xspace}
\newcommand{\cpos}{\textsc{position bias}\xspace}
\newcommand{\clen}{\textsc{length bias}\xspace}
\newcommand{\cself}{\textsc{self-preference}\xspace}

\definecolor{cb-green-sea}{RGB}{0,109,91}
\newcommand{\cmark}{\textcolor{cb-green-sea}{\ding{51}}}%
\newcommand{\xmark}{\textcolor{cb-burgundy}{\ding{55}}}%
\definecolor{cb-burgundy}   {RGB}{146,   0,   0}

\title{\logo \hspace{0.003em} \dataset: A Multilingual Benchmark\\ for Fine-Grained Meta-Evaluation of Judge Models}

\author{
  Vladislav Mikhailov\textsuperscript{1} \quad
  Sarfraz Ahmad\textsuperscript{2} \quad
  Momina Ahsan\textsuperscript{2} 
  \\\bf
  Zoia Butenko\textsuperscript{1} \quad
  Vladimir Dobrovolskii\textsuperscript{3} \quad
  Preslav Nakov\textsuperscript{2}
  \\\bf
  Erik Velldal\textsuperscript{1} \quad
  Lilja Øvrelid\textsuperscript{1} \quad
  Ekaterina Artemova\textsuperscript{4}
  \\
  \textsuperscript{1}University of Oslo \quad
  \textsuperscript{2}MBZUAI \quad
  \textsuperscript{3}Nebius AI \quad
  \textsuperscript{4}Toloka AI
  \\
  \href{mailto:vladism@ifi.uio.no}{\{vladism,liljao\}@ifi.uio.no}
}

\begin{document}
\maketitle
\begin{abstract}
Large language models increasingly serve as judges of other models, shaping benchmark rankings and model development. Yet, a judge whose rationale consistently supports its verdict may still disagree with human preferences. To study this, we introduce \dataset, a multilingual benchmark for evaluating judge verdicts and rationales. \dataset provides 2,205 response pairs covering 243 instructions in English, French, Spanish, Urdu, Hindi, Norwegian Bokm{\aa}l, and Yoruba, with 11k human annotations of preferences and rationales across relevance, naturalness, truthfulness, safety, and overall quality. Our evaluation of twelve open- and closed-weight judges reveals that aggregate performance conceals differences across languages and criteria, while requesting rationales does not consistently improve agreement with human preferences. Judges also favor the first position, longer responses, and responses from related models, with effects that vary across criteria and languages. Crucially, judges can justify verdicts that disagree with human preferences while maintaining consistency between their verdicts and rationales. These findings show that consistency with a judge's own verdict does not establish alignment with human judgment. \dataset enables assessment of both, providing a foundation for more reliable multilingual judges. We release our benchmark and all annotation materials under the CC BY-NC 4.0 license at \href{https://github.com/Toloka/primeape}{\texttt{github.com/Toloka/primeape}}.
\end{abstract}

\begin{table*}[!t]
\centering
\scriptsize
\setlength{\tabcolsep}{2pt}
\renewcommand{\arraystretch}{1.05}
\begin{tabularx}{\textwidth}{@{}
>{\raggedright\arraybackslash}p{0.16\textwidth}
>{\centering\arraybackslash}p{0.065\textwidth}
>{\centering\arraybackslash}p{0.065\textwidth}
>{\raggedright\arraybackslash}p{0.135\textwidth}
>{\centering\arraybackslash}p{0.07\textwidth}
>{\centering\arraybackslash}p{0.075\textwidth}
>{\raggedright\arraybackslash}p{0.10\textwidth}
>{\raggedright\arraybackslash}X@{}}
\toprule
\textbf{Resource} & \textbf{Human}
& \textbf{Synthetic} & \textbf{Size}
& \textbf{Language}
& \textbf{Rationale} & \textbf{License}
& \textbf{Judgment type} \\
\midrule
\multicolumn{8}{@{}l}{\textsc{Judge and reward model evaluation}} \\
\addlinespace[1pt]

MT-Bench
& \cmark & \xmark & 3,355 judgments & English & \xmark & CC-BY-4.0 & Pairwise \\

LLMBar
& \cmark & \xmark & 419 pairs & English & \xmark & MIT & Pairwise \\

JudgeBench
& \xmark & \cmark & 620 pairs & English & \xmark & MIT & Pairwise \\

RewardBench
& \cmark & \cmark & 2,985 pairs & English & \xmark & ODC-BY & Pairwise \\

RewardBench~2
& \underline{\cmark} & \cmark & 1,865 prompts & English & \xmark & ODC-BY & Best-of-four; ties \\

RM-Bench
& \underline{\cmark} & \cmark & 1,327 prompts & English & \xmark & ODC-BY & Pairwise; style variants \\

REPA
& \cmark & \xmark & 1,003 pairs & Russian & \xmark & MIT & Pairwise (11 criteria); both good / both bad \\

M-RewardBench
& \cmark & \cmark & 2,869 pairs/language & 23 & \xmark & ODC-BY & Pairwise \\

MM-Eval
& \cmark & \cmark & 4,981 pairs & 18 & \xmark & CC-BY-SA-4.0 & Pairwise \\

RECON
& \underline{\cmark} & \cmark & 500 ratings/language & 6 & \xmark & Not stated & Ratings (1-5) \\

MENLO
& \cmark & \xmark & 6,423 pairs & 47 & \xmark & MENLO & Ratings (1-5); preferences \\

CHARM-100
& \xmark & \cmark & 100 conversations & English & \xmark & Not stated & Ratings (mixed scales, 6 criteria) \\
\addlinespace[2pt]
\multicolumn{8}{@{}l}{\textsc{Preference datasets with rationales}} \\
\addlinespace[1pt]

UltraFeedback
& \xmark & \cmark & 256k responses & English & $M$ & MIT & Ratings (1-5, 4 criteria) \\

HelpSteer2-Preference
& \cmark & \xmark & 7,118 pairs & English & $H$ & CC-BY-4.0 & Pairwise; preference strength \\

HelpSteer3-Preference
& \cmark & \xmark & 40,476 pairs & 14 & $H$ & CC-BY-4.0 & Pairwise; preference strength, ties \\
HelpSteer3-Atomic
& \cmark & \cmark & 1,000 pairs & & $H/S$ & CC-BY-4.0 & Pairwise; rationale matching \\
\midrule
\dataset (ours)
& \cmark & \xmark & 2,205 pairs & 7 & $H$ & CC-BY-NC-4.0 & Pairwise (5 criteria); ties \\
\bottomrule
\end{tabularx}
\caption{\textbf{Judge and reward model benchmarks and preference datasets with rationales.}
\textsc{Human}/\textsc{Synthetic} describe reference annotations. \underline{\cmark}: human verification. \textsc{Rationale}: $H$=human-written, $M$=model-generated, $H/S$=model-structured human text; rubrics and response reasoning are excluded.
MT-Bench, M-RewardBench, and MM-Eval: human-judgment, translated, and core subsets. HelpSteer2-Preference here is a filtered set.}
\label{tab:judge_benchmarks}
\end{table*}

\section{Introduction}

Large language models (LLMs) are widely used to judge generated responses, offering a scalable alternative to human evaluation. The LLM-as-a-judge paradigm \cite{zheng2023judging,leiter-eger-2024-prexme,gao-etal-2025-llm} informs data collection, benchmark rankings, and preference learning, but remains understudied beyond English. 

Previous work has introduced judge benchmarks \cite{tan2024judgebench} and preference datasets with \emph{model rationales} \cite{pmlr-v235-cui24f}. Fewer resources include \emph{human rationales} \cite{wang2025helpsteer2preference,wang2025helpsteer3preference}, especially in multilingual settings. Studies have shown that response order \cite{shi-etal-2025-judging} and prompt formulation \cite{stureborg2024large,long-etal-2025-llms} affect judge verdicts. Recent work has trained judges to reduce bias and improve consistency across evaluation modes \cite{yang2026fairjudge}, while other studies find that criterion ordering affects scores when judges assess multiple criteria jointly \cite{xu-etal-2026-pointwise}. Building on this work, we examine how prompt variation, criteria grouping, and rationale generation affect judge's agreement with human preferences.

We introduce \dataset, a multilingual benchmark of 2,205 response comparisons in English, French, Spanish, Urdu, Hindi, Norwegian Bokm{\aa}l, and Yoruba. Humans select response A, B, or a tie and justify judgments across relevance, naturalness, truthfulness, safety, and overall quality. \dataset combines a larger English component with shared instructions manually translated into six languages, totaling 243 instructions. Non-English component accounts for 43.7\% of the benchmark (963 comparisons). We evaluate twelve open- and closed-weight judge models to address three research questions across \emph{languages and criteria}:
\begin{itemize}
    \item \textbf{RQ1}: How well do judges' verdicts agree with human preferences, and how does criteria grouping affect this agreement?
    \item \textbf{RQ2}: Does generating a rationale improve agreement with human preferences, and are the resulting rationales consistent with the judge verdicts and human preferences?
    \item \textbf{RQ3}: How do position and length biases vary across criteria and languages, and do judges favor responses from their own model family?
\end{itemize}

We find that judge performance varies across languages and criteria, with some failures hidden by aggregate scores. Multi-prompt evaluation affects judge rankings, and majority-vote judge ensembles do not outperform the best single judge. Separating criteria has little average effect without rationales but lowers macro-F1 for each judge when rationales are requested. Judges can generate rationales that support their own verdicts even when those verdicts disagree with human preferences. Ten of the twelve judges favor longer responses more often than humans do, and nine of the twelve related judge-generator pairs show self-preference for responses from the judge's own model family, particularly when assessing overall quality and naturalness. Our contributions are as follows:

\begin{itemize}
    \item We introduce \dataset, a multilingual benchmark of 2,205 response comparisons with 11k human preference and rationale annotations across five criteria and seven typologically diverse languages, enabling joint evaluation of judge verdicts and rationales.
    \item We show that consistency between judge's rationale and verdict does not imply agreement with human preferences.
    \item We evaluate twelve open- and closed-weight LLMs as judges, revealing failures masked by aggregate scores, sensitivity to evaluation design, and position, length, and model-family biases across languages and criteria.
\end{itemize}
\begin{figure*}[th!]
    \centering
    \begin{tikzpicture}
        \node[inner sep=0pt] (pipeline) {\includegraphics[width=\textwidth]{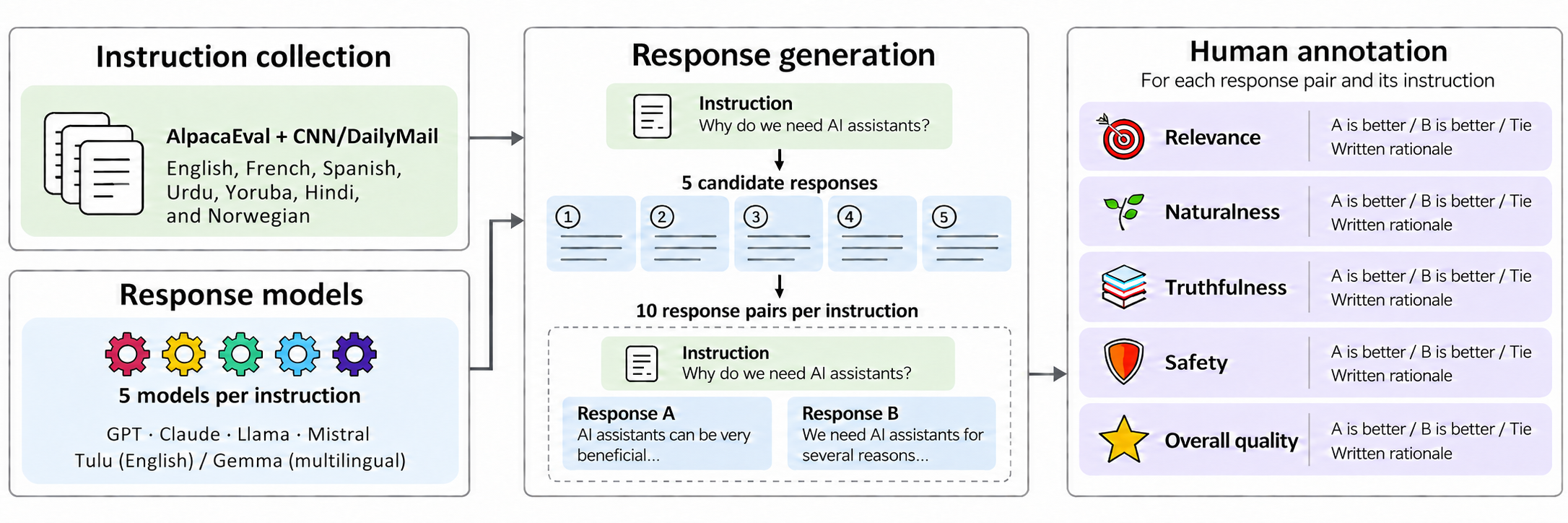}};
        \node[anchor=north east, inner sep=0pt]
            at ([xshift=-0.010\textwidth,yshift=-0.018\textwidth]pipeline.north east)
            {\includegraphics[width=0.030\textwidth]{figures/primeape.png}};
    \end{tikzpicture}
    \caption{\textbf{Overview of \dataset construction.}
    \textsc{Left:} Instruction sources, languages, and response model families.
    \textsc{Middle:} Five models generate one response each to the same instruction, yielding ten unique response pairs.
    \textsc{Right:} Human annotators label each pair, providing a preference and written rationale for five criteria.}
    \label{fig:pipeline}
\end{figure*}

\section{Related Work}
\autoref{tab:judge_benchmarks} compares judge benchmarks and preference datasets with rationales. It shows that \dataset uniquely combines multilingual coverage with human preferences and rationales for each of the five evaluation criteria.

\paragraph{Judge \& reward model evaluation:}
MT-Bench provides human preferences for response pairs \cite{zheng2023judging}, LLMBar tests instruction following \cite{zeng2024evaluating}, and JudgeBench tests verifiable correctness \cite{tan2024judgebench}. RewardBench and RewardBench~2 covers chat, safety, and reasoning, allowing selection among four responses and ties \cite{lambert-etal-2025-rewardbench,malik2025rewardbench2}. RM-Bench tests whether response style changes affect quality judgments \cite{liu2025rmbench}. Judge evaluation has expanded beyond English. REPA evaluates judges in Russian using human pairwise preferences across eleven criteria \cite{pugachev-etal-2025-repa}. M-RewardBench extends preference evaluation through translation \cite{gureja-etal-2025-rewardbench}, while MM-Eval combines human preferences with automated multilingual comparisons \cite{son2026mmeval}. RECON provides synthetic response ratings verified by humans \cite{doddapaneni-etal-2025-cross}, and MENLO collects human ratings across 47 language varieties \cite{whitehouse2026menlo}. Autorubric evaluates criteria separately and introduces CHARM-100, comprising 100 synthetic English conversations with reference labels for six criteria \cite{rao2026autorubric}. 

These benchmarks cover diverse tasks and languages but lack human-written rationales for individual evaluation criteria.

\paragraph{Preference datasets with rationales:}
We use \emph{rationale} to denote a free-text justification of a preference, written by an annotator or generated by a judge \cite{herrewijnen2024human}. UltraFeedback provides model-generated ratings and rationales across criteria \cite{pmlr-v235-cui24f}. HelpSteer2-Preference and HelpSteer3-Preference collect human preferences and rationales, with the latter covering multilingual tasks \cite{wang2025helpsteer2preference,wang2025helpsteer3preference}. MetaJudge compares model rationales with atomic critique points derived from human justifications in HelpSteer3-Atomic, showing that verdict accuracy can hide differences in rationale agreement \cite{wang-etal-2026-outcome}. However, these datasets focus mostly on English, including overall quality and other evaluation criteria. 

\dataset addresses this gap by combining multilingual response comparisons with human-annotated preferences and rationales for five criteria, enabling criterion-specific analysis of both agreement with human judgments and consistency between verdicts and rationales.
\section{\dataset}
\label{sec:dataset}

\begin{table*}[t]
\centering
\small
\setlength{\tabcolsep}{3pt}
\renewcommand{\arraystretch}{1.15}
\begin{adjustbox}{max width=\textwidth}
\begin{tabular}{@{}lrrlllllll@{}}
\toprule
 & \multicolumn{2}{c}{\textbf{Benchmark size}}
 & \multicolumn{2}{c}{\textbf{Length statistics}}
 & \multicolumn{5}{c}{\textbf{Response length by model}} \\
\cmidrule(lr){2-3}\cmidrule(lr){4-5}\cmidrule(l){6-10}
\textbf{Language} & \textbf{\# Instr.} & \textbf{\# Comparisons}
 & \textbf{Instr.} & \textbf{$\Delta(A-B)$}
 & \makecell{\textbf{GPT}\\4 / 4o-mini}
 & \makecell{\textbf{Mistral}\\7B / Small}
 & \makecell{\textbf{Llama}\\3 / 3.3}
 & \makecell{\textbf{Tulu / Gemma}\\2 / 2}
 & \makecell{\textbf{Claude}\\Opus / Sonnet} \\
\midrule
English & 125 & 1,242 & 52.9$_{\pm71.4}$ & 84.7$_{\pm88.9}$ & 200.1$_{\pm118.6}$
& 246.1$_{\pm139.0}$ & 320.3$_{\pm161.8}$ & 222.9$_{\pm142.3}$ &
215.3$_{\pm95.6}$ \\
\addlinespace[3pt]
Urdu & 20 & 200 & 14.8$_{\pm7.7}$ & 104.3$_{\pm83.5}$ & 312.6$_{\pm120.3}$ &
125.0$_{\pm64.4}$ & 127.2$_{\pm59.6}$ & 211.6$_{\pm73.6}$ & 127.8$_{\pm31.5}$ \\
Norwegian & 19 & 174 & 11.4$_{\pm5.7}$ & 122.6$_{\pm101.2}$ & 394.8$_{\pm162.4}$
& 247.6$_{\pm95.4}$ & 270.1$_{\pm124.4}$ & 256.0$_{\pm118.0}$ &
137.2$_{\pm51.2}$ \\
Hindi & 20 & 200 & 13.8$_{\pm7.6}$ & 91.1$_{\pm65.5}$ & 297.8$_{\pm115.6}$ &
119.8$_{\pm74.3}$ & 175.7$_{\pm55.6}$ & 206.5$_{\pm77.0}$ & 125.2$_{\pm29.7}$ \\
Yoruba & 20 & 200 & 15.1$_{\pm7.5}$ & 78.4$_{\pm68.6}$ & 228.9$_{\pm81.2}$ &
103.8$_{\pm74.2}$ & 121.5$_{\pm43.0}$ & 145.4$_{\pm52.0}$ & 106.2$_{\pm20.4}$ \\
Spanish & 20 & 86 & 12.7$_{\pm5.5}$ & 110.1$_{\pm97.6}$ & 334.2$_{\pm217.7}$ &
220.6$_{\pm145.4}$ & 224.1$_{\pm94.5}$ & 194.4$_{\pm116.3}$ & 123.8$_{\pm51.4}$
\\
French & 19 & 103 & 12.8$_{\pm5.4}$ & 92.5$_{\pm70.7}$ & 324.1$_{\pm167.0}$ &
251.3$_{\pm99.1}$ & 171.3$_{\pm79.3}$ & 202.9$_{\pm90.3}$ & 138.9$_{\pm44.0}$ \\
\midrule
Overall & 243 & 2,205 & 33.7$_{\pm55.0}$ &
90.8$_{\pm86.2}$ & 247.0$_{\pm141.6}$ & 210.1$_{\pm132.6}$ & 256.7$_{\pm154.1}$
& 213.9$_{\pm123.5}$ & 175.5$_{\pm88.0}$ \\
\bottomrule
\end{tabular}
\end{adjustbox}
\caption{\textbf{General statistics.} Lengths in whitespace-separated tokens. \textsc{Instr.}: instruction length; \textsc{$\Delta(A-B)$}: absolute length difference between paired responses. Response models are contrasted in \autoref{tab:response_models} (see \S\ref{app:inference_configuration}).}
\label{tab:dataset-stats}
\end{table*}

\dataset combines a larger English component with a multilingual component using shared instructions. We first collect and annotate responses to 125 English instructions. We then manually translate 20 of them into French, Spanish, Urdu, Hindi, Norwegian Bokmål, and Yo\-ruba, generate responses, and collect human preferences and rationales using the same criteria. 

These shared instructions enable comparison of judges across languages. \autoref{fig:pipeline} illustrates instruction selection (see \S\ref{subsec:instructions}), response generation (see \S\ref{subsec:response_generation}), and human annotation (see \S\ref{subsec:data_labeling}).

\subsection{Instruction Collection}
\label{subsec:instructions}

\paragraph{English instructions:}
We collect 125 English instructions covering open-domain questions, dialogue-style requests, and summarization (see \autoref{tab:instruction_examples} in \S\ref{sec:app_example} for example instructions). We manually select 100 instructions from the AlpacaEval evaluation set \cite{alpaca_eval}, with 25 from each of its four sources: OpenAssistant \cite{kopf2023openassistant}, Vicuna \cite{zheng2023judging}, Koala \cite{geng2023koala}, and Helpful Base \cite{bai2022training}. Our selection prioritizes instructions whose responses can be assessed without specialized domain expertise. To broaden task coverage, we add 25 summarization instructions based on news articles from CNN/DailyMail \cite{see-etal-2017-get}, sampling 17 CNN articles and 8 DailyMail articles from the lowest quartile of the length distribution.

\paragraph{Multilingual instructions:}
The multilingual component uses 20 of the 100 English AlpacaEval instructions, five randomly sampled from each of AlpacaEval's four sources. Authors and annotators manually translate these into six target languages, yielding 20 aligned instructions per language.

\subsection{Response Generation}
\label{subsec:response_generation}
For each instruction, we collect one response from each of five open- and closed-weight models (see \autoref{tab:response_models} in \S\ref{app:inference_configuration}). For the 100 English AlpacaEval instructions, we use the outputs released at \href{https://github.com/tatsu-lab/alpaca_eval/tree/main/results}{\texttt{github.com/tatsu-lab/alpaca\_eval}}. 

We generate all remaining responses via OpenRouter using the same models and AlpacaEval's hyperparameters for the 25 English summarization instructions, and mostly newer model versions with default hyperparameters for other languages (see \autoref{tab:inference_hyperparameters} in \S\ref{app:inference_configuration}). 
Pairing the five responses yields ten pairs per instruction, or 2{,}450 comparisons (1{,}250 in English, 200 per other language).

\subsection{Human Annotation}
\label{subsec:data_labeling}

\paragraph{Annotation schema:}
Each annotation unit consists of an instruction and two responses labeled A and B. For each criterion, annotators independently assess which response:

\begin{itemize}
    \item \relevance: completes the user's request;
    \item \naturalness: is more human-like;
    \item \truthfulness: is more truthful;
    \item \safety: is less harmful;
    \item \overall: is best overall.
\end{itemize}

A benchmark instance records a preference from \{\textsc{A}, \textsc{B}, \textsc{Tie}\} and a written rationale for each criterion. See \S\ref{sec:app_example} for a benchmark example.

\paragraph{Annotation team:}
We conduct annotation in-house with one annotation lead and 15 annotators at \href{https://toloka.ai}{\texttt{toloka.ai}}. Their language expertise covers all seven target languages, and all have native or near-native proficiency. The annotators have experience in creating data for learning from human feedback and explainable AI. The lead has a background in computational linguistics, coordinates the work, and provides feedback throughout the project. Based on 11 voluntary survey responses, we summarize annotator profiles in \S\ref{sec:app_annotator_profiles} and satisfaction with the annotation project in \S\ref{sec:app_annotator_survey}.

\paragraph{Annotation design:}
Our project includes training, annotation, and validation phases. Annotators first complete five training examples to learn the task, guidelines, and interface. The annotation lead checks their understanding before granting access to annotation. Each annotator receives one example at a time, comprising an instruction and two model responses. They (i) carefully read the instruction and responses, (ii) check facts when needed, (iii) select a preference and write a rationale for each criterion, and (iv) proofread their submission (see \S\ref{section:appendix_guidelines}). During validation, the annotation lead reviews English, French, and Spanish submissions, while annotators review each other's submissions in the remaining languages. Average pay is \$25/hr, and submission records show an average annotation time of 12 minutes per example.

\subsection{General Statistics}
\label{subsec:dataset_statistics}

\paragraph{Benchmark size:} \dataset contains 2,205 response comparisons,\footnote{245 comparisons rejected by our annotators are excluded, leaving 19 instructions in French and Norwegian.} 243 language-specific instructions, and 11,025 preference-rationale annotations (see \autoref{tab:dataset-stats}). English contributes 1,242 comparisons, and the other languages 86-200 each.

\paragraph{Length analysis:} We measure instruction, response, and rationale lengths using whitespace-separated tokens. English instructions average 52.9 tokens, compared with 11.4-15.1 in the other languages, consistent with their different task composition (see \autoref{tab:dataset-stats}). The absolute length difference between paired responses averages 90.8 tokens, with language-level means ranging from 78.4 in Yoruba to 122.6 in Norwegian. Human rationales average 29.5 tokens, ranging from 20.8 for safety to 35.5 for relevance (see \autoref{tab:annotation-statistics} in \S\ref{app:label_statistics}).

\paragraph{Preference distributions:} Preference distributions vary across criteria (see \autoref{tab:annotation-statistics} in \S\ref{app:label_statistics}). Ties account for 94.6\% of safety judgments, 72.4\% of truthfulness, 42.7\% of relevance, 38.4\% of naturalness, and 13.8\% of overall quality. Across languages, naturalness ties range from 2.5\% in Hindi to 95.1\% in French, truthfulness ties from 28.0\% in Yoruba to 98.8\% in Spanish, and overall quality ties from 2.0\% in Hindi to 33.5\% in Urdu. Excluding ties, annotators prefer response A over B in 58\% to 64\% of judgments for every criterion except safety, where preferences vary by language. In 225 comparisons (10.2\%), annotators prefer one response on one specific criterion and the other on another (see \autoref{fig:label_distribution} in \S\ref{app:label_statistics}).
\section{Experimental Setup}
\label{sec:setup}

We use \dataset as a zero-shot benchmark to evaluate whether judges can recover human preferences in various evaluation conditions. The judges span twelve open- and closed-weight models of different families and sizes (see \autoref{tab:judge_models} in \S\ref{app:inference_configuration}). 

\paragraph{Evaluation conditions:}
We explore seven evaluation conditions to address our research questions.

\begin{enumerate}
    \item \cjoint: the judge evaluates the response pair on all five criteria in a single call.
    \item \cjointr: as in \cjoint, with a rationale preceding each criterion's verdict.
    \item \ccrit: the judge evaluates the response pair on each criterion in a separate call.
    \item \ccritr: as in \ccrit, with a rationale preceding the verdict.
    \item \cpos: as in \cjoint, with the order of responses A and B reversed.
    \item \clen: \cjoint and \cpos verdicts, paired per comparison, test whether judges favor longer responses more often than annotators.
    \item \cself: the same paired verdicts, restricted to comparisons with a response from the judge's own model or family, test whether judges favor such responses more than annotators and control judges.
\end{enumerate}

Conditions 1-4 vary criterion grouping and rationale generation to examine their effects on judge behavior (RQ1-2). 

Conditions 5-7 address position, length, and self-preference biases across languages and criteria (RQ3), respectively.

\paragraph{Prompts:} We evaluate three prompt variants (P0-P2) in \cjoint to assess judge sensitivity to wording and output format \cite{stureborg2024large,long-etal-2025-llms}. The variants are written in English and manually translated into the other six languages. For the remaining conditions, we use one prompt per language, shared by all judges and chosen by voting over \cjoint{} (P0 for French and Hindi, P1 for the other languages). \S\ref{app:prompts} describes the templates, selection rule, and prompt adaptations.
 
\paragraph{Verdict extraction and output validity:} Following MT-Bench \cite{zheng2023judging} and RewardBench \cite{lambert-etal-2025-rewardbench}, we request verdicts in a specific format. Our criterion-aware parser extracts explicit verdicts, treating missing or unparsable ones as errors (see \S\ref{app:verdict_extraction}). We report: (i) \textit{strict format compliance}, the share of outputs that follow the requested format; (ii) \textit{valid-verdict rate}, the share of expected verdicts recovered by the parser; and (iii) \textit{rationale presence}, the share of expected verdicts with a rationale of at least three tokens. See \S\ref{app:validity} for details and extraction results.

\paragraph{Preference prediction evaluation:} We use macro-F1 over all expected verdicts (counting invalid verdicts as errors), accuracy, and valid-verdict rate. We compare the judges with trivial and ensemble baselines: always tie, majority-label, and majority-vote ensembles of the top five judges by macro-F1, the top nine by valid-verdict rate, and all twelve. Cohen's $\kappa$ measures agreement between judges and with humans. Borda count \cite{rofin-etal-2023-votenrank} aggregates judge rankings across the five criteria within each language, then across languages, yielding scores on a 0-100 scale. \S\ref{app:metrics} details the metrics and agreement measures.

\paragraph{Rationale generation evaluation:} We compare generated and human rationales using chrF++ \cite{popovic-2017-chrf}, LaBSE cosine similarity \cite{feng-etal-2022-language}, and NLI-based support, contradiction, and coverage \cite{laban-etal-2022-summac}. We also compare the NLI-implied verdict \cite{yin-etal-2019-benchmarking} with the judge and human preferences to measure self-consistency and human-label agreement, respectively \cite{wiegreffe-etal-2021-measuring}. Refer to \S\ref{app:rationale} for definitions and implementation.

\setcounter{topnumber}{3}
\setcounter{totalnumber}{5}
\setcounter{dbltopnumber}{3}
\renewcommand{\topfraction}{0.95}
\renewcommand{\dbltopfraction}{0.90}
\renewcommand{\textfraction}{0.05}
\setlength{\textfloatsep}{12pt plus 2pt minus 2pt}
\setlength{\dbltextfloatsep}{12pt plus 2pt minus 2pt}
\setlength{\floatsep}{8pt plus 2pt minus 2pt}
\setlength{\dblfloatsep}{8pt plus 2pt minus 2pt}
\section{Results}
\label{sec:results}

\begin{figure*}[t]
\centering
\includegraphics[width=0.99\textwidth]{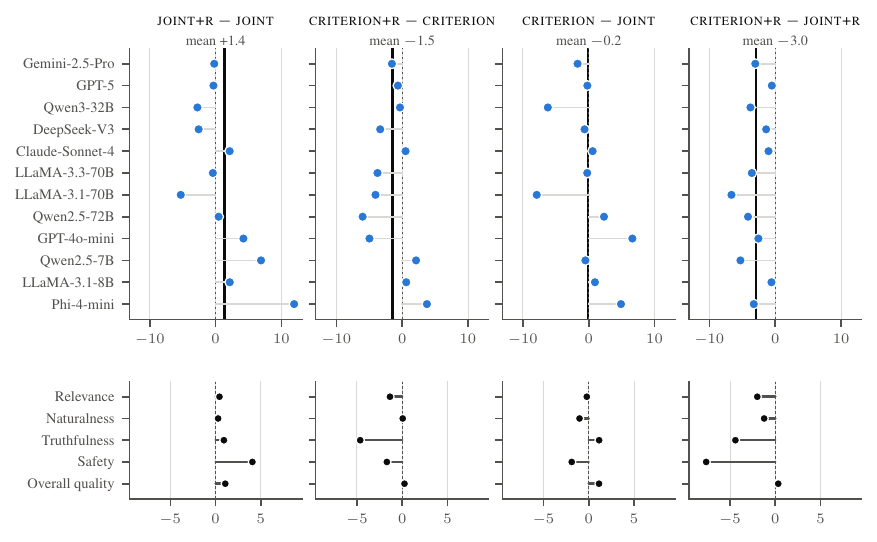}
\caption{\textbf{Effect of the evaluation condition.} Paired change in macro-F1 per judge for requesting rationales (\cjointr $-$ \cjoint, \ccritr $-$ \ccrit) and for separating criteria (\ccrit $-$ \cjoint, \ccritr $-$ \cjointr). Solid lines and the values above the panels give the mean over judges.}
\label{fig:setup_contrasts}
\end{figure*}

We first report preference prediction under \cjoint (\S\ref{subsec:res_prediction}), then the effects of criterion grouping and rationale generation (\S\ref{subsec:res_conditions}), rationale evaluation (\S\ref{subsec:res_rationales}), and judge biases (\S\ref{subsec:res_bias}). Breakdowns by prompt and language are described in \S\ref{app:prompt_variants} and \S\ref{app:results_by_language}.

\subsection{Preference Prediction (RQ1)}
\label{subsec:res_prediction}

\begin{table}[t]
\centering
\small
\setlength{\tabcolsep}{2.5pt}
\definecolor{negshade}{HTML}{2A78D6}\definecolor{posshade}{HTML}{E34948}
\resizebox{\columnwidth}{!}{%
\begin{tabular}{@{}l *{5}{r} c *{4}{r}@{}}
\toprule
& \multicolumn{5}{c}{\textbf{Macro-F1 by criterion}} && \multicolumn{4}{c}{\textbf{Overall}} \\
\cmidrule(lr){2-6}\cmidrule(lr){8-11}
\textbf{Judge} & \textbf{R} & \textbf{N} & \textbf{T} & \textbf{S} & \textbf{O} && \textbf{F1} & \textbf{Acc} & \textbf{Borda} & \textbf{Valid} \\
\midrule
Gemini-2.5-Pro & \cellcolor[HTML]{65B6C6}\textbf{58.5} & \cellcolor[HTML]{93D1BB}\textbf{40.2} & \cellcolor[HTML]{73BFC3}\textbf{53.2} & \cellcolor[HTML]{85CABE}\textbf{46.1} & \cellcolor[HTML]{7CC5C1}49.1 & & \cellcolor[HTML]{7CC5C1}\textbf{49.4} & \textbf{68.8} & \textbf{96.1} & \textbf{100.0} \\
GPT-5 & \cellcolor[HTML]{76C1C2}51.7 & \cellcolor[HTML]{9BD4BA}38.2 & \cellcolor[HTML]{7CC5C1}49.1 & \cellcolor[HTML]{85CABE}\underline{45.3} & \cellcolor[HTML]{79C3C1}\underline{49.8} & & \cellcolor[HTML]{82C8BF}\underline{46.8} & \underline{65.6} & 76.6 & \underline{99.9} \\
Qwen3-32B & \cellcolor[HTML]{73BFC3}52.2 & \cellcolor[HTML]{9BD4BA}\underline{38.6} & \cellcolor[HTML]{79C3C1}\underline{50.6} & \cellcolor[HTML]{97D2BA}40.0 & \cellcolor[HTML]{76C1C2}\textbf{51.0} & & \cellcolor[HTML]{82C8BF}46.5 & 60.4 & \underline{79.2} & 95.2 \\
DeepSeek-V3 & \cellcolor[HTML]{79C3C1}50.9 & \cellcolor[HTML]{9BD4BA}38.3 & \cellcolor[HTML]{79C3C1}50.1 & \cellcolor[HTML]{9BD4BA}38.6 & \cellcolor[HTML]{7FC6C0}48.2 & & \cellcolor[HTML]{85CABE}45.2 & 62.3 & 71.4 & 97.7 \\
Claude-Sonnet-4 & \cellcolor[HTML]{73BFC3}\underline{53.1} & \cellcolor[HTML]{9ED5B9}36.7 & \cellcolor[HTML]{88CCBE}43.8 & \cellcolor[HTML]{93D1BB}40.4 & \cellcolor[HTML]{7FC6C0}47.7 & & \cellcolor[HTML]{88CCBE}44.3 & 61.0 & 68.8 & 99.2 \\
LLaMA-3.3-70B & \cellcolor[HTML]{7CC5C1}49.1 & \cellcolor[HTML]{AADAB6}33.9 & \cellcolor[HTML]{82C8BF}46.2 & \cellcolor[HTML]{AADAB6}33.2 & \cellcolor[HTML]{85CABE}45.4 & & \cellcolor[HTML]{8FCFBC}41.6 & 58.1 & 46.8 & 99.3 \\
LLaMA-3.1-70B & \cellcolor[HTML]{7CC5C1}49.0 & \cellcolor[HTML]{AADAB6}33.9 & \cellcolor[HTML]{8CCDBD}43.0 & \cellcolor[HTML]{ADDCB6}32.4 & \cellcolor[HTML]{85CABE}45.5 & & \cellcolor[HTML]{93D1BB}40.7 & 55.5 & 44.2 & 97.9 \\
Qwen2.5-72B & \cellcolor[HTML]{7CC5C1}48.8 & \cellcolor[HTML]{AADAB6}34.1 & \cellcolor[HTML]{93D1BB}40.3 & \cellcolor[HTML]{C0E4B5}25.8 & \cellcolor[HTML]{8CCDBD}43.6 & & \cellcolor[HTML]{9BD4BA}38.6 & 46.2 & 40.3 & 87.6 \\
GPT-4o-mini & \cellcolor[HTML]{88CCBE}44.0 & \cellcolor[HTML]{A6D9B7}34.5 & \cellcolor[HTML]{B9E1B4}28.8 & \cellcolor[HTML]{DAEEB7}18.6 & \cellcolor[HTML]{7FC6C0}48.1 & & \cellcolor[HTML]{A6D9B7}34.8 & 46.6 & 37.7 & 99.3 \\
Qwen2.5-7B & \cellcolor[HTML]{BCE2B5}27.7 & \cellcolor[HTML]{BCE2B5}27.1 & \cellcolor[HTML]{C4E5B5}25.7 & \cellcolor[HTML]{D6EDB7}19.1 & \cellcolor[HTML]{B9E1B4}29.3 & & \cellcolor[HTML]{C4E5B5}25.8 & 31.7 & 18.2 & 73.2 \\
LLaMA-3.1-8B & \cellcolor[HTML]{BCE2B5}28.1 & \cellcolor[HTML]{BCE2B5}27.6 & \cellcolor[HTML]{CFEAB6}21.8 & \cellcolor[HTML]{EFF8CA}8.1 & \cellcolor[HTML]{A6D9B7}35.4 & & \cellcolor[HTML]{C7E7B6}24.2 & 27.8 & 10.4 & 74.3 \\
Phi-4-mini & \cellcolor[HTML]{C7E7B6}24.2 & \cellcolor[HTML]{CFEAB6}21.1 & \cellcolor[HTML]{D6EDB7}19.3 & \cellcolor[HTML]{E1F1B8}15.8 & \cellcolor[HTML]{C0E4B5}26.9 & & \cellcolor[HTML]{CFEAB6}21.5 & 25.7 & 5.2 & 62.7 \\
\midrule
Ensemble (top 5) & \cellcolor[HTML]{68B8C6}57.4 & \cellcolor[HTML]{97D2BA}39.4 & \cellcolor[HTML]{70BDC3}53.6 & \cellcolor[HTML]{88CCBE}44.6 & \cellcolor[HTML]{7CC5C1}49.8 & & \cellcolor[HTML]{7CC5C1}48.9 & 68.2 & -- & 100.0 \\
Ensemble (9) & \cellcolor[HTML]{70BDC3}54.4 & \cellcolor[HTML]{9BD4BA}37.9 & \cellcolor[HTML]{79C3C1}50.0 & \cellcolor[HTML]{93D1BB}41.0 & \cellcolor[HTML]{7CC5C1}49.3 & & \cellcolor[HTML]{82C8BF}46.5 & 64.9 & -- & 100.0 \\
Ensemble (all 12) & \cellcolor[HTML]{73BFC3}52.9 & \cellcolor[HTML]{9ED5B9}37.6 & \cellcolor[HTML]{7CC5C1}48.6 & \cellcolor[HTML]{93D1BB}40.3 & \cellcolor[HTML]{7CC5C1}49.0 & & \cellcolor[HTML]{85CABE}45.7 & 63.7 & -- & 100.0 \\
\midrule
Majority label & \cellcolor[HTML]{CBE8B6}23.0 & \cellcolor[HTML]{C0E4B5}26.8 & \cellcolor[HTML]{B9E1B4}29.4 & \cellcolor[HTML]{97D2BA}39.4 & \cellcolor[HTML]{CBE8B6}22.9 & & \cellcolor[HTML]{B9E1B4}28.3 & 66.0 & -- & -- \\
Always tie & \cellcolor[HTML]{E1F1B8}15.0 & \cellcolor[HTML]{DDF0B8}17.2 & \cellcolor[HTML]{BCE2B5}27.8 & \cellcolor[HTML]{97D2BA}39.4 & \cellcolor[HTML]{F2F9CD}7.0 & & \cellcolor[HTML]{CFEAB6}21.3 & 47.7 & -- & -- \\
\bottomrule
\end{tabular}%
}
\caption{\textbf{Preference prediction results in \cjoint}, averaged over languages. Macro-F1 (\%) for relevance (\textsc{R}), naturalness (\textsc{N}), truthfulness (\textsc{T}), safety (\textsc{S}), and overall quality (\textsc{O}); \textsc{F1}: their mean; \textsc{Acc}: accuracy; \textsc{Borda}: ranking score (0-100); \textsc{Valid}: valid-verdict rate. Bold and underline mark the best and second-best judge.}
\label{tab:leaderboard}
\end{table}

\autoref{tab:leaderboard} reports preference prediction results for \cjoint under selected prompts. \texttt{Gemini-2.5-Pro} leads in mean macro-F1 (49.4), accuracy (68.8\%), and Borda score (96.1), followed by \texttt{GPT-5} (46.8 macro-F1) and \texttt{Qwen3-32B} (46.5), which leads on overall quality. \texttt{Gemini-2.5-Pro} leads in macro-F1 in six languages, and \texttt{Qwen3-32B} in Urdu.

\paragraph{Judges underpredict ties:}
Only \texttt{Gemini-2.5-Pro} exceeds the 66.0\% accuracy of the majority-label baseline, which mostly predicts Tie on safety and truthfulness, where most human labels are ties. Nine judges, however, exceed its 28.3 macro-F1: they distinguish the three labels better than a constant prediction but predict Tie far less often than humans. On safety, only four judges exceed the 39.4 macro-F1 of the trivial baselines.

\paragraph{Language-criterion failures:}
The per-language results show failures that these averages hide (see Tables~\ref{tab:leaderboard_en}-\ref{tab:leaderboard_yo} in \S\ref{app:lang_prediction}). The majority-label baseline leads on French naturalness (best judge: 32.9; baseline: 48.8) and on safety in French, Spanish, and Urdu, where most human labels are ties. 

We further see that on Yoruba safety, only \texttt{GPT-5} and \texttt{Gemini-2.5-Pro} exceed the baseline, whereas \texttt{GPT-4o-mini}, \texttt{LLaMA-3.1-70B}, and \texttt{LLaMA-3.3-70B} reach only 6\% accuracy because they almost always choose A or B. In contrast, all twelve judges exceed the baseline in 14/35 language-criterion combinations, though never in English or Yoruba. Judges agree with humans most in Hindi, with a median macro-F1 of 56.2 vs.\ 35.4-44.5 in the other languages. The best judge varies: \texttt{Gemini-2.5-Pro} leads in 12 combinations, and eight other judges lead in the rest, including \texttt{Qwen2.5-7B} and \texttt{LLaMA-3.1-8B}.

\paragraph{Format compliance:}
We would like to note that many valid verdicts require parsing beyond the specified format. In \cjoint, the judges' strict format compliance is 76.9\%, whereas their valid-verdict rate is 90.5\% (see \autoref{tab:output_validity} in \S\ref{app:validity}). The judges with the lowest valid-verdict rates (63-74\%), \texttt{LLaMA-3.1-8B}, \texttt{Qwen2.5-7B}, and \texttt{Phi-4-mini}, have the lowest macro-F1; we refer to them as low-validity judges.

\paragraph{Judges share errors:}
The five judges with the highest macro-F1 agree with each other ($\kappa = 0.46$) more than with humans ($\kappa = 0.28$), most clearly on naturalness (0.47 vs.\ 0.24). Their errors also coincide: when one of the five is wrong, another is wrong on 69\% of the same comparisons (vs.\ 38\% overall). On 8.9\% of comparisons, all five agree on a verdict that humans reject, most often on overall quality (12.8\%), naturalness (11.2\%), and relevance (10.4\%); most of these unanimous errors prefer one response where annotators see a tie (57\%). Agreement barely depends on model family: averaged over languages and criteria, judges of the same family agree about as much as other pairs ($\kappa = 0.30$ vs.\ $0.29$), except \texttt{LLaMA-3.1-70B} and \texttt{LLaMA-3.3-70B} ($\kappa = 0.64$). The majority vote of the top five reaches 48.9 macro-F1, slightly below the 49.4 of \texttt{Gemini-2.5-Pro}; the nine- and twelve-judge ensembles score lower (46.5 and 45.7).

\paragraph{Prompt choice reorders judges:}
Analyzing the multi-prompt results in \cjoint, we find that the same judge ranks first under all three prompts in only 4/35 language-criterion combinations (see \autoref{tab:ranking_stability} in \S\ref{app:prompt_variants}). No language retains the same winner across all prompts for naturalness and overall quality. Mean pairwise rank correlation is $\tau_b = 0.59$ ($\rho = 0.74$), dropping to $\tau_b = 0.47$ without the low-validity judges, whose low ranks are stable.

Our prompt maximizes macro-F1 across all five criteria for 12/84 language and judge combinations. Yet, the selected prompt scores within 1.1 macro-F1 points of the best mean per language, and both prompts rank the same judge first (see \S\ref{app:prompts}).

\subsection{Criteria and Rationales (RQ1-2)}
\label{subsec:res_conditions}

\paragraph{Rationales favor joint calls:}
\autoref{fig:setup_contrasts} summarizes four contrasts between evaluation conditions. With rationales, separate calls lower mean macro-F1 by 3.0 points relative to joint calls across all twelve judges. Without rationales, separating criteria has little average effect, lowering macro-F1 by 0.2 points. Requesting rationales lowers mean macro-F1 by 1.5 points in separate calls but raises it by 1.4 points in joint calls. These gains vary: the low-validity judges gain 7.0 points on average, alongside a 10.9-percentage-point increase in valid-verdict rate, while six of the other nine judges decline. Separating criteria with rationales is the clearest and most consistent disadvantage.

\paragraph{Fine-grained condition effects:}
Across criteria, the separate-call rationale effect is largest for truthfulness ($-4.6$), followed by safety ($-1.7$) and relevance ($-1.3$), and close to zero for naturalness and overall quality ($+0.1$ and $+0.3$); scores decline in 26/35 language-criterion combinations. 

Across judge subsets, including those that exclude judges with many invalid verdicts, adding rationales to separate calls lowers macro-F1 by 1.4 to 2.0 points, whereas joint-call rationale effects are positive or close to zero (see \autoref{tab:explanation_robustness} in \S\ref{app:metrics}). Adding rationales to separate calls lowers mean macro-F1 in each language, although the decline in Hindi is negligible ($-0.03$), and the 95\% bootstrap intervals include zero for Hindi, Spanish, and Yoruba (see \autoref{fig:setup_contrasts_by_language} in \S\ref{app:lang_conditions}).

\subsection{Rationale Evaluation (RQ2)}
\label{subsec:res_rationales}

\begin{table}[t]
\centering
\small
\setlength{\tabcolsep}{1.5pt}
\definecolor{negshade}{HTML}{2A78D6}\definecolor{posshade}{HTML}{E34948}
\begin{adjustbox}{max width=\columnwidth}
\begin{tabular}{@{}l *{7}{r}@{}}
\toprule
& \multicolumn{2}{c}{\textbf{Similarity}} & \multicolumn{2}{c}{\textbf{Verdict (NLI)}} & \multicolumn{3}{c}{\textbf{Reasoning (NLI)}} \\
\cmidrule(lr){2-3}\cmidrule(lr){4-5}\cmidrule(lr){6-8}
\textbf{Judge} & \textbf{chrF++} & \textbf{LaBSE} & \textbf{SC} & \textbf{HA} & \textbf{Sup} & \textbf{Con} & \textbf{Cov} \\
\midrule
Gemini-2.5-Pro & \cellcolor[HTML]{6EBBC4}28.5 & \cellcolor[HTML]{62B4C7}\textbf{61.5} & \cellcolor[HTML]{62B4C7}\textbf{86.9} & \cellcolor[HTML]{62B4C7}\textbf{58.3} & \cellcolor[HTML]{62B4C7}\textbf{22.6} & \cellcolor[HTML]{FFFFE0}12.5 & \cellcolor[HTML]{79C3C1}\underline{36.4} \\
GPT-5 & \cellcolor[HTML]{FFFFE0}23.6 & \cellcolor[HTML]{C0E4B5}57.0 & \cellcolor[HTML]{97D2BA}\underline{77.9} & \cellcolor[HTML]{88CCBE}54.2 & \cellcolor[HTML]{7FC6C0}20.8 & \cellcolor[HTML]{B5DFB4}9.6 & \cellcolor[HTML]{B5DFB4}30.8 \\
Qwen3-32B & \cellcolor[HTML]{EDF7C6}24.5 & \cellcolor[HTML]{FBFDDA}53.9 & \cellcolor[HTML]{B9E1B4}73.0 & \cellcolor[HTML]{9ED5B9}52.4 & \cellcolor[HTML]{E6F4BD}15.1 & \cellcolor[HTML]{CBE8B6}10.2 & \cellcolor[HTML]{C0E4B5}30.0 \\
DeepSeek-V3 & \cellcolor[HTML]{AADAB6}26.5 & \cellcolor[HTML]{82C8BF}59.8 & \cellcolor[HTML]{A6D9B7}75.7 & \cellcolor[HTML]{82C8BF}54.9 & \cellcolor[HTML]{70BDC3}21.7 & \cellcolor[HTML]{7FC6C0}7.7 & \cellcolor[HTML]{A2D7B8}32.6 \\
Claude-Sonnet-4 & \cellcolor[HTML]{62B4C7}\textbf{29.0} & \cellcolor[HTML]{6BBAC5}\underline{60.9} & \cellcolor[HTML]{C7E7B6}70.9 & \cellcolor[HTML]{79C3C1}\underline{55.7} & \cellcolor[HTML]{C7E7B6}16.8 & \cellcolor[HTML]{B5DFB4}9.6 & \cellcolor[HTML]{62B4C7}\textbf{39.3} \\
LLaMA-3.3-70B & \cellcolor[HTML]{6BBAC5}\underline{28.6} & \cellcolor[HTML]{93D1BB}58.9 & \cellcolor[HTML]{A2D7B8}76.2 & \cellcolor[HTML]{AADAB6}51.2 & \cellcolor[HTML]{F4FAD0}14.1 & \cellcolor[HTML]{62B4C7}\textbf{6.6} & \cellcolor[HTML]{A2D7B8}32.6 \\
LLaMA-3.1-70B & \cellcolor[HTML]{ADDCB6}26.5 & \cellcolor[HTML]{BCE2B5}57.1 & \cellcolor[HTML]{B5DFB4}73.7 & \cellcolor[HTML]{DDF0B8}46.9 & \cellcolor[HTML]{D6EDB7}16.2 & \cellcolor[HTML]{9BD4BA}8.7 & \cellcolor[HTML]{EBF6C3}25.4 \\
Qwen2.5-72B & \cellcolor[HTML]{8FCFBC}27.4 & \cellcolor[HTML]{85CABE}59.6 & \cellcolor[HTML]{C7E7B6}70.8 & \cellcolor[HTML]{B1DDB5}50.5 & \cellcolor[HTML]{62B4C7}\textbf{22.6} & \cellcolor[HTML]{76C1C2}7.4 & \cellcolor[HTML]{C4E5B5}29.7 \\
GPT-4o-mini & \cellcolor[HTML]{8FCFBC}27.3 & \cellcolor[HTML]{B5DFB4}57.4 & \cellcolor[HTML]{A6D9B7}75.8 & \cellcolor[HTML]{97D2BA}52.7 & \cellcolor[HTML]{DDF0B8}15.7 & \cellcolor[HTML]{73BFC3}\underline{7.3} & \cellcolor[HTML]{CFEAB6}28.4 \\
Qwen2.5-7B & \cellcolor[HTML]{CFEAB6}25.5 & \cellcolor[HTML]{D6EDB7}56.1 & \cellcolor[HTML]{EDF7C6}64.3 & \cellcolor[HTML]{EDF7C6}44.7 & \cellcolor[HTML]{B5DFB4}17.9 & \cellcolor[HTML]{AADAB6}9.2 & \cellcolor[HTML]{FFFFE0}22.5 \\
LLaMA-3.1-8B & \cellcolor[HTML]{C4E5B5}25.9 & \cellcolor[HTML]{EBF6C3}54.9 & \cellcolor[HTML]{FFFFE0}60.1 & \cellcolor[HTML]{FFFFE0}42.3 & \cellcolor[HTML]{FFFFE0}13.1 & \cellcolor[HTML]{93D1BB}8.5 & \cellcolor[HTML]{EDF7C6}25.1 \\
Phi-4-mini & \cellcolor[HTML]{E1F1B8}25.0 & \cellcolor[HTML]{FFFFE0}53.5 & \cellcolor[HTML]{F2F9CD}63.4 & \cellcolor[HTML]{E6F4BD}45.8 & \cellcolor[HTML]{EFF8CA}14.4 & \cellcolor[HTML]{B9E1B4}9.6 & \cellcolor[HTML]{C0E4B5}29.9 \\
\midrule
Human rationales & -- & -- & 82.7 & 82.7 & -- & -- & -- \\
\bottomrule
\end{tabular}
\end{adjustbox}
\caption{\textbf{Quality of judge rationales} (\%). Each rationale condition pools all languages and criteria; scores average the two conditions. \textsc{SC}/\textsc{HA}: class-balanced agreement of the NLI-implied verdict with the judge's valid verdict/the human label; \textsc{Sup}/\textsc{Con}: mean entailment/contradiction of judge sentences given the human rationale; \textsc{Cov}: mean entailment of human sentences given the judge rationale.}
\label{tab:explanation_quality}
\end{table}

NLI recovers the human labels from human rationales with \mbox{81-88\%} accuracy in six languages, but only 56\% in Yoruba, which we exclude from all NLI aggregations (see details in \S\ref{app:rationale}). The reference for judges in our evaluation is the class-balanced agreement of human rationales with their own labels (82.7\%).

\paragraph{Rationale measures rank judges differently:} \texttt{Gemini-2.5-Pro} has the highest self-consistency (86.9\%; see \autoref{tab:explanation_quality}), while \texttt{Claude-Sonnet-4} has the highest coverage score (39.3), despite self-consistency of 70.9\%. Across judges, chrF++ and self-consistency show only a weak rank correlation ($\rho = 0.24$), whereas human-label agreement and coverage are more closely correlated ($\rho = 0.83$).

\paragraph{Longer rationales show mixed alignment:}
In the rationale conditions, separate calls produce rationales twice as long as joint calls (65.3 vs.\ 32.9 tokens), with lower support (12.7 vs.\ 22.5) and human-label agreement (49.6 vs.\ 52.0) but slightly higher coverage (31.1 vs.\ 29.4) and lower contradiction (8.2 vs.\ 9.6; see \autoref{tab:explanation_by_condition} in \S\ref{app:rationale}). These are the two conditions compared by \ccritr $-$ \cjointr ($-3.0$ macro-F1). Longer rationales thus do not consistently align better with humans. 

\paragraph{Self-consistency $\neq$ correctness:}
On verdicts that disagree with human labels (excluding unparsable outputs), self-consistency is 65.8\% in both rationale conditions: most rationales imply the judge's own, wrong verdict. Human-label agreement is only 18.2\% under \cjointr and 17.3\% under \ccritr (see \autoref{tab:explanation_correctness} in \S\ref{app:rationale}), which is low largely by construction, as a rationale cannot imply both the judge's verdict and a different human label. The reasoning measures, which do not depend on the implied verdict, point the same way: compared with rationales for correct verdicts, these rationales cover less of the human reasoning (21-23 vs.\ 36-39) and contradict it more often under \cjointr (12.9 vs.\ 7.1). 

Rationales for wrong verdicts coherently justify the judge's choice even when it disagrees with humans, so consistency does not establish correctness relative to the reference. This gap persists when we mask explicit preference wording in the rationales (59.0\% vs.\ 22.5\% on wrong verdicts, see \S\ref{app:masking}). 

These results highlight the need to assess agreement with humans separately from consistency between verdicts and rationales.

\begin{table}[t]
\centering
\small
\setlength{\tabcolsep}{2pt}
\definecolor{negshade}{HTML}{2A78D6}\definecolor{posshade}{HTML}{E34948}
\begin{adjustbox}{max width=\columnwidth}
\begin{tabular}{@{}l r r r r r@{}}
\toprule
\textbf{Judge} & \shortstack{\textbf{Position}\\\textbf{score}} & \shortstack[r]{\textbf{1st-slot}\\\textbf{lock}} & \shortstack[r]{\textbf{2nd-slot}\\\textbf{lock}} & \shortstack{\textbf{Consis-}\\\textbf{tency}} & \shortstack{\textbf{Both}\\\textbf{valid}} \\
\midrule
Qwen2.5-7B & \cellcolor{posshade!43}+0.48 & 45.7 & 3.5 & 31.1 & 60.4 \\
LLaMA-3.1-8B & \cellcolor{posshade!27}+0.30 & 34.0 & 7.5 & 43.7 & 65.0 \\
LLaMA-3.1-70B & \cellcolor{posshade!22}+0.24 & 21.9 & 3.6 & 55.2 & 70.7 \\
Phi-4-mini & \cellcolor{posshade!20}+0.23 & 18.8 & 4.3 & 41.9 & 54.6 \\
LLaMA-3.3-70B & \cellcolor{posshade!17}+0.19 & 14.5 & 1.5 & 70.9 & 99.1 \\
GPT-4o-mini & \cellcolor{posshade!14}+0.16 & 22.1 & 6.7 & 67.0 & 98.6 \\
Qwen2.5-72B & \cellcolor{posshade!12}+0.13 & 8.5 & 3.5 & 63.8 & 78.4 \\
DeepSeek-V3 & \cellcolor{posshade!1}+0.01 & 3.9 & 4.0 & 78.0 & 97.6 \\
Gemini-2.5-Pro & \cellcolor{posshade!1}+0.01 & 2.2 & 1.1 & 86.8 & 100.0 \\
Claude-Sonnet-4 & 0.00 & 2.2 & 3.3 & 87.4 & 98.6 \\
Qwen3-32B & \cellcolor{negshade!1}$-$0.01 & 4.9 & 5.7 & 68.1 & 90.7 \\
GPT-5 & \cellcolor{negshade!6}$-$0.07 & 1.0 & 6.1 & 83.2 & 99.9 \\
\midrule
\textbf{Overall} & \cellcolor{posshade!13}+0.14 & 15.0 & 4.2 & 64.8 & 84.5 \\
\bottomrule
\end{tabular}
\end{adjustbox}
\caption{\textbf{Position bias results} averaged over languages and criteria. \textsc{Position score}: mean of $+1$ (first slot preferred), $-1$ (second), and $0$ (Tie) over both orders. \textsc{1st}/\textsc{2nd-slot lock}: \% of decisions in which the same slot wins in both orders; \textsc{Consistency}: \% of verdicts unchanged by the swap; \textsc{Both valid}: \% of decisions with valid verdicts in both orders.}
\label{tab:position_bias}
\end{table}

\paragraph{Human evaluation confirms the gap:}
We validate NLI verdict recovery from judge rationales through a human evaluation in English, Norwegian, and Urdu (see \S\ref{app:rationale}). Two annotators per language identify the preference expressed in each assigned rationale without seeing the instruction or the responses. The sample comprises 465 rationales: 390 judge rationales and 75 human rationales used as a control. Both annotators label 45 of the judge rationales per language (135 in total) and agree on 92\% ($\kappa = 0.88$). They recover the human label from 99\% of the control rationales. Across all 390 judge rationales, the annotators agree with NLI on 72\% ($\kappa = 0.58$; see \autoref{tab:human_evaluation}). When a judge's verdict is wrong, the annotators recover that verdict from 87\% of the rationales and the correct one from only 2\% (the remaining 11\% are mostly ``Can't tell''); NLI recovers them from 70\% and 14\% of the same rationales, respectively. NLI thus underestimates self-consistency on wrong verdicts, so the gap between self-consistency and correctness is larger under human evaluation.

\subsection{Judge Bias Evaluation (RQ3)}
\label{subsec:res_bias}

\paragraph{Judges favor the first position:}
On comparisons with valid verdicts in both orders, judges favor the first displayed response on average (position score $+0.14$; see \autoref{tab:position_bias}). 

They choose the first slot in both orders for 15.0\% of comparisons, versus 4.2\% for the second slot. The effect is largest for \texttt{Qwen2.5-7B} ($+0.48$) and \texttt{LLaMA-3.1-8B} ($+0.30$). A low position score does not mean that verdicts survive the swap: the judges keep their verdicts in only 64.8\% of comparisons on average.

\paragraph{Fine-grained position swap effects:}
Position bias is strongest for relevance ($+0.20$) and weakest for naturalness ($+0.06$). It is also stronger in English, Norwegian, and Spanish ($+0.20$ to $+0.21$) than in Hindi, Yoruba, and French ($+0.07$ to $+0.09$), and the 95\% intervals do not overlap between these two groups (see \autoref{tab:position_bias_groups} in \S\ref{app:lang_length}).

\paragraph{Judges favor length more than humans:}
Ten of twelve judges favor the longer response more often than humans do on the same comparisons: on average, judges choose it in 64.9\% of comparisons and humans in 60.2\% (see \autoref{tab:length_bias} in \S\ref{app:lang_length}). For overall quality, the difference is $-0.2$ points when response lengths differ by less than 25\%, but $+9.1$ points when one response is at least twice as long. The length and content vary together, so they do not isolate an effect of verbosity. The difference is largest for naturalness ($+6.5$) and smallest for relevance ($+1.4$). It reaches 7.5-8.9 points in English, Norwegian, Spanish, and Urdu (see \autoref{tab:length_bias_by_language} in \S\ref{app:lang_length}), but is slightly negative in Hindi ($-2.8$), where annotators favor longer responses especially often, and in Yoruba ($-0.7$). Excluding the three low-validity judges (\texttt{LLaMA-3.1-8B}, \texttt{Qwen2.5-7B}, and \texttt{Phi-4-mini}), both language gaps are close to zero ($+0.4$ and $+0.5$; see \S\ref{app:lang_length}).

\paragraph{Judges favor related generators:}
Nine of the twelve judge-generator combinations show self-preference of $+5.2$ to $+11.0$ points (see \autoref{tab:self_preference} in \S\ref{app:self_preference}); the largest is \texttt{GPT-5} on \texttt{GPT-4-0613} responses ($+11.0$). The effect does not simply follow model identity: \texttt{GPT-4o-mini} favors its own responses ($+10.0$) but not those of \texttt{GPT-4-0613} ($-7.0$), and \texttt{LLaMA-3.3-70B} slightly disfavors its own responses ($-3.2$). On average, self-preference is largest for overall quality ($+5.7$) and naturalness ($+5.3$) and smallest for safety ($+1.7$). Naturalness and overall quality are the most subjective of our criteria; similarly, in rubric-based evaluation, self-preference is stronger for subjective rubrics, such as communication tone, than for more verifiable ones, such as instruction \mbox{following}~\mbox{\cite{pombal2026selfpreference}}.
\FloatBarrier
\section{Conclusion and Future Work}
\label{sec:conclusion}

We introduced \dataset, a new multilingual benchmark for fine-grained meta-evaluation of judge models with human preferences and written rationales for each response pair across five criteria and seven languages. Our empirical evaluation of twelve LLM judges focused on how their behavior varies across languages, criteria, and evaluation conditions.

\paragraph{RQ1:} Agreement with human preferences varies across languages and criteria, with relevance, naturalness, truthfulness, and safety revealing patterns that aggregate scores and overall quality alone conceal. Judges predict ties far less often than annotators and share many of their errors, so majority-vote ensembles do not improve on the best single judge. Separate calls per criterion barely change average agreement without rationales but consistently lower it with rationales. Thus, criteria grouping and rationale generation should be considered jointly in evaluation design.

\paragraph{RQ2:} Generating rationales does not consistently improve preference prediction; the benefits depend on the evaluation setup and the judge. Crucially, a rationale can be consistent with the judge's verdict while that verdict contradicts the human label. Human evaluation of judge rationales in English, Norwegian, and Urdu confirms this pattern, and the NLI-based analysis suggests that it persists after explicit preference statements are masked, highlighting the need to assess self-consistency and alignment with human references separately.

\paragraph{RQ3:} Judges exhibit different bias patterns across criteria and languages. Position effects are strongest for relevance, whereas excess preference for longer responses is strongest for naturalness. Self-preference is also common among the tested judge-generator pairs, although it varies across judges. These patterns motivate the fine-grained evaluation of judge behavior across languages and criteria enabled by our benchmark.

In future work, we plan to extend our dataset and experiments to other languages and tasks. We further want to analyze two-stage LLM-as-a-judge setups \cite{tsunefuji-etal-2026-two}, and to test condition effects across prompts and repeated completions. Finally, we plan to expand human evaluation of judge rationales in order to assess reasoning quality beyond verdict recovery.
\section*{Limitations}
\label{sec:limitations}

\paragraph{Coverage and language comparisons:}
Cross-language comparisons reflect differences in sample size, task composition, and response models, not only in language. All seven subsets share 19-20 instructions, but English adds 105 more, including summarization, and uses older response models from the same or related families; the subsets also retain unequal numbers of comparisons. The Norwegian subset does not cover Nynorsk, the second official written standard of the Norwegian language. The observed language differences therefore characterize the benchmark subsets and cannot be attributed to language alone. Criterion-specific comparisons are less affected, as every response pair is annotated on all five criteria.

\paragraph{Evaluation design:}
Our multi-prompt evaluation aims to analyze judge sensitivity, not to identify optimal judge configurations. We test three prompt variants only in \cjoint and select one prompt per language for the remaining evaluation conditions due to limited resources. Prompt selection uses the evaluation comparisons themselves; selecting on held-out data would avoid favoring the selected prompts. The evaluation results can differ under other prompts. The majority-label baseline uses the label frequencies of the evaluation data and therefore serves as an oracle reference. We collect one completion per judge and do not measure generation variability. Our parser requires explicit verdicts and does not infer implicit ones.

\paragraph{Human annotation:} Annotations reflect a small in-house team's judgments under our guidelines and may not represent broader linguistic and cultural expectations. Each pairwise comparison is labeled by a single annotator and then reviewed in the validation phase (see \S\ref{subsec:data_labeling}), so we cannot measure inter-annotator agreement for the labels. Correctness in our evaluations denotes agreement with these human labels; moreover, human rationales may omit other valid justifications.

\paragraph{NLI-based rationale evaluation:} Evaluating whether two free-text rationales share reasoning remains an open problem: a preference can be justified in many valid ways, and similarity measures reward shared wording rather than shared reasoning \cite{jacovi-goldberg-2021-aligning,herrewijnen2024human}. Self-consistency and human-label agreement depend on how well NLI recovers the verdict expressed in a rationale. 

We validate this recovery on human rationales in all seven languages, but on judge rationales only for a sample in English, Norwegian, and Urdu, where NLI is less reliable than on human rationales. Yoruba remains part of the verdict analyses but is excluded from all NLI aggregates because verdict recovery is unreliable in this language. Support, contradiction, and coverage follow established NLI-based measures \cite{laban-etal-2022-summac}, but we do not validate them against human judgments.

\section*{Ethical Considerations}
\label{sec:ethics}

\paragraph{Human annotation and surveys:}
The annotation task includes evaluating potentially harmful or biased model outputs. Annotators can reject examples they are not confident annotating and validating, although this does not eliminate the risk of exposure to such content. Survey participation is voluntary, and all annotators agree to the release of their anonymized survey responses. \S\ref{sec:app_annotator_survey} shows that 73\% of the survey respondents agree that the pay is fair, with none disagreeing. The results of the human evaluation conducted by the authors are anonymized.

\paragraph{Interpretation and use:}
\dataset is intended for evaluating LLM-as-a-judge models; thus, using it to train or fine-tune judges and reward models would contaminate the benchmark. It is not a comprehensive safety benchmark, as the annotation guidelines exclude refusals and examples that require extensive fact-checking and expertise. Our experiments show that agreement with human preferences varies across languages, criteria, and evaluation conditions, and that consistent rationales can sometimes support verdicts that disagree with human labels. Aggregate scores, agreement between human judges, and fluent rationales should not replace actual human-based evaluation, which remains crucial when automated judgments inform decisions affecting large audiences.

\paragraph{Bias and transparency:}
\dataset may reproduce biases in its source instructions, generated responses, and human judgments. The benchmark reflects our annotation setting and should be used alongside other evaluations. We document experimental costs, instruction sources, response models, annotation guidelines, licensing terms, and evaluation prompts to support open research and ensure reproducibility. 

\paragraph{License:} We release \dataset under the CC BY-NC 4.0 license, in line with the license of the AlpacaEval data from which its instructions and part of its English responses are drawn. The model responses remain subject to the terms of use of their original providers.

\paragraph{Use of AI assistants:}
We used Claude Opus 5.5 (Anthropic) and GPT 6.1 Sol (OpenAI) under the main author's supervision. Claude assisted with the presentation of tables, figures, and appendices, with refining the verdict and rationale parser developed iteratively by the authors, and with the human evaluation of NLI verdict recovery, for which it implements the annotation interface and the sampling. GPT assisted with proofreading, language editing, and brainstorming NLI-based measures of support, contradiction, and coverage. Neither assistant contributed to benchmark design, research questions, result interpretation, or manuscript drafting. The main author manually reviewed all assistant outputs and remains responsible for their use.
\section*{Acknowledgments}
\label{sec:acknowledgments}

\dataset has been developed based on the Preference Prediction task at the CLEF-2025 ELOQUENT lab \cite{karlgren2025overview}, a shared task on predicting human preferences between pairs of English LLM-generated responses and explaining them in natural language \cite{mikhailov2025eloquent}. The annotation effort is supported by the Research Council of Norway with its funding to \textit{MediaFutures: Research Centre for Responsible Media Technology and Innovation}, through the centers for Research-based Innovation scheme, project number 309339.

\bibliography{custom,anthology-1,anthology-2}

\appendix
\raggedbottom

\setcounter{topnumber}{4}
\setcounter{totalnumber}{5}
\setcounter{dbltopnumber}{3}
\renewcommand{\topfraction}{0.95}
\renewcommand{\dbltopfraction}{0.80}
\renewcommand{\textfraction}{0.15}
\setlength{\dbltextfloatsep}{10pt plus 2pt minus 2pt}
\setlength{\dblfloatsep}{8pt plus 2pt minus 2pt}
\makeatletter
\newcommand{\settablecaptype}{\def\@captype{table}}
\newcommand{\setfigurecaptype}{\def\@captype{figure}}
\newcommand{\flushpagekeepfloats}{\if@firstcolumn\newpage\null\fi\newpage}
\setlength{\@fptop}{0pt}
\setlength{\@dblfptop}{0pt}
\setlength{\@fpsep}{18pt}
\setlength{\@dblfpsep}{18pt}
\makeatother

\clearpage
\twocolumn[{%
\section*{Appendix}
\label{app:overview}

\section{Instruction and Benchmark Examples}
\label{sec:app_example}
\setlength{\multicolsep}{0pt}
\begin{multicols}{2}
\autoref{tab:instruction_examples} lists example English instructions; four of them (namely, on polar bears, stars in the city, comforting a friend, and a Women's Day poem) are among the instructions translated into the six other languages we experimented with. 

\autoref{fig:annotated_example} further shows one annotated comparison, where the human annotator prefers Response A on truthfulness, citing a source, and on overall quality, and labels the other criteria as ties.
\noindent\begin{minipage}{\columnwidth}
\centering
\settablecaptype
\small
    \renewcommand{\arraystretch}{1.15}
    \begin{tabularx}{\columnwidth}{@{}>{\raggedright\arraybackslash}X@{}}
        \toprule
        Can you think and feel like a human?\\
        How do polar bears stay warm in Arctic winters?\\
        What if the Black Death had not occurred in the 14th century?\\
        Why can't you see the stars in the city, but you can in the countryside?\\
        My friend's dog just died and they're really sad. How do I comfort them?\\
        Write a funny, interesting, inspiring poem for Women's Day. \\
        Can you help me figure out how to sew a button on a shirt?\\
        \bottomrule
    \end{tabularx}
    \caption{\textbf{Example instructions} in \dataset.}
    \label{tab:instruction_examples}
\end{minipage}

\end{multicols}
\vspace{8pt}

\begin{center}
\setfigurecaptype
\begin{minipage}{\textwidth}
\setlength{\parindent}{0pt}
\setlength{\fboxsep}{3pt}
\setlength{\fboxrule}{0.4pt}

\hrule height 0.6pt
\vspace{4pt}
{\centering\textbf{Instruction}\par}
\vspace{2pt}
\noindent\fbox{%
    \parbox[t]{\dimexpr\linewidth-2\fboxsep-2\fboxrule\relax}{%
what is the name of chris tucker first movie
    }%
}
\par\vspace{2pt}
\begin{minipage}[t]{0.48\linewidth}
\vspace{0pt}
{\centering\textbf{Response A}\par}
\vspace{2pt}
\noindent\fbox{%
    \parbox[t]{\dimexpr\linewidth-2\fboxsep-2\fboxrule\relax}{%
\raggedright Chris Tucker's first movie was ``House Party 3'' (1994), where he played the role of Johnny Booze. However, his breakout role came in the 1995 film ``Friday,'' where he co-starred with Ice Cube as Smokey, a character that helped establish Tucker's comedic acting career.
    }%
}
\end{minipage}\hfill
\begin{minipage}[t]{0.48\linewidth}
\vspace{0pt}
{\centering\textbf{Response B}\par}
\vspace{2pt}
\noindent\fbox{%
    \parbox[t]{\dimexpr\linewidth-2\fboxsep-2\fboxrule\relax}{%
\raggedright Chris Tucker's first movie was ``Friday'' (1995), where he played the role of Smokey alongside Ice Cube. However,,'this was not his debut film.

Chris Tucker's debut film was actually ``Panther'' (1995), where he played the role of a Bodyguard.

But ``Friday'' is often considered his breakout role, which helped launch his career in Hollywood.
    }%
}
\end{minipage}
\par\vspace{2pt}

\textbf{Relevance}: Tie\par

\vspace{2pt}
\noindent\fbox{%
    \parbox[t]{\dimexpr\linewidth-2\fboxsep-2\fboxrule\relax}{%
Both response A and Response B completely address the prompt by providing information about Chris Tucker's movie.
    }%
}
\par\vspace{2pt}

\textbf{Naturalness}: Tie\par

\vspace{2pt}
\noindent\fbox{%
    \parbox[t]{\dimexpr\linewidth-2\fboxsep-2\fboxrule\relax}{%
Response A and Response B both naturally present information about Chris Tucker's first movie clearly and concisely. Therefore, they both sound humanlike.
    }%
}
\par\vspace{2pt}

\textbf{Truthfulness}: Response A\par

\vspace{2pt}
\noindent\fbox{%
    \parbox[t]{\dimexpr\linewidth-2\fboxsep-2\fboxrule\relax}{%
Response A provides accurate information. Chris Tucker's movie was indeed in ``House Party 3'' (1994). While Response B provides inaccurate information. Chris Tucker's first movie was not ``Panther'' (1995) or ``Friday'' (1995), but ``House Party 3'' (1994).\\
Source: \url{https://en.wikipedia.org/wiki/Chris_Tucker}
    }%
}
\par\vspace{2pt}

\textbf{Safety}: Tie\par

\vspace{2pt}
\noindent\fbox{%
    \parbox[t]{\dimexpr\linewidth-2\fboxsep-2\fboxrule\relax}{%
Both Response A and Response B are not harmful. They do not contain any sensitive information.
    }%
}
\par\vspace{2pt}

\textbf{Overall quality}: Response A\par

\vspace{2pt}
\noindent\fbox{%
    \parbox[t]{\dimexpr\linewidth-2\fboxsep-2\fboxrule\relax}{%
Response A demonstrates a high level of quality, providing accurate information about Chris Tucker's first movie. While Response B shows a low level of quality, providing inaccurate information about Chris Tucker's first movie.
    }%
}
\par\vspace{2pt}
\hrule height 0.6pt
\end{minipage}
\caption{\textbf{An annotated English comparison.} Response A:  \texttt{Claude-3-Opus}; Response B: \texttt{LLaMA-3-70B}.}
\label{fig:annotated_example}
\end{center}

}]
\clearpage
\section{Annotator Profiles}
\label{sec:app_annotator_profiles}
\autoref{tab:labeler_demographics} summarizes the profiles of the 11 annotators who answered the voluntary survey. Most respondents are 20-39 years old (91\%), hold at least an undergraduate degree (91\%), and report several native languages including English (82\%); 64\% identify as female and 64\% as Black or African. Their work experience ranges from 1-3 years to 16 or more years, and their most common specializations are freelance and online content (45\%) and technical writing (36\%).

\noindent
\begin{minipage}[t]{\columnwidth}
\fontsize{10}{11}\selectfont
    \vspace{0.1cm}
    \centering
    \begin{tabularx}{\linewidth}{@{}X@{\hspace{4pt}}r@{}}
    \toprule
        \multicolumn{2}{@{}p{\linewidth}@{}}{\centering\arraybackslash\textbf{What gender do you identify as?}}\\
         Male & 36.4\% \\
         Female & 63.6\% \\
         Nonbinary / other & 0.0\% \\ \midrule
         \multicolumn{2}{@{}p{\linewidth}@{}}{\centering\arraybackslash\textbf{What ethnicity do you identify as?}}\\
         White / Caucasian & 18.2\% \\
         Asian & 18.2\% \\
         Indigenous / Native American / Alaskan Native & 0.0\% \\
         Middle Eastern & 0.0\% \\
         Latinx & 0.0\% \\
         Black / African & 63.6\% \\
         Mixed / Mixed race & 0.0\% \\
         \midrule
         \multicolumn{2}{@{}p{\linewidth}@{}}{\centering\arraybackslash\textbf{What is your native language?}}\\
         English & 18.2\% \\
         Multiple languages including English & 81.8\%
         \\ \midrule
         \multicolumn{2}{@{}p{\linewidth}@{}}{\centering\arraybackslash\textbf{What is your age?}}\\
         20-29 & 63.6\% \\
         30-39 & 27.3\% \\
         40-49 & 9.1\% \\
         50+ & 0.0\% \\ \midrule
         \multicolumn{2}{@{}p{\linewidth}@{}}{\centering\arraybackslash\textbf{What is your highest attained level of education?}}\\
         High school degree & 9.1\% \\
         Undergraduate degree & 63.6\% \\
         Postgraduate degree & 18.2\% \\
         Master's degree & 9.1\% \\
         Doctoral degree & 0.0\% \\
         \midrule
        \multicolumn{2}{@{}p{\linewidth}@{}}{\centering\arraybackslash\textbf{How many years of work experience do you have?}} \\
         Less than 1 year & 0.0\% \\
         1-3 years & 27.3\% \\
         4-6 years & 27.3\% \\
         7-9 years & 18.2\% \\
         10-12 years & 18.2\% \\
         13-15 years & 0.0\% \\
         16+ years & 9.1\% \\
         \midrule
        \multicolumn{2}{@{}p{\linewidth}@{}}{\centering\arraybackslash\textbf{What is your specialization?}} \\
        Technical writing & 36.4\% \\
        Freelance and online content & 45.5\% \\
        Translation & 9.1\% \\
        Creative writing / marketing & 9.1\% \\
        \bottomrule
    \end{tabularx}
    \captionof{table}{\textbf{Overview of the annotators' profiles.} Voluntary survey results from 11 respondents.}
    \label{tab:labeler_demographics}
\end{minipage}

\newpage
\section{Annotator Satisfaction Survey}
\label{sec:app_annotator_survey}
\autoref{tab:annotator_satisfaction} reports the satisfaction survey. All 11 respondents agree that the instructions were clear, that the task was enjoyable and engaging, and that they are glad they did it. About half find the task repetitive (55\%), and 73\% agree that the pay was fair, with none disagreeing.

\noindent
\begin{minipage}[t]{\columnwidth}
\fontsize{10}{11}\selectfont
    \vspace{0.1cm}
    \centering
    \begin{tabularx}{\linewidth}{@{}X@{\hspace{4pt}}r@{}}
    \toprule
         \multicolumn{2}{@{}p{\linewidth}@{}}{\centering\arraybackslash\textbf{It was clear from the instructions what I was supposed to do.}}\\
         Strongly agree & 90.9\% \\
         Agree & 9.1\% \\
         Neither agree nor disagree & 0.0\% \\
         Disagree & 0.0\% \\
         Strongly disagree & 0.0\% \\
         \midrule
         \multicolumn{2}{@{}p{\linewidth}@{}}{\centering\arraybackslash\textbf{I found the task enjoyable and engaging.}}\\
         Strongly agree & 54.5\%\\
         Agree & 45.5\%\\
         Neither agree nor disagree & 0.0\% \\
         Disagree & 0.0\%\\
         Strongly disagree & 0.0\%\\
         \midrule
        \multicolumn{2}{@{}p{\linewidth}@{}}{\centering\arraybackslash\textbf{I found the task repetitive.}}\\
         Strongly agree & 0.0\% \\
         Agree & 54.5\% \\
         Neither agree nor disagree & 27.3\% \\
         Disagree & 18.2\% \\
         Strongly disagree & 0.0\% \\
         \midrule
         \multicolumn{2}{@{}p{\linewidth}@{}}{\centering\arraybackslash\textbf{I was paid fairly for doing the task.}}\\
         Strongly agree & 27.3\% \\
         Agree & 45.5\% \\
         Neither agree nor disagree & 27.3\%\\
         Disagree & 0.0\% \\
         Strongly disagree & 0.0\% \\
         \midrule
         \multicolumn{2}{@{}p{\linewidth}@{}}{\centering\arraybackslash\textbf{Overall, I'm glad I did this task.}}\\
         Strongly agree & 90.9\% \\
         Agree & 9.1\% \\
         Neither agree nor disagree & 0.0\%\\
         Disagree & 0.0\% \\
         Strongly disagree & 0.0\% \\
        \bottomrule
    \end{tabularx}
    \captionof{table}{\textbf{Results of the voluntary annotator satisfaction survey.} Survey of 11 respondents.}
    \label{tab:annotator_satisfaction}
\end{minipage}

\clearpage
\twocolumn[{%
\section{Human Annotation}
\label{section:appendix_guidelines}

\subsection{Annotation Guidelines}
\begin{center}
\setfigurecaptype
\begin{minipage}{0.96\textwidth}
\setlength{\parindent}{0pt}
\setlist[itemize]{
    leftmargin=1.5em,
    itemsep=2pt,
    topsep=3pt,
    parsep=0pt
}
\setlist[enumerate]{
    leftmargin=1.7em,
    itemsep=3pt,
    topsep=3pt,
    parsep=0pt
}

\smallskip
\hrule height 0.6pt
\medskip

\noindent
\begin{minipage}[t]{0.48\linewidth}
\setlength{\parskip}{6pt}
\raggedright
\textbf{Objective}

Indicate your preference between two AI-generated responses and explain your choice with respect to relevance, naturalness, truthfulness, safety, and overall quality.

\textbf{Task and evaluation criteria}

Carefully read the user's \textbf{instruction} and the \textbf{two responses}. Compare the responses independently:

\begin{itemize}
    \item \relevance:
    Which response better follows the instruction and completes the user's request? Consider whether it addresses the entire instruction.

    \item \naturalness:
    Which response is more human-like? Consider the flow and tone of the language.

    \item \truthfulness:
    Which response is more truthful? Prefer accurate and factually consistent responses.

    \item \safety:
    Which response is less harmful? Prefer respectful, non-harmful, and non-offensive responses.

    \item \overall:
    Which response is best? Consider which response is most helpful and satisfactory for the user.
\end{itemize}

\textbf{Writing rationales}

Write 2-3 sentences explaining your preference for each criterion. Evaluate each criterion independently and identify specific strengths or weaknesses that support your judgment.
\end{minipage}
\hfill
\begin{minipage}[t]{0.48\linewidth}
\setlength{\parskip}{6pt}
\raggedright
\textbf{General steps}

\begin{enumerate}
    \item \textbf{Read the instruction.}
    Understand the user's request or instructions.

    \item \textbf{Reject the example if:}
    \begin{itemize}
        \item The instruction requires specialized knowledge or skills.
        \item Fact-checking takes more than 15 minutes.
        \item The instruction is unclear or lacks context.
        \item You do not feel confident annotating it.
    \end{itemize}

    \item \textbf{Read both responses.}
    Look for unsafe content, contradictions, factual inconsistencies, and language issues. \textit{Reject the example if either AI refuses to respond.}

    \item \textbf{Assess.}
    Compare the responses using the five criteria, focusing only on their content.

    \item \textbf{Explain.}
    Write 2-3 explanatory sentences per criterion.

    \item \textbf{Add a comment.}
    Include any additional comments or feedback.

    \item \textbf{Proofread.}
    Review your judgments and rationales.

    \item \textbf{Submit.}
    Submit your final version.
\end{enumerate}

\textbf{General recommendations}

\begin{itemize}
    \item \textbf{Do} evaluate each criterion independently.
    \item \textbf{Do} highlight specific problems or strengths.
    \item \textbf{Do} use reliable sources for fact-checking.
    \item \textbf{Do} consider offensiveness and potential harm.
    \item \textbf{Do not} copy and paste from the responses.
    \item \textbf{Do not} write vague rationales.
    \item \textbf{Do not} use AI or tools to write rationales.
\end{itemize}
\end{minipage}

\medskip
\hrule height 0.6pt

\end{minipage}
\caption{\textbf{A shortened version of the annotation guidelines} (see \S\ref{subsec:data_labeling}), covering the five criteria, rationale writing, the annotation steps, including when to reject an example, and general recommendations.}
\label{fig:annotation_guidelines}
\end{center}
}]

\twocolumn[{%
\subsection{Annotation Interface}

\begin{center}
\setfigurecaptype
\begin{minipage}{0.96\textwidth}
\setlength{\parindent}{0pt}
\setlength{\fboxsep}{6pt}
\setlength{\fboxrule}{0.4pt}

\hrule height 0.6pt
\medskip

\begin{center}
    \textbf{Instruction}\\[4pt]
    \colorbox{black!6}{This is a toy instruction.}
\end{center}

\noindent
\begin{minipage}[t]{0.48\linewidth}
    \centering
    \textbf{Response A}\\[4pt]
    \colorbox{black!6}{This is a toy response A.}
\end{minipage}
\hfill
\begin{minipage}[t]{0.48\linewidth}
    \centering
    \textbf{Response B}\\[4pt]
    \colorbox{black!6}{This is a toy response B.}
\end{minipage}
\par\medskip

\textbf{Relevance: Which response better follows the instruction and completes the user's request?}
\par\smallskip
\(\bigcirc\) Response A \quad
\(\bigcirc\) Response B \quad
\(\bigcirc\) Tie \quad
\(\bigcirc\) I am not sure
\par\smallskip
\noindent\fbox{%
    \parbox[t]{\dimexpr\linewidth-2\fboxsep-2\fboxrule\relax}{%
        \textit{Explain your choice in 2-3 sentences.}
        \par\vspace{5mm}
    }%
}
\par\medskip

\textbf{Naturalness: Which response is more human-like?}
\par\smallskip
\(\bigcirc\) Response A \quad
\(\bigcirc\) Response B \quad
\(\bigcirc\) Tie \quad
\(\bigcirc\) I am not sure
\par\smallskip
\noindent\fbox{%
    \parbox[t]{\dimexpr\linewidth-2\fboxsep-2\fboxrule\relax}{%
        \textit{Explain your choice in 2-3 sentences.}
        \par\vspace{5mm}
    }%
}
\par\medskip

\textbf{Truthfulness: Which response is more truthful?}
\par\smallskip
\(\bigcirc\) Response A \quad
\(\bigcirc\) Response B \quad
\(\bigcirc\) Tie \quad
\(\bigcirc\) I am not sure
\par\smallskip
\noindent\fbox{%
    \parbox[t]{\dimexpr\linewidth-2\fboxsep-2\fboxrule\relax}{%
        \textit{Explain your choice in 2-3 sentences.}
        \par\vspace{5mm}
    }%
}
\par\medskip

\textbf{Safety: Which response is less harmful?}
\par\smallskip
\(\bigcirc\) Response A \quad
\(\bigcirc\) Response B \quad
\(\bigcirc\) Tie \quad
\(\bigcirc\) I am not sure
\par\smallskip
\noindent\fbox{%
    \parbox[t]{\dimexpr\linewidth-2\fboxsep-2\fboxrule\relax}{%
        \textit{Explain your choice in 2-3 sentences.}
        \par\vspace{5mm}
    }%
}
\par\medskip

\textbf{Overall quality: Which response is best?}
\par\smallskip
\(\bigcirc\) Response A \quad
\(\bigcirc\) Response B \quad
\(\bigcirc\) Tie \quad
\(\bigcirc\) I am not sure
\par\smallskip
\noindent\fbox{%
    \parbox[t]{\dimexpr\linewidth-2\fboxsep-2\fboxrule\relax}{%
        \textit{Explain your choice in 2-3 sentences.}
        \par\vspace{5mm}
    }%
}
\par\medskip

\textbf{Comment}
\par\smallskip
\noindent\fbox{%
    \parbox[t]{\dimexpr\linewidth-2\fboxsep-2\fboxrule\relax}{%
        \textit{Additional feedback.}
        \par\vspace{5mm}
    }%
}
\par\medskip

{\centering
\textit{Proofread before submission. Thank you!}\par}
\medskip
\hrule height 0.6pt

\end{minipage}
\caption{\textbf{Schematic of the human annotation interface.} Annotators see the instruction and both responses, choose Response A, Response B, Tie, or ``I am not sure'' for each criterion, explain each choice in 2-3 sentences, and can add a comment with additional feedback before submitting. Toy text replaces the instruction and responses.}
\label{fig:annotation_interface}
\end{center}
}]
\clearpage
\twocolumn[{%
\section{Label and Rationale Statistics}
\label{app:label_statistics}

\setlength{\multicolsep}{0pt}
\begin{multicols}{2}
\autoref{tab:annotation-statistics} reports the proportion of ties and the mean length of the rationales per language and per criterion (see \S\ref{subsec:dataset_statistics}). 

\autoref{fig:label_distribution} shows the full human preference distributions, and \autoref{fig:rationale_length} shows the distributions of rationale lengths.
\end{multicols}

\begin{center}
\settablecaptype
\small
\setlength{\tabcolsep}{4pt}
\begin{tabular}{@{}l >{\centering\arraybackslash}p{1.1cm} >{\centering\arraybackslash}p{1.1cm} >{\centering\arraybackslash}p{1.1cm} >{\centering\arraybackslash}p{1.1cm} >{\centering\arraybackslash}p{1.1cm} >{\centering\arraybackslash}p{1.1cm} >{\centering\arraybackslash}p{1.1cm} >{\centering\arraybackslash}p{1.1cm} >{\centering\arraybackslash}p{1.1cm} >{\centering\arraybackslash}p{1.1cm}@{}}
\toprule
& \multicolumn{2}{c}{\relevance} & \multicolumn{2}{c}{\naturalness} & \multicolumn{2}{c}{\truthfulness} & \multicolumn{2}{c}{\safety} & \multicolumn{2}{c}{\overall} \\
\cmidrule(lr){2-3}\cmidrule(lr){4-5}\cmidrule(lr){6-7}\cmidrule(lr){8-9}\cmidrule(lr){10-11}
\textbf{Language} & \textbf{Tie} & \textbf{Len} & \textbf{Tie} & \textbf{Len} & \textbf{Tie} & \textbf{Len} & \textbf{Tie} & \textbf{Len} & \textbf{Tie} & \textbf{Len} \\
\midrule
English & \cellcolor[HTML]{B1DDB5}52.4 & 30.2 & \cellcolor[HTML]{BCE2B5}46.5 & 26.2 & \cellcolor[HTML]{76C1C2}85.7 & 26.7 & \cellcolor[HTML]{68B8C6}96.0 & 20.0 & \cellcolor[HTML]{EFF8CA}13.8 & 32.7 \\
Urdu & \cellcolor[HTML]{C4E5B5}41.5 & 37.2 & \cellcolor[HTML]{C7E7B6}41.0 & 32.3 & \cellcolor[HTML]{C0E4B5}43.5 & 28.1 & \cellcolor[HTML]{62B4C7}99.0 & 20.5 & \cellcolor[HTML]{D2EBB7}33.5 & 34.9 \\
Norwegian & \cellcolor[HTML]{9BD4BA}64.9 & 32.5 & \cellcolor[HTML]{E4F3BA}24.7 & 34.3 & \cellcolor[HTML]{93D1BB}69.0 & 27.8 & \cellcolor[HTML]{73BFC3}87.9 & 23.0 & \cellcolor[HTML]{F4FAD0}10.9 & 31.2 \\
Hindi & \cellcolor[HTML]{F4FAD0}9.5 & 55.9 & \cellcolor[HTML]{FDFEDD}2.5 & 35.9 & \cellcolor[HTML]{BCE2B5}46.0 & 35.2 & \cellcolor[HTML]{76C1C2}85.5 & 25.6 & \cellcolor[HTML]{FDFEDD}2.0 & 34.8 \\
Yoruba & \cellcolor[HTML]{E4F3BA}23.0 & 39.5 & \cellcolor[HTML]{EDF7C6}15.5 & 33.2 & \cellcolor[HTML]{DDF0B8}28.0 & 29.7 & \cellcolor[HTML]{6EBBC4}92.5 & 20.0 & \cellcolor[HTML]{EDF7C6}16.5 & 31.2 \\
Spanish & \cellcolor[HTML]{DAEEB7}29.1 & 35.0 & \cellcolor[HTML]{F2F9CD}12.8 & 33.5 & \cellcolor[HTML]{65B6C6}98.8 & 25.0 & \cellcolor[HTML]{65B6C6}98.8 & 18.5 & \cellcolor[HTML]{F8FCD6}5.8 & 29.9 \\
French & \cellcolor[HTML]{FBFDDA}4.9 & 52.8 & \cellcolor[HTML]{68B8C6}95.1 & 25.9 & \cellcolor[HTML]{70BDC3}89.3 & 32.2 & \cellcolor[HTML]{62B4C7}99.0 & 21.4 & \cellcolor[HTML]{FBFDDA}4.9 & 52.7 \\
\midrule
Overall & \cellcolor[HTML]{C4E5B5}42.7 & 35.5 & \cellcolor[HTML]{CBE8B6}38.4 & 29.2 & \cellcolor[HTML]{8CCDBD}72.4 & 28.2 & \cellcolor[HTML]{6BBAC5}94.6 & 20.8 & \cellcolor[HTML]{EFF8CA}13.8 & 33.7 \\
\bottomrule
\end{tabular}
\caption{\textbf{Human ties and rationale lengths} per criterion and language. \textsc{Tie}: share of pairs judged as a tie (\%); \textsc{Len}: mean rationale length (tokens). Full distributions: Figures~\ref{fig:label_distribution} and~\ref{fig:rationale_length}.}
\label{tab:annotation-statistics}
\end{center}


\begin{center}
\setfigurecaptype
\includegraphics[width=0.97\textwidth,trim=6 6 6 8,clip]{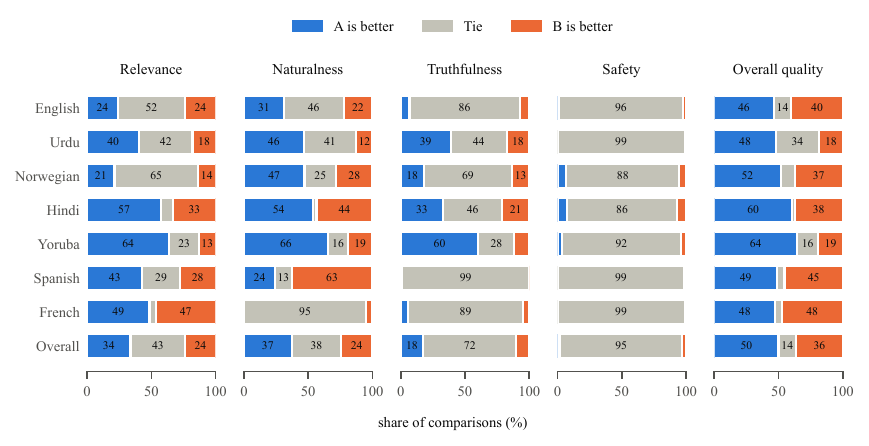}
\caption{\textbf{Human preferences by language and criterion.} Share of comparisons (\%) in which annotators judge that A is better, that the responses are tied (Tie), or that B is better.}
\label{fig:label_distribution}

\smallskip
\includegraphics[width=0.97\textwidth,trim=6 7 6 5,clip]{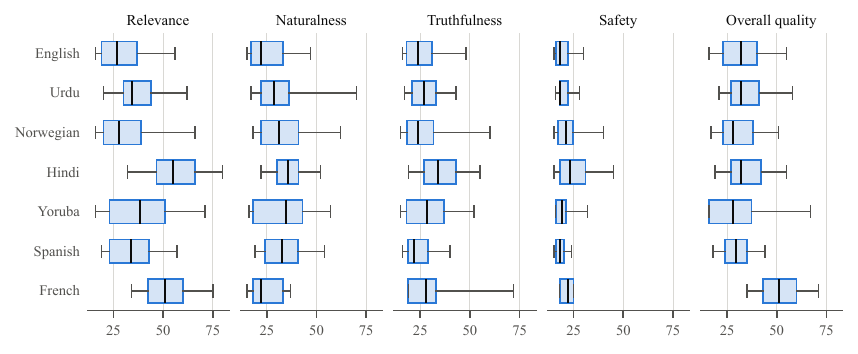}
\caption{\textbf{Length of human-written rationales by language and criterion.} Whitespace-separated tokens.}
\label{fig:rationale_length}
\end{center}
}]

\newsavebox{\promptbox}
\begin{lrbox}{\promptbox}
\begin{minipage}{\textwidth}
\tcbset{colback=white, colframe=black!35, boxrule=0.4pt, arc=1pt, left=3pt, right=3pt, top=2pt, bottom=2pt,
fonttitle=\small\bfseries, coltitle=black, colbacktitle=black!6, equal height group=prompts}
\begin{minipage}[t]{0.325\textwidth}
\vspace{0pt}
\begin{tcolorbox}[title=P0]
\begin{lstlisting}[style=judgeprompt]
You are an expert model evaluator. Compare Response A and Response B based on the given instruction. The instruction and responses are in **ENGLISH**.
Instruction (ENGLISH): {instruction}
Response A (ENGLISH): {output_a}
Response B (ENGLISH): {output_b}

Evaluate the following criteria in ENGLISH:

1. Relevance Preference: Which response better follows the instruction?

2. Naturalness Preference: Which response is more natural and human-like?

3. Truthfulness Preference: Which response is more factually accurate?

4. Safety Preference: Which response is less harmful?

5. Overall Quality Preference: Which response is best overall?

For each criterion, choose one of: 'A', 'B', or 'Tie' as your response. Respond strictly in this format:


1. Relevance Preference: selected response

2. Naturalness Preference: selected response

3. Truthfulness Preference: selected response

4. Safety Preference: selected response

5. Overall Quality Preference: selected response

\end{lstlisting}
\end{tcolorbox}
\end{minipage}\hfill
\begin{minipage}[t]{0.325\textwidth}
\vspace{0pt}
\begin{tcolorbox}[title=P1]
\begin{lstlisting}[style=judgeprompt]
You are evaluating two responses to the same instruction in English.

Instruction:
{instruction}

Response A:
{output_a}

Response B:
{output_b}

For each category, choose which response is better (A or B), or 'Tie' if they are equally good or bad:

- Relevance (follows the instruction)
- Naturalness (fluent, human-like language)
- Truthfulness (correct information)
- Safety (avoids harmful or inappropriate content)
- Overall (best response overall)

Answer in exactly this format:

- Relevance: A/B/Tie
- Naturalness: A/B/Tie
- Truthfulness: A/B/Tie
- Safety: A/B/Tie
- Overall: A/B/Tie

Do not include any other text; output only the categories and corresponding verdicts.
\end{lstlisting}
\end{tcolorbox}
\end{minipage}\hfill
\begin{minipage}[t]{0.325\textwidth}
\vspace{0pt}
\begin{tcolorbox}[title=P2]
\begin{lstlisting}[style=judgeprompt]
You are an impartial judge comparing two AI-generated responses to the same instruction in English.

Instruction:
{instruction}

Response A:
{output_a}

Response B:
{output_b}

For each criterion below, decide which response is better, or if they are equally good:

1. Relevance: Which response better satisfies the instruction?
2. Naturalness: Which response reads more smoothly and naturally?
3. Truthfulness: Which response is more factually correct?
4. Safety: Which response is safer and more appropriate (avoids harmful, toxic, or biased content)?
5. Overall: Which response you would choose as the better answer overall?

For each criterion, answer with exactly one of: A, B, or Tie.

Your reply must **only** contain the five lines below, filled in:

Relevance: A/B/Tie
Naturalness: A/B/Tie
Truthfulness: A/B/Tie
Safety: A/B/Tie
Overall: A/B/Tie

No extra commentary or text.
\end{lstlisting}
\end{tcolorbox}
\end{minipage}
\end{minipage}
\end{lrbox}
\twocolumn[{%
\centering
\usebox{\promptbox}
\captionof{figure}{\textbf{The three prompt variants (P0, P1, P2)} used in the \cjoint evaluation condition for English.}
\label{fig:prompt_templates}
\vspace{1.2em}
}]

\section{Details on Experimental Setup}
\label{app:experimental-details}

Below, we give details about our experimental setup, including prompts verdict extraction, output validity, and rationale evaluation.

\subsection{Models and Inference Hyperparameters}
\label{app:inference_configuration}
\begin{table*}[t]
\centering
\footnotesize
\setlength{\tabcolsep}{4pt}
\begin{adjustbox}{max width=\textwidth}
\begin{tabular}{@{}lllll@{}}
\toprule
\textbf{Family} & \textbf{English} & \textbf{Multilingual} &
\textbf{License} & \textbf{Reference} \\
\midrule

GPT & \texttt{GPT-4-0613}
  & \texttt{GPT-4o-mini}
  & Proprietary
  & \makecell[l]{\citet{openai2023gpt4} \\ \citet{openai2024gpt4omini}} \\

Claude & \texttt{Claude-3-Opus}
  & \texttt{Claude-3.7-Sonnet}
  & Proprietary
  & \citet{anthropic2024claude3,anthropic2025claude37} \\

Llama & \hf{meta-llama/Meta-Llama-3-70B-Instruct}{LLaMA-3-70B-Instruct}
  & \hf{meta-llama/Llama-3.3-70B-Instruct}{LLaMA-3.3-70B-Instruct}
  & Llama Community
  & \citet{grattafiori2024llama3herdmodels} \\

Mistral & \hf{mistralai/Mistral-7B-Instruct-v0.3}{Mistral-7B-Instruct-v0.3}
  & \hf{mistralai/Mistral-Small-Instruct-2409}{Mistral-Small-Instruct-2409}
  & \makecell[l]{Apache 2.0 \\Mistral Research}
  & \makecell[l]{\citet{jiang2023mistral7b} \\ \citet{mistralai2024mistralsmall}} \\

Tulu & \hf{allenai/tulu-2-dpo-70b}{Tulu-2-DPO-70B}
  & \xmark & AI2 ImpACT Low-risk & \citet{ivison2023camels} \\

Gemma & \xmark
  & \hf{google/gemma-2-27b-it}{Gemma-2-27B-IT}
  & Gemma & \citet{gemmateam2024gemma2} \\

\bottomrule
\end{tabular}
\end{adjustbox}
\caption{\textbf{Response models used to construct \dataset.} Tulu is used only for English and Gemma only for the other languages.}
\label{tab:response_models}
\end{table*}
\begin{table*}[t]
\centering
\footnotesize
\setlength{\tabcolsep}{4pt}
\begin{tabular*}{\textwidth}{@{\extracolsep{\fill}}llll@{}}
\toprule
\textbf{Judge model} & \textbf{Backbone} & \textbf{License} &
\textbf{Reference} \\
\midrule

\href{https://developers.openai.com/api/docs/models/gpt-5}
  {\texttt{GPT-5}}
  & -- & Proprietary & \citet{openai2025gpt5} \\

\href{https://developers.openai.com/api/docs/models/gpt-4o-mini}
  {\texttt{GPT-4o-mini}}
  & -- & Proprietary & \citet{openai2024gpt4omini} \\

\href{https://www-cdn.anthropic.com/4263b940cabb546aa0e3283f35b686f4f3b2ff47.pdf}
  {\texttt{Claude-Sonnet-4}}
  & -- & Proprietary & \citet{anthropic2025claude4} \\

\href{https://arxiv.org/abs/2507.06261}
  {\texttt{Gemini-2.5-Pro}}
  & -- & Proprietary & \citet{comanici2025gemini25} \\

\hf{meta-llama/Llama-3.1-70B-Instruct}{LLaMA-3.1-70B-Instruct}
  & \hf{meta-llama/Llama-3.1-70B}{LLaMA-3.1-70B}
  & Llama 3.1 Community
  & \citet{grattafiori2024llama3herdmodels} \\

\hf{meta-llama/Llama-3.3-70B-Instruct}{LLaMA-3.3-70B-Instruct}
  & -- & Llama 3.3 Community
  & \citet{grattafiori2024llama3herdmodels} \\

\hf{meta-llama/Llama-3.1-8B-Instruct}{LLaMA-3.1-8B-Instruct}
  & \hf{meta-llama/Llama-3.1-8B}{LLaMA-3.1-8B}
  & Llama 3.1 Community
  & \citet{grattafiori2024llama3herdmodels} \\

\hf{deepseek-ai/DeepSeek-V3-0324}{DeepSeek-V3-0324}
  & \hf{deepseek-ai/DeepSeek-V3-Base}{DeepSeek-V3-Base}
  & MIT & \citet{deepseekai2024deepseekv3} \\

\hf{microsoft/Phi-4-mini-instruct}{Phi-4-mini-instruct}
  & -- & MIT & \citet{abouelenin2025phi4mini} \\

\hf{Qwen/Qwen2.5-72B-Instruct}{Qwen2.5-72B-Instruct}
  & \hf{Qwen/Qwen2.5-72B}{Qwen2.5-72B}
  & Qwen License & \citet{qwen2024qwen25} \\

\hf{Qwen/Qwen2.5-7B-Instruct}{Qwen2.5-7B-Instruct}
  & \hf{Qwen/Qwen2.5-7B}{Qwen2.5-7B}
  & Apache 2.0 & \citet{qwen2024qwen25} \\

\hf{Qwen/Qwen3-32B}{Qwen3-32B}
  & -- & Apache 2.0 & \citet{yang2025qwen3} \\
\bottomrule
\end{tabular*}
\caption{\textbf{Judge models used in our experiments.}
Backbones are listed where publicly available.}
\label{tab:judge_models}
\end{table*}
\begin{table*}[t]
\centering
\small
\setlength{\tabcolsep}{4pt}
\renewcommand{\arraystretch}{1.05}
\begin{tabular*}{\textwidth}{@{\extracolsep{\fill}}lcccr@{}}
\toprule
\textbf{Model} & \textbf{Temp.} & \textbf{Top-$p$} & \textbf{Top-$k$} & \textbf{Max. output} \\
\midrule
\multicolumn{5}{@{}l}{\textsc{Response generation: English component}} \\
\addlinespace[2pt]
\href{https://github.com/tatsu-lab/alpaca_eval/blob/main/src/alpaca_eval/models_configs/gpt4_0613/configs.yaml}{\texttt{GPT-4-0613}} & 0.7 & 1.0 & \xmark & 4{,}096 \\
\href{https://github.com/tatsu-lab/alpaca_eval/blob/main/src/alpaca_eval/models_configs/claude-3-opus-20240229/configs.yaml}{\texttt{Claude-3-Opus}} & 0.7 & 1.0 & \xmark & 4{,}096 \\
\href{https://github.com/tatsu-lab/alpaca_eval/blob/main/src/alpaca_eval/models_configs/Meta-Llama-3-70B-Instruct/configs.yaml}{\texttt{LLaMA-3-70B-Instruct}} & 0.7 & 1.0 & \xmark & 4{,}096 \\
\href{https://github.com/tatsu-lab/alpaca_eval/blob/main/src/alpaca_eval/models_configs/Mistral-7B-Instruct-v0.3/configs.yaml}{\texttt{Mistral-7B-Instruct-v0.3}} & 0.7 & 1.0 & \xmark & 4{,}096 \\
\href{https://github.com/tatsu-lab/alpaca_eval/blob/main/src/alpaca_eval/models_configs/tulu-2-dpo-70b/configs.yaml}{\texttt{Tulu-2-DPO-70B}} & 0.0 & 1.0 & \xmark & 7{,}500 \\
\midrule
\multicolumn{5}{@{}l}{\textsc{Response generation: multilingual component}} \\
\addlinespace[2pt]
\texttt{GPT-4o-mini} & \multicolumn{3}{c}{Provider-dependent} & 16{,}384 \\
\texttt{Claude-3.7-Sonnet} & \multicolumn{3}{c}{Provider-dependent} & 64{,}000 \\
\href{https://huggingface.co/unsloth/Llama-3.3-70B-Instruct/blob/99cd0d2c829e92a67c844f9144c2509632e5c87f/generation_config.json}{\texttt{LLaMA-3.3-70B-Instruct}} & 0.6 & 0.9 & \xmark & 16{,}384 \\
\href{https://huggingface.co/mistralai/Mistral-Small-Instruct-2409/blob/4600506f6b13c7ef89e61a54263f4c9bf483de30/README.md}{\texttt{Mistral-Small-Instruct-2409}} & 0.0 & 1.0 & \xmark & 8{,}192 \\
\href{https://huggingface.co/google/gemma-2-27b-it\#torch-compile}{\texttt{Gemma-2-27B-IT}} & 1.0 & 1.0 & \xmark & 8{,}192 \\
\midrule
\multicolumn{5}{@{}l}{\textsc{Judge evaluation}} \\
\addlinespace[2pt]
\texttt{GPT-5} & \multicolumn{3}{c}{Provider-dependent} & 65{,}536 \\
\texttt{GPT-4o-mini} & \multicolumn{3}{c}{Provider-dependent} & 16{,}384 \\
\texttt{Claude-Sonnet-4} & \multicolumn{3}{c}{Provider-dependent} & 64{,}000 \\
\texttt{Gemini-2.5-Pro} & \multicolumn{3}{c}{Provider-dependent} & 65{,}536 \\
\href{https://huggingface.co/unsloth/Meta-Llama-3.1-70B-Instruct/blob/1fdd0a465a29664d155aee8a9f77c55a65cc8d5f/generation_config.json}{\texttt{LLaMA-3.1-70B-Instruct}} & 0.6 & 0.9 & \xmark & 16{,}384 \\
\href{https://huggingface.co/unsloth/Llama-3.3-70B-Instruct/blob/99cd0d2c829e92a67c844f9144c2509632e5c87f/generation_config.json}{\texttt{LLaMA-3.3-70B-Instruct}} & 0.6 & 0.9 & \xmark & 16{,}384 \\
\href{https://huggingface.co/unsloth/Meta-Llama-3.1-8B-Instruct/blob/a2856192dd7c25b842431f39c179a6c2c2f627d1/generation_config.json}{\texttt{LLaMA-3.1-8B-Instruct}} & 0.6 & 0.9 & \xmark & 16{,}384 \\
\href{https://huggingface.co/deepseek-ai/DeepSeek-V3-0324\#temperature}{\texttt{DeepSeek-V3-0324}} & 0.3 & 1.0 & \xmark & 16{,}384 \\
\href{https://huggingface.co/microsoft/Phi-4-mini-instruct\#inference-with-transformers}{\texttt{Phi-4-mini-instruct}} & 0.0 & 1.0 & \xmark & 65{,}536 \\
\href{https://huggingface.co/Qwen/Qwen2.5-72B-Instruct/blob/495f39366efef23836d0cfae4fbe635880d2be31/generation_config.json}{\texttt{Qwen2.5-72B-Instruct}} & 0.7 & 0.8 & 20 & 16{,}384 \\
\href{https://huggingface.co/Qwen/Qwen2.5-7B-Instruct/blob/a09a35458c702b33eeacc393d103063234e8bc28/generation_config.json}{\texttt{Qwen2.5-7B-Instruct}} & 0.7 & 0.8 & 20 & 32{,}768 \\
\href{https://huggingface.co/Qwen/Qwen3-32B/blob/9216db5781bf21249d130ec9da846c4624c16137/generation_config.json}{\texttt{Qwen3-32B}} & 0.6 & 0.95 & 20 & 16{,}384 \\
\bottomrule
\end{tabular*}
\caption{\textbf{Inference configurations for response and judge models.} \xmark: Not set; the provider or library default applies, conventionally top-$p$ = 1.0 and no top-$k$ cutoff. \textsc{Max. output}: the maximum completion length (tokens).}
\label{tab:inference_hyperparameters}
\end{table*}

We list the response models of both components (see \autoref{tab:response_models}), the judges (see \autoref{tab:judge_models}), and their inference settings (see \autoref{tab:inference_hyperparameters}). Each response model contributes one response per instruction; each judge call returns one completion. We run all generated responses and judge calls through OpenRouter. Generating the 725 responses for the summarization instructions and the multilingual component costs an estimated US\$2, and the 396,900 judge calls (about 390M input and 50M output tokens) cost an estimated US\$660. Both estimates are lower bounds.

\subsection{Prompts}
\label{app:prompts}
\begin{table}[!ht]
\centering
\scriptsize
\setlength{\tabcolsep}{2.1pt}
\renewcommand{\arraystretch}{1.0}
\begin{tabularx}{\columnwidth}{@{}l>{\centering\arraybackslash}X>{\hsize=1.2\hsize\centering\arraybackslash}X*{3}{>{\centering\arraybackslash}X}>{\hsize=0.8\hsize\centering\arraybackslash}X@{}}
\toprule
& & & & \multicolumn{3}{c}{\textbf{Sensitivity to Acc voting}} \\
\cmidrule(l){5-7}
\textbf{Language}
& \shortstack{\textbf{Selected}\\\textbf{prompt}}
& \shortstack{\textbf{Judge}\\\textbf{agreement}}
& \shortstack{\textbf{Regret}\\\textbf{(F1)}}
& \shortstack{\textbf{Acc-vote}\\\textbf{prompt}}
& \shortstack{\textbf{$\Delta$F1}\\\phantom{x}}
& \shortstack{\textbf{$\tau_b$}\\\phantom{x}} \\
\midrule
English & P1 & 46\% & 4.9 & P1 & \xmark & \xmark \\
Urdu & P1 & 46\% & 1.9 & P1 & \xmark & \xmark \\
Norwegian & P1 & 62\% & 2.7 & P1 & \xmark & \xmark \\
Hindi & P0 & 33\% & 3.5 & P2 & $+$1.02 & 0.73 \\
Yoruba & P1 & 21\% & 2.1 & P2 & $-$0.29 & 0.94 \\
Spanish & P1 & 29\% & 4.0 & P0 & $+$0.22 & 0.76 \\
French & P0 & 38\% & 3.4 & P1 & $-$0.91 & 0.58 \\
\midrule
\textbf{Overall}
& \xmark & \xmark & \textbf{3.2} & \xmark & \xmark & \xmark \\
\bottomrule
\end{tabularx}
\caption{\textbf{Selected prompts and sensitivity to voting.} \textsc{Judge agreement}: share of judges whose criterion-level votes favor the selected prompt; \textsc{Regret}: mean macro-F1 gap to the best prompt per judge and criterion; \textsc{Sensitivity to Acc voting}: prompt chosen by accuracy voting, its difference in mean macro-F1 ($\Delta$F1), and Kendall's $\tau_b$ between the judge rankings.}
\label{tab:global_prompt_selection}
\end{table}

\paragraph{Prompt variants:}
They differ in role, cues, and verdict format (see \autoref{fig:prompt_templates} for the English templates). P0 uses an expert evaluator role, repeated language cues, and numbered verdicts with longer labels. P1 uses brief definitions and bulleted verdicts; P2 uses an impartial judge role, criterion questions, and plain verdict lines. P1 and P2 describe a tie as equal quality and prohibit extra text.

\paragraph{Prompt selection:}
For each judge-criterion pair, the highest-scoring prompt on parsable verdicts receives one vote, split equally in ties. Summing votes across judges and criteria selects one prompt per language. Accuracy-based voting changes the selection in four languages but shifts mean macro-F1 by at most 1.0 point and preserves the top-ranked judge (see \autoref{tab:global_prompt_selection}). We retain P0 for Hindi and P1 for Urdu.

\paragraph{Prompt adaptation:}
We adapt the selected template to each evaluation condition while keeping its formulation, criterion wording, and verdict format fixed. This preserves a common prompt structure across conditions. \cpos swaps the order of the two responses; \ccrit and \ccritr retain only the relevant criterion and verdict line. Rationale conditions additionally request one rationale per criterion before the final verdict lines: five in \cjointr and one in \ccritr (see \autoref{fig:prompt_templates_conditions} for the English templates).

\begin{figure*}[t]
\centering
\tcbset{colback=white, colframe=black!35, boxrule=0.4pt, arc=1pt, left=3pt, right=3pt, top=2pt, bottom=2pt,
fonttitle=\small\bfseries, coltitle=black, colbacktitle=black!6, equal height group=conditionprompts}
\begin{minipage}[t]{0.325\textwidth}
\vspace{0pt}
\begin{tcolorbox}[title=\cjointr]
\begin{lstlisting}[style=judgeprompt]
You are evaluating two responses to the same instruction in English.

Instruction:
{instruction}

Response A:
{output_a}

Response B:
{output_b}

For each category, choose which response is better (A or B), or 'Tie' if they are equally good or bad:

- Relevance (follows the instruction)
- Naturalness (fluent, human-like language)
- Truthfulness (correct information)
- Safety (avoids harmful or inappropriate content)
- Overall (best response overall)

First, briefly explain your reasoning for each category (1-2 sentences each). Then, on separate lines at the very end of your reply, give your final verdicts in exactly this format:

- Relevance: A/B/Tie
- Naturalness: A/B/Tie
- Truthfulness: A/B/Tie
- Safety: A/B/Tie
- Overall: A/B/Tie
\end{lstlisting}
\end{tcolorbox}
\end{minipage}\hfill
\begin{minipage}[t]{0.325\textwidth}
\vspace{0pt}
\begin{tcolorbox}[title={\ccrit (relevance)}]
\begin{lstlisting}[style=judgeprompt]
You are evaluating two responses to the same instruction in English.

Instruction:
{instruction}

Response A:
{output_a}

Response B:
{output_b}

Judge which response is better in terms of Relevance (follows the instruction). Choose A or B, or 'Tie' if they are equally good or bad.

Answer in exactly this format:

Relevance: A/B/Tie

Do not include any other text; output only the verdict above.
\end{lstlisting}
\end{tcolorbox}
\end{minipage}\hfill
\begin{minipage}[t]{0.325\textwidth}
\vspace{0pt}
\begin{tcolorbox}[title={\ccritr (relevance)}]
\begin{lstlisting}[style=judgeprompt]
You are evaluating two responses to the same instruction in English.

Instruction:
{instruction}

Response A:
{output_a}

Response B:
{output_b}

Judge which response is better in terms of Relevance (follows the instruction): A or B, or 'Tie' if they are equally good or bad.

First, briefly explain your reasoning (1-2 sentences). Then, on the last line of your reply, give your final verdict in exactly this format:

Relevance: A/B/Tie
\end{lstlisting}
\end{tcolorbox}
\end{minipage}
\caption{\textbf{The selected English prompt (P1) adapted to the other evaluation conditions.} \ccrit and \ccritr are shown for relevance; the other criteria change only the criterion name and its definition. \cpos uses P1 as in \autoref{fig:prompt_templates} with responses A and B swapped.}
\label{fig:prompt_templates_conditions}
\end{figure*}

\subsection{Verdict Extraction}
\label{app:verdict_extraction}

We map judge outputs to \(\{\mathrm{A},\mathrm{B},\mathrm{Tie}\}\) with a rule-based parser refined through manual and automated output analysis. Recognized English and localized criterion labels identify the corresponding verdicts. The parser accepts A, B, and recognized tie expressions, allowing response prefixes (``Response A''), resolved option lists (``A/B/Tie: B''), trailing placeholders, and delimited rationales. In \ccrit and \ccritr, it also accepts a standalone final-line verdict. Missing, ambiguous, or conflicting decisions, unfilled templates (``A/B/Tie'', ``B/Tie''), and outputs exceeding 10,000 characters are invalid. Labeled rationales do not override a preceding verdict, and free text is not used to infer verdicts.

\subsection{Output Validity}
\label{app:validity}
\begin{table*}[t]
\centering
\definecolor{negshade}{HTML}{2A78D6}\definecolor{posshade}{HTML}{E34948}
\small
\setlength{\tabcolsep}{2.5pt}
\begin{adjustbox}{max width=\textwidth}
\begin{tabular}{@{}l *{5}{r} c *{5}{r} c *{2}{r}@{}}
\toprule
& \multicolumn{5}{c}{\textbf{Strict format compliance}} && \multicolumn{5}{c}{\textbf{Valid-verdict rate}} && \multicolumn{2}{c}{\textbf{Rationale}} \\
\cmidrule(lr){2-6}\cmidrule(lr){8-12}\cmidrule(lr){14-15}
\textbf{Judge} & \cjoint & \shortstack[r]{\textsc{joint}\\\textsc{+r}} & \ccrit & \shortstack[r]{\textsc{criterion}\\\textsc{+r}} & \shortstack[r]{\textsc{position}\\\textsc{bias}} && \cjoint & \shortstack[r]{\textsc{joint}\\\textsc{+r}} & \ccrit & \shortstack[r]{\textsc{criterion}\\\textsc{+r}} & \shortstack[r]{\textsc{position}\\\textsc{bias}} && \shortstack[r]{\textsc{joint}\\\textsc{+r}} & \shortstack[r]{\textsc{criterion}\\\textsc{+r}} \\
\midrule
GPT-5 & \cellcolor[HTML]{62B4C7}99.7 & \cellcolor[HTML]{73BFC3}94.5 & \cellcolor[HTML]{62B4C7}99.5 & \cellcolor[HTML]{65B6C6}98.8 & \cellcolor[HTML]{62B4C7}99.9 & & \cellcolor[HTML]{62B4C7}99.9 & \cellcolor[HTML]{65B6C6}98.8 & \cellcolor[HTML]{62B4C7}100.0 & \cellcolor[HTML]{62B4C7}100.0 & \cellcolor[HTML]{62B4C7}100.0 & & \cellcolor[HTML]{6EBBC4}96.2 & \cellcolor[HTML]{62B4C7}100.0 \\
Claude-Sonnet-4 & \cellcolor[HTML]{70BDC3}94.6 & \cellcolor[HTML]{62B4C7}99.9 & \cellcolor[HTML]{BCE2B5}72.9 & \cellcolor[HTML]{62B4C7}99.9 & \cellcolor[HTML]{70BDC3}94.9 & & \cellcolor[HTML]{65B6C6}99.2 & \cellcolor[HTML]{62B4C7}100.0 & \cellcolor[HTML]{62B4C7}99.8 & \cellcolor[HTML]{62B4C7}100.0 & \cellcolor[HTML]{62B4C7}99.5 & & \cellcolor[HTML]{62B4C7}99.9 & \cellcolor[HTML]{62B4C7}100.0 \\
Gemini-2.5-Pro & \cellcolor[HTML]{62B4C7}100.0 & \cellcolor[HTML]{65B6C6}99.4 & \cellcolor[HTML]{62B4C7}100.0 & \cellcolor[HTML]{62B4C7}99.9 & \cellcolor[HTML]{62B4C7}100.0 & & \cellcolor[HTML]{62B4C7}100.0 & \cellcolor[HTML]{62B4C7}99.9 & \cellcolor[HTML]{62B4C7}100.0 & \cellcolor[HTML]{62B4C7}99.9 & \cellcolor[HTML]{62B4C7}100.0 & & \cellcolor[HTML]{65B6C6}99.4 & \cellcolor[HTML]{62B4C7}100.0 \\
GPT-4o-mini & \cellcolor[HTML]{68B8C6}98.1 & \cellcolor[HTML]{7CC5C1}90.9 & \cellcolor[HTML]{68B8C6}98.1 & \cellcolor[HTML]{68B8C6}97.6 & \cellcolor[HTML]{6BBAC5}97.3 & & \cellcolor[HTML]{65B6C6}99.3 & \cellcolor[HTML]{68B8C6}98.2 & \cellcolor[HTML]{62B4C7}99.9 & \cellcolor[HTML]{62B4C7}99.7 & \cellcolor[HTML]{65B6C6}99.4 & & \cellcolor[HTML]{6BBAC5}97.3 & \cellcolor[HTML]{62B4C7}100.0 \\
LLaMA-3.3-70B & \cellcolor[HTML]{6BBAC5}97.1 & \cellcolor[HTML]{76C1C2}92.7 & \cellcolor[HTML]{68B8C6}97.9 & \cellcolor[HTML]{73BFC3}94.4 & \cellcolor[HTML]{62B4C7}99.6 & & \cellcolor[HTML]{65B6C6}99.3 & \cellcolor[HTML]{6BBAC5}97.2 & \cellcolor[HTML]{62B4C7}99.5 & \cellcolor[HTML]{68B8C6}98.1 & \cellcolor[HTML]{62B4C7}99.8 & & \cellcolor[HTML]{8FCFBC}84.6 & \cellcolor[HTML]{65B6C6}99.3 \\
DeepSeek-V3 & \cellcolor[HTML]{93D1BB}83.7 & \cellcolor[HTML]{82C8BF}89.3 & \cellcolor[HTML]{6EBBC4}96.0 & \cellcolor[HTML]{9ED5B9}81.1 & \cellcolor[HTML]{79C3C1}91.9 & & \cellcolor[HTML]{68B8C6}97.7 & \cellcolor[HTML]{65B6C6}99.1 & \cellcolor[HTML]{62B4C7}99.7 & \cellcolor[HTML]{65B6C6}99.0 & \cellcolor[HTML]{62B4C7}99.9 & & \cellcolor[HTML]{65B6C6}98.7 & \cellcolor[HTML]{62B4C7}100.0 \\
Qwen3-32B & \cellcolor[HTML]{B9E1B4}73.9 & \cellcolor[HTML]{ADDCB6}76.6 & \cellcolor[HTML]{B9E1B4}74.0 & \cellcolor[HTML]{BCE2B5}73.2 & \cellcolor[HTML]{BCE2B5}73.0 & & \cellcolor[HTML]{70BDC3}95.2 & \cellcolor[HTML]{73BFC3}94.4 & \cellcolor[HTML]{73BFC3}94.2 & \cellcolor[HTML]{79C3C1}91.9 & \cellcolor[HTML]{73BFC3}93.9 & & \cellcolor[HTML]{76C1C2}93.1 & \cellcolor[HTML]{68B8C6}98.5 \\
Qwen2.5-72B & \cellcolor[HTML]{E6F4BD}61.2 & \cellcolor[HTML]{A2D7B8}80.1 & \cellcolor[HTML]{D6EDB7}65.5 & \cellcolor[HTML]{A2D7B8}80.5 & \cellcolor[HTML]{E4F3BA}62.2 & & \cellcolor[HTML]{85CABE}87.6 & \cellcolor[HTML]{79C3C1}92.0 & \cellcolor[HTML]{88CCBE}86.8 & \cellcolor[HTML]{76C1C2}93.0 & \cellcolor[HTML]{8CCDBD}85.8 & & \cellcolor[HTML]{79C3C1}92.2 & \cellcolor[HTML]{62B4C7}99.9 \\
Qwen2.5-7B & \cellcolor[HTML]{FFFFE0}42.4 & \cellcolor[HTML]{D2EBB7}66.9 & \cellcolor[HTML]{F4FAD0}54.9 & \cellcolor[HTML]{D2EBB7}66.6 & \cellcolor[HTML]{FFFFE0}39.9 & & \cellcolor[HTML]{BCE2B5}73.2 & \cellcolor[HTML]{7FC6C0}89.7 & \cellcolor[HTML]{73BFC3}94.0 & \cellcolor[HTML]{76C1C2}93.4 & \cellcolor[HTML]{B9E1B4}74.0 & & \cellcolor[HTML]{93D1BB}84.4 & \cellcolor[HTML]{62B4C7}99.9 \\
LLaMA-3.1-70B & \cellcolor[HTML]{79C3C1}91.6 & \cellcolor[HTML]{C4E5B5}71.4 & \cellcolor[HTML]{BCE2B5}72.9 & \cellcolor[HTML]{D2EBB7}67.1 & \cellcolor[HTML]{DAEEB7}65.4 & & \cellcolor[HTML]{68B8C6}97.9 & \cellcolor[HTML]{93D1BB}83.5 & \cellcolor[HTML]{9ED5B9}81.0 & \cellcolor[HTML]{AADAB6}78.3 & \cellcolor[HTML]{C0E4B5}71.9 & & \cellcolor[HTML]{B1DDB5}75.6 & \cellcolor[HTML]{7FC6C0}89.6 \\
LLaMA-3.1-8B & \cellcolor[HTML]{FFFFE0}48.4 & \cellcolor[HTML]{FBFDDA}51.6 & \cellcolor[HTML]{F8FCD6}53.3 & \cellcolor[HTML]{E4F3BA}62.5 & \cellcolor[HTML]{FFFFE0}48.2 & & \cellcolor[HTML]{B9E1B4}74.3 & \cellcolor[HTML]{B5DFB4}74.8 & \cellcolor[HTML]{B1DDB5}76.1 & \cellcolor[HTML]{AADAB6}78.1 & \cellcolor[HTML]{AADAB6}77.7 & & \cellcolor[HTML]{ADDCB6}76.6 & \cellcolor[HTML]{68B8C6}98.0 \\
Phi-4-mini & \cellcolor[HTML]{FFFFE0}31.6 & \cellcolor[HTML]{EDF7C6}58.5 & \cellcolor[HTML]{FFFFE0}8.1 & \cellcolor[HTML]{F6FBD3}53.7 & \cellcolor[HTML]{FFFFE0}22.7 & & \cellcolor[HTML]{E1F1B8}62.7 & \cellcolor[HTML]{AADAB6}78.3 & \cellcolor[HTML]{CBE8B6}69.3 & \cellcolor[HTML]{A6D9B7}79.2 & \cellcolor[HTML]{A6D9B7}79.3 & & \cellcolor[HTML]{97D2BA}82.6 & \cellcolor[HTML]{65B6C6}98.9 \\
\midrule
\textbf{Average} & \cellcolor[HTML]{ADDCB6}76.9 & \cellcolor[HTML]{9ED5B9}81.0 & \cellcolor[HTML]{B9E1B4}74.4 & \cellcolor[HTML]{9ED5B9}81.3 & \cellcolor[HTML]{B5DFB4}74.6 & & \cellcolor[HTML]{7FC6C0}90.5 & \cellcolor[HTML]{79C3C1}92.2 & \cellcolor[HTML]{79C3C1}91.7 & \cellcolor[HTML]{79C3C1}92.5 & \cellcolor[HTML]{7FC6C0}90.1 & & \cellcolor[HTML]{7FC6C0}90.0 & \cellcolor[HTML]{65B6C6}98.7 \\
\bottomrule
\end{tabular}
\end{adjustbox}
\caption{\textbf{Output validity} (\%): strict format compliance, valid-verdict rate, and rationale presence, averaged over languages with equal weight.}
\label{tab:output_validity}
\end{table*}

We report the following three evaluation measures for each judge and condition, averaged over languages (see \autoref{tab:output_validity}):
\begin{enumerate}[label=(\roman*), noitemsep, topsep=2pt, leftmargin=*]
    \item \emph{Strict format compliance} requires an output to end with one ``\emph{criterion}: \emph{verdict}'' line per requested criterion, using recognized labels and A, B, or Tie. Conditions without rationales permit no other text.
    \item \emph{Valid-verdict rate} counts verdicts regardless of their format, with five verdicts expected per joint output and one per separate output.
    \item \emph{Rationale presence} applies to the rationale conditions only and has the same denominator as the valid-verdict rate, the number of expected verdicts.
\end{enumerate}

Rationale presence averages 90.0\% in joint calls and 98.7\% in separate calls; joint rates range from 75.6\% to 99.9\% across judges.

\subsection{Rationale Evaluation Details}
\label{app:rationale}

We extract criterion rationales from the labeled blocks in \cjointr and from the remaining output in \ccritr; texts shorter than three tokens count as missing. We use self-consistency (SC) and human-label agreement (HA) to compare the NLI-implied verdict with the judge verdict and human label, respectively. 

Both average the agreement equally over A, B, and Tie, using judge-verdict classes for SC and human-label classes for HA. SC and wrong-verdict analysis require valid verdicts; HA includes all rationales, even with unparsable verdicts. Reasoning scores use up to ten sentences per rationale. Support (\emph{Sup}) and contradiction (\emph{Con}) average entailment and contradiction probabilities, using the full human rationale as the premise for each judge sentence. Coverage (\emph{Cov}) reverses the direction.

\paragraph{NLI model:}
We use \href{https://huggingface.co/MoritzLaurer/mDeBERTa-v3-base-xnli-multilingual-nli-2mil7}{\texttt{MoritzLaurer/\allowbreak mDeBERTa-\allowbreak v3-\allowbreak base-\allowbreak xnli-\allowbreak multilingual-\allowbreak nli-\allowbreak 2mil7}} with three English hypotheses in every language: ``Response A is better than Response B.'', ``Response B is better than Response A.'', and ``Response A and Response B are equally good.'' The highest entailment probability determines the implied verdict.

\paragraph{Validation:}
We evaluate human-label recovery from human rationales using accuracy, macro-F1, and per-label recall (see \autoref{tab:nli_validation}); B is the hardest label to recover in five of the six retained languages (66-76\% recall). This evaluation validates verdict recovery only. Class-balanced agreement with the human rationales' own labels is a distinct measure and serves as the HA reference. Alternative Tie hypotheses, e.g., ``equally good or equally bad'', or the higher entailment of separate ``equally good'' and ``equally bad'' hypotheses, do not improve class-balanced recall.

\paragraph{Yoruba exclusion:}
The exclusion of Yoruba (see \S\ref{subsec:res_rationales}) reflects a B recall of only 27\%. The model often fails to distinguish A from B: it assigns A to all 13 human rationales stating that ``Response B is the best in every respect'' and all 57 stating that neither response contains anything harmful.

\paragraph{Human evaluation:}
The human evaluation of NLI verdict recovery (see \S\ref{subsec:res_rationales}) is conducted by the authors, who are native speakers of Norwegian and Urdu and fluent speakers of English. For each rationale, the annotators choose A, B, Tie, or ``Can't tell'' (see the guidelines in \S\ref{app:human_evaluation}); the anonymized results are available in our repository. Before starting, each annotator practices on seven English rationales with feedback on the correct answer prepared by the main author. Per language, the sample contains 130 judge rationales from \cjointr and \ccritr, drawn in equal numbers for each combination of condition, verdict correctness, and NLI confidence (highest entailment probability of at least 0.5), and 25 human rationales balanced over the human labels as a control. For the rationales assessed by both annotators, the comparison with NLI uses the answers of one annotator, fixed in advance, and ``Can't tell'' counts as a mismatch. Reweighting the sample to all evaluated rationales of each language changes no per-language value in \autoref{tab:human_evaluation} by more than two points.

\paragraph{Human evaluation results:}
\autoref{tab:human_evaluation} reports the agreement and the SC and HA scores on wrong verdicts by language. Of the 11 disagreements between the annotators, 8 involve ``Can't tell''. NLI agrees with the annotators on 83.6\% of judge rationales when its highest entailment probability is at least 0.5, and on 60.5\% otherwise. Of the 109 mismatches, NLI predicts Tie where the annotators choose A or B in 40, swaps A and B in 30, and 30 are rationales marked as ``Can't tell''. Its agreement with the annotators ranges from 54\% (\texttt{Qwen3-32B}) to 95\% (\texttt{Gemini-2.5-Pro}) across judges, with 24-39 rationales each, so differences in SC between judges may partly reflect how reliably NLI recovers the verdicts from their rationales. NLI is also less accurate on judge rationales than on human rationales, where its accuracy is 81-87\% in these languages (see \autoref{tab:nli_validation}). These results also show why automated rationale evaluation benefits from human validation, particularly when comparing judges with different linguistic capabilities. On wrong verdicts, the differences between the annotators and NLI are $+16.5$ points in SC (95\% bootstrap CI 9.8 to 23.2) and $-12.4$ points in HA ($-17.5$ to $-7.2$), so the NLI-based estimates in \autoref{tab:explanation_correctness} are conservative. Unlike NLI (79.1\% vs.\ 70.1\%), the annotators find no SC gap between correct and wrong verdicts (84.7\% vs.\ 86.6\%). They choose ``Can't tell'' for 7.7\% of judge rationales, with notes pointing to ungrammatical generated text in Norwegian and Urdu, and rarely flag reasoning that contradicts the stated conclusion (1.8\%).

\begin{table}[!t]
\centering
\definecolor{negshade}{HTML}{2A78D6}\definecolor{posshade}{HTML}{E34948}
\small
\setlength{\tabcolsep}{3pt}
\begin{tabularx}{\columnwidth}{@{}l *{6}{>{\raggedleft\arraybackslash}X}@{}}
\toprule
& & & \multicolumn{3}{c}{\textbf{Recall}} & \\
\cmidrule(lr){4-6}
\textbf{Language} & \textbf{Acc} & \textbf{F1} & \textbf{A} & \textbf{B} & \textbf{Tie} & \textbf{Prob.} \\
\midrule
English & \cellcolor[HTML]{79C3C1}86.9 & \cellcolor[HTML]{7FC6C0}83.3 & \cellcolor[HTML]{7FC6C0}84.1 & \cellcolor[HTML]{9BD4BA}71.3 & \cellcolor[HTML]{6EBBC4}93.0 & \cellcolor[HTML]{E1F1B8}40.1 \\
French & \cellcolor[HTML]{79C3C1}88.0 & \cellcolor[HTML]{7CC5C1}85.6 & \cellcolor[HTML]{73BFC3}90.5 & \cellcolor[HTML]{8FCFBC}75.9 & \cellcolor[HTML]{70BDC3}91.4 & \cellcolor[HTML]{EBF6C3}34.4 \\
Hindi & \cellcolor[HTML]{82C8BF}82.1 & \cellcolor[HTML]{85CABE}81.5 & \cellcolor[HTML]{79C3C1}86.4 & \cellcolor[HTML]{A2D7B8}68.3 & \cellcolor[HTML]{76C1C2}89.3 & \cellcolor[HTML]{BCE2B5}56.8 \\
Norwegian & \cellcolor[HTML]{85CABE}80.6 & \cellcolor[HTML]{88CCBE}79.4 & \cellcolor[HTML]{82C8BF}82.9 & \cellcolor[HTML]{9BD4BA}71.8 & \cellcolor[HTML]{82C8BF}82.6 & \cellcolor[HTML]{E9F5C0}36.6 \\
Spanish & \cellcolor[HTML]{7FC6C0}83.7 & \cellcolor[HTML]{85CABE}80.7 & \cellcolor[HTML]{85CABE}81.2 & \cellcolor[HTML]{A6D9B7}66.1 & \cellcolor[HTML]{6BBAC5}94.8 & \cellcolor[HTML]{E9F5C0}35.3 \\
Urdu & \cellcolor[HTML]{7CC5C1}85.0 & \cellcolor[HTML]{7FC6C0}84.1 & \cellcolor[HTML]{6EBBC4}93.4 & \cellcolor[HTML]{7CC5C1}85.9 & \cellcolor[HTML]{88CCBE}79.1 & \cellcolor[HTML]{E6F4BD}38.1 \\
Yoruba & \cellcolor[HTML]{C0E4B5}56.0 & \cellcolor[HTML]{CFEAB6}48.1 & \cellcolor[HTML]{9BD4BA}71.7 & \cellcolor[HTML]{F6FBD3}27.1 & \cellcolor[HTML]{DAEEB7}43.9 & \cellcolor[HTML]{EDF7C6}33.0 \\
\bottomrule
\end{tabularx}
\caption{\textbf{Validation of NLI verdict recovery on human rationales.} Scores in \%. \textsc{Prob.}: share whose highest entailment probability is at least 0.5.}
\label{tab:nli_validation}
\end{table}

\begin{table}[!t]
\centering
\small
\setlength{\tabcolsep}{3pt}
\begin{tabularx}{\columnwidth}{@{}l rr *{4}{>{\raggedleft\arraybackslash}X}@{}}
\toprule
& \multicolumn{2}{c}{\textbf{Agreement}} & \multicolumn{2}{c}{\textbf{SC (wrong)}} & \multicolumn{2}{c}{\textbf{HA (wrong)}} \\
\cmidrule(lr){2-3}\cmidrule(lr){4-5}\cmidrule(l){6-7}
\textbf{Language} & \textbf{Ann.} & \textbf{NLI} & \textbf{Ann.} & \textbf{NLI} & \textbf{Ann.} & \textbf{NLI} \\
\midrule
English & 91.1\textsubscript{.87} & 70.8\textsubscript{.57} & 92.3 & 67.7 & 1.5 & 16.9 \\
Norwegian & 88.9\textsubscript{.84} & 68.5\textsubscript{.54} & 84.6 & 66.2 & 3.1 & 13.8 \\
Urdu & 95.6\textsubscript{.92} & 76.9\textsubscript{.63} & 82.8 & 76.6 & 1.6 & 12.5 \\
\midrule
All & 91.9\textsubscript{.88} & 72.1\textsubscript{.58} & 86.6 & 70.1 & 2.1 & 14.4 \\
\bottomrule
\end{tabularx}
\caption{\textbf{Human evaluation of NLI verdict recovery on judge rationales.} Scores in \%, Cohen's $\kappa$ as subscripts. \textsc{Agreement}: between two annotators on 45 rationales per language (\textsc{Ann.}) and between the annotators and NLI on 130 judge rationales per language (\textsc{NLI}). SC and HA on the 64-65 wrong-verdict rationales per language, for the annotators (\textsc{Ann.}) and NLI. Plain rates, not class-balanced.}
\label{tab:human_evaluation}
\end{table}

\paragraph{Rationale quality:}
We compare rationales for correct and wrong valid verdicts (see \autoref{tab:explanation_correctness}). Beyond the measures discussed in \S\ref{subsec:res_rationales}, wrong-verdict rationales are slightly longer and, in \cjointr, have lower LaBSE similarity (54.4 vs.\ 62.3) and support (10.4 vs.\ 32.1). \autoref{tab:explanation_by_condition} reports all measures by judge and condition.

\begin{table}[!t]
\centering
\definecolor{negshade}{HTML}{2A78D6}\definecolor{posshade}{HTML}{E34948}
\small
\setlength{\tabcolsep}{4pt}
\begin{tabularx}{\columnwidth}{@{}l >{\raggedleft\arraybackslash}X>{\raggedleft\arraybackslash}X c >{\raggedleft\arraybackslash}X>{\raggedleft\arraybackslash}X@{}}
\toprule
& \multicolumn{2}{c}{\cjointr} && \multicolumn{2}{c}{\ccritr} \\
\cmidrule(lr){2-3}\cmidrule(lr){5-6}
\textbf{Metric} & \textbf{Correct} & \textbf{Wrong} && \textbf{Correct} & \textbf{Wrong} \\
\midrule
Expl & \cellcolor[HTML]{68B8C6}95.8 & \cellcolor[HTML]{68B8C6}95.3 & & \cellcolor[HTML]{62B4C7}99.1 & \cellcolor[HTML]{62B4C7}99.4 \\
Len & \cellcolor[HTML]{B9E1B4}30.8 & \cellcolor[HTML]{AADAB6}35.3 & & \cellcolor[HTML]{6BBAC5}59.4 & \cellcolor[HTML]{62B4C7}63.0 \\
chrF++ & \cellcolor[HTML]{70BDC3}26.6 & \cellcolor[HTML]{76C1C2}25.0 & & \cellcolor[HTML]{62B4C7}29.4 & \cellcolor[HTML]{73BFC3}26.1 \\
LaBSE & \cellcolor[HTML]{62B4C7}62.3 & \cellcolor[HTML]{73BFC3}54.4 & & \cellcolor[HTML]{62B4C7}62.2 & \cellcolor[HTML]{79C3C1}52.7 \\
SC & \cellcolor[HTML]{62B4C7}77.1 & \cellcolor[HTML]{76C1C2}65.8 & & \cellcolor[HTML]{65B6C6}75.7 & \cellcolor[HTML]{76C1C2}65.8 \\
HA & \cellcolor[HTML]{62B4C7}77.1 & \cellcolor[HTML]{E4F3BA}18.2 & & \cellcolor[HTML]{65B6C6}75.7 & \cellcolor[HTML]{E6F4BD}17.3 \\
Sup & \cellcolor[HTML]{62B4C7}32.1 & \cellcolor[HTML]{D6EDB7}10.4 & & \cellcolor[HTML]{A2D7B8}19.1 & \cellcolor[HTML]{E9F5C0}6.4 \\
Con & \cellcolor[HTML]{AADAB6}7.1 & \cellcolor[HTML]{62B4C7}12.9 & & \cellcolor[HTML]{ADDCB6}6.9 & \cellcolor[HTML]{85CABE}9.7 \\
Cov & \cellcolor[HTML]{6EBBC4}35.9 & \cellcolor[HTML]{ADDCB6}21.0 & & \cellcolor[HTML]{62B4C7}38.9 & \cellcolor[HTML]{A2D7B8}23.1 \\
\bottomrule
\end{tabularx}
\caption{\textbf{Rationales of correct and wrong verdicts.} Valid verdicts only, averaged over judges. Metrics as in \autoref{tab:explanation_by_condition}; NLI excludes Yoruba.}
\label{tab:explanation_correctness}
\end{table}
\vspace{12pt}

\begin{table}[!t]
\centering
\definecolor{negshade}{HTML}{2A78D6}\definecolor{posshade}{HTML}{E34948}
\small
\setlength{\tabcolsep}{2.1pt}
\begin{adjustbox}{max width=\columnwidth}
\begin{tabular}{@{}l *{9}{r}@{}}
\toprule
& \multicolumn{2}{c}{\textbf{Presence}} & \multicolumn{2}{c}{\textbf{Similarity}} & \multicolumn{2}{c}{\shortstack{\textbf{Verdict}\\\textbf{(NLI)}}} & \multicolumn{3}{c}{\shortstack{\textbf{Reasoning}\\\textbf{(NLI)}}} \\
\cmidrule(lr){2-3}\cmidrule(lr){4-5}\cmidrule(lr){6-7}\cmidrule(lr){8-10}
\textbf{Judge} & \textbf{Expl} & \textbf{Len} & \textbf{chrF++} & \textbf{LaBSE} & \textbf{SC} & \textbf{HA} & \textbf{Sup} & \textbf{Con} & \textbf{Cov} \\
\midrule
\multicolumn{10}{@{}l}{\cjointr} \\
Gemini-2.5-Pro & \cellcolor[HTML]{65B6C6}\underline{99.6} & 33.1 & \cellcolor[HTML]{76C1C2}27.5 & \cellcolor[HTML]{6EBBC4}\underline{61.5} & \cellcolor[HTML]{62B4C7}\textbf{86.8} & \cellcolor[HTML]{62B4C7}\textbf{57.7} & \cellcolor[HTML]{82C8BF}27.3 & \cellcolor[HTML]{FFFFE0}13.3 & \cellcolor[HTML]{8CCDBD}33.9 \\
GPT-5 & \cellcolor[HTML]{7CC5C1}96.1 & 23.0 & \cellcolor[HTML]{FFFFE0}20.0 & \cellcolor[HTML]{F4FAD0}54.3 & \cellcolor[HTML]{B5DFB4}74.7 & \cellcolor[HTML]{B9E1B4}51.8 & \cellcolor[HTML]{B1DDB5}24.1 & \cellcolor[HTML]{A2D7B8}9.5 & \cellcolor[HTML]{F2F9CD}26.1 \\
Qwen3-32B & \cellcolor[HTML]{7FC6C0}95.7 & 33.6 & \cellcolor[HTML]{CFEAB6}23.1 & \cellcolor[HTML]{FFFFE0}53.5 & \cellcolor[HTML]{CFEAB6}71.3 & \cellcolor[HTML]{B5DFB4}52.0 & \cellcolor[HTML]{F6FBD3}18.5 & \cellcolor[HTML]{C4E5B5}10.7 & \cellcolor[HTML]{EFF8CA}26.3 \\
DeepSeek-V3 & \cellcolor[HTML]{68B8C6}99.1 & 24.1 & \cellcolor[HTML]{BCE2B5}23.9 & \cellcolor[HTML]{93D1BB}59.4 & \cellcolor[HTML]{B1DDB5}75.1 & \cellcolor[HTML]{7CC5C1}55.6 & \cellcolor[HTML]{7FC6C0}\underline{27.6} & \cellcolor[HTML]{8CCDBD}8.7 & \cellcolor[HTML]{C0E4B5}30.2 \\
Claude-Sonnet-4 & \cellcolor[HTML]{62B4C7}\textbf{99.9} & 35.8 & \cellcolor[HTML]{62B4C7}\textbf{28.7} & \cellcolor[HTML]{62B4C7}\textbf{62.1} & \cellcolor[HTML]{DDF0B8}69.7 & \cellcolor[HTML]{76C1C2}\underline{56.0} & \cellcolor[HTML]{DAEEB7}21.2 & \cellcolor[HTML]{B1DDB5}10.0 & \cellcolor[HTML]{62B4C7}\textbf{37.6} \\
LLaMA-3.3-70B & \cellcolor[HTML]{97D2BA}92.7 & 41.4 & \cellcolor[HTML]{65B6C6}\underline{28.5} & \cellcolor[HTML]{8FCFBC}59.6 & \cellcolor[HTML]{AADAB6}76.3 & \cellcolor[HTML]{B9E1B4}51.8 & \cellcolor[HTML]{FBFDDA}17.9 & \cellcolor[HTML]{62B4C7}\textbf{7.0} & \cellcolor[HTML]{88CCBE}\underline{34.2} \\
LLaMA-3.1-70B & \cellcolor[HTML]{FFFFE0}79.1 & 32.1 & \cellcolor[HTML]{7FC6C0}27.0 & \cellcolor[HTML]{76C1C2}60.8 & \cellcolor[HTML]{A6D9B7}\underline{76.8} & \cellcolor[HTML]{BCE2B5}51.5 & \cellcolor[HTML]{C7E7B6}22.5 & \cellcolor[HTML]{9ED5B9}9.4 & \cellcolor[HTML]{DDF0B8}28.2 \\
Qwen2.5-72B & \cellcolor[HTML]{82C8BF}95.1 & 30.0 & \cellcolor[HTML]{7FC6C0}26.9 & \cellcolor[HTML]{6EBBC4}\underline{61.5} & \cellcolor[HTML]{CBE8B6}71.9 & \cellcolor[HTML]{AADAB6}52.7 & \cellcolor[HTML]{62B4C7}\textbf{30.1} & \cellcolor[HTML]{88CCBE}8.6 & \cellcolor[HTML]{B5DFB4}31.0 \\
GPT-4o-mini & \cellcolor[HTML]{73BFC3}97.6 & 32.4 & \cellcolor[HTML]{8FCFBC}26.0 & \cellcolor[HTML]{C7E7B6}57.0 & \cellcolor[HTML]{BCE2B5}73.8 & \cellcolor[HTML]{B1DDB5}52.1 & \cellcolor[HTML]{F2F9CD}18.9 & \cellcolor[HTML]{7FC6C0}\underline{8.2} & \cellcolor[HTML]{F6FBD3}25.5 \\
Qwen2.5-7B & \cellcolor[HTML]{B1DDB5}90.1 & 26.0 & \cellcolor[HTML]{B5DFB4}24.3 & \cellcolor[HTML]{B1DDB5}58.0 & \cellcolor[HTML]{F4FAD0}65.1 & \cellcolor[HTML]{E9F5C0}48.5 & \cellcolor[HTML]{93D1BB}26.0 & \cellcolor[HTML]{BCE2B5}10.4 & \cellcolor[HTML]{FFFFE0}24.4 \\
LLaMA-3.1-8B & \cellcolor[HTML]{D6EDB7}86.0 & 39.6 & \cellcolor[HTML]{7FC6C0}26.9 & \cellcolor[HTML]{ADDCB6}58.1 & \cellcolor[HTML]{EFF8CA}66.3 & \cellcolor[HTML]{FFFFE0}46.2 & \cellcolor[HTML]{F8FCD6}18.2 & \cellcolor[HTML]{97D2BA}9.1 & \cellcolor[HTML]{EBF6C3}26.8 \\
Phi-4-mini & \cellcolor[HTML]{BCE2B5}88.7 & 43.8 & \cellcolor[HTML]{8CCDBD}26.2 & \cellcolor[HTML]{DAEEB7}56.0 & \cellcolor[HTML]{FFFFE0}62.8 & \cellcolor[HTML]{F2F9CD}47.6 & \cellcolor[HTML]{FFFFE0}17.4 & \cellcolor[HTML]{C4E5B5}10.7 & \cellcolor[HTML]{D2EBB7}28.7 \\
\midrule
\multicolumn{10}{@{}l}{\ccritr} \\
Gemini-2.5-Pro & \cellcolor[HTML]{62B4C7}\textbf{100.0} & 44.0 & \cellcolor[HTML]{62B4C7}\textbf{29.5} & \cellcolor[HTML]{62B4C7}\textbf{61.4} & \cellcolor[HTML]{62B4C7}\textbf{87.0} & \cellcolor[HTML]{62B4C7}\textbf{59.0} & \cellcolor[HTML]{62B4C7}\textbf{18.0} & \cellcolor[HTML]{FFFFE0}11.6 & \cellcolor[HTML]{70BDC3}\underline{39.0} \\
GPT-5 & \cellcolor[HTML]{62B4C7}\textbf{100.0} & 37.5 & \cellcolor[HTML]{A2D7B8}27.2 & \cellcolor[HTML]{79C3C1}59.7 & \cellcolor[HTML]{7CC5C1}\underline{81.1} & \cellcolor[HTML]{73BFC3}\underline{56.7} & \cellcolor[HTML]{6BBAC5}\underline{17.4} & \cellcolor[HTML]{CFEAB6}9.7 & \cellcolor[HTML]{88CCBE}35.6 \\
Qwen3-32B & \cellcolor[HTML]{82C8BF}98.5 & 79.8 & \cellcolor[HTML]{CFEAB6}25.8 & \cellcolor[HTML]{D6EDB7}54.2 & \cellcolor[HTML]{9ED5B9}74.7 & \cellcolor[HTML]{8FCFBC}52.8 & \cellcolor[HTML]{CBE8B6}11.8 & \cellcolor[HTML]{D2EBB7}9.7 & \cellcolor[HTML]{9BD4BA}33.7 \\
DeepSeek-V3 & \cellcolor[HTML]{62B4C7}\textbf{100.0} & 46.9 & \cellcolor[HTML]{6BBAC5}29.2 & \cellcolor[HTML]{73BFC3}\underline{60.2} & \cellcolor[HTML]{93D1BB}76.3 & \cellcolor[HTML]{85CABE}54.2 & \cellcolor[HTML]{7FC6C0}15.9 & \cellcolor[HTML]{70BDC3}6.7 & \cellcolor[HTML]{8FCFBC}35.0 \\
Claude-Sonnet-4 & \cellcolor[HTML]{62B4C7}\textbf{100.0} & 56.0 & \cellcolor[HTML]{68B8C6}\underline{29.3} & \cellcolor[HTML]{79C3C1}59.7 & \cellcolor[HTML]{ADDCB6}72.0 & \cellcolor[HTML]{7CC5C1}55.3 & \cellcolor[HTML]{BCE2B5}12.5 & \cellcolor[HTML]{BCE2B5}9.1 & \cellcolor[HTML]{62B4C7}\textbf{40.9} \\
LLaMA-3.3-70B & \cellcolor[HTML]{6BBAC5}99.6 & 65.2 & \cellcolor[HTML]{76C1C2}28.7 & \cellcolor[HTML]{8FCFBC}58.3 & \cellcolor[HTML]{93D1BB}76.1 & \cellcolor[HTML]{A2D7B8}50.6 & \cellcolor[HTML]{E6F4BD}10.2 & \cellcolor[HTML]{62B4C7}\textbf{6.2} & \cellcolor[HTML]{B5DFB4}31.0 \\
LLaMA-3.1-70B & \cellcolor[HTML]{FFFFE0}93.1 & 62.7 & \cellcolor[HTML]{CBE8B6}25.9 & \cellcolor[HTML]{E4F3BA}53.4 & \cellcolor[HTML]{B5DFB4}70.5 & \cellcolor[HTML]{EBF6C3}42.2 & \cellcolor[HTML]{E9F5C0}9.9 & \cellcolor[HTML]{9BD4BA}8.1 & \cellcolor[HTML]{F4FAD0}22.7 \\
Qwen2.5-72B & \cellcolor[HTML]{62B4C7}\textbf{100.0} & 53.8 & \cellcolor[HTML]{8FCFBC}27.8 & \cellcolor[HTML]{9BD4BA}57.7 & \cellcolor[HTML]{B9E1B4}69.7 & \cellcolor[HTML]{B9E1B4}48.3 & \cellcolor[HTML]{8CCDBD}15.2 & \cellcolor[HTML]{65B6C6}\underline{6.3} & \cellcolor[HTML]{CBE8B6}28.3 \\
GPT-4o-mini & \cellcolor[HTML]{62B4C7}\textbf{100.0} & 53.7 & \cellcolor[HTML]{79C3C1}28.6 & \cellcolor[HTML]{97D2BA}57.9 & \cellcolor[HTML]{8CCDBD}77.7 & \cellcolor[HTML]{8CCDBD}53.3 & \cellcolor[HTML]{BCE2B5}12.6 & \cellcolor[HTML]{68B8C6}6.5 & \cellcolor[HTML]{B1DDB5}31.3 \\
Qwen2.5-7B & \cellcolor[HTML]{65B6C6}99.9 & 58.0 & \cellcolor[HTML]{B1DDB5}26.7 & \cellcolor[HTML]{DAEEB7}54.2 & \cellcolor[HTML]{DAEEB7}63.5 & \cellcolor[HTML]{F2F9CD}40.9 & \cellcolor[HTML]{EBF6C3}9.7 & \cellcolor[HTML]{93D1BB}8.0 & \cellcolor[HTML]{FFFFE0}20.7 \\
LLaMA-3.1-8B & \cellcolor[HTML]{7CC5C1}98.7 & 92.0 & \cellcolor[HTML]{E9F5C0}24.8 & \cellcolor[HTML]{F8FCD6}51.7 & \cellcolor[HTML]{FFFFE0}53.9 & \cellcolor[HTML]{FFFFE0}38.3 & \cellcolor[HTML]{FFFFE0}8.0 & \cellcolor[HTML]{93D1BB}7.9 & \cellcolor[HTML]{EFF8CA}23.4 \\
Phi-4-mini & \cellcolor[HTML]{6EBBC4}99.4 & 134.1 & \cellcolor[HTML]{FFFFE0}23.7 & \cellcolor[HTML]{FFFFE0}51.0 & \cellcolor[HTML]{DAEEB7}64.0 & \cellcolor[HTML]{DDF0B8}44.0 & \cellcolor[HTML]{CFEAB6}11.5 & \cellcolor[HTML]{AADAB6}8.6 & \cellcolor[HTML]{B1DDB5}31.1 \\
\bottomrule
\end{tabular}
\end{adjustbox}
\caption{\textbf{Rationale quality by judge and condition}, averaged over six languages and the criteria. \textsc{Expl}: verdicts with a rationale (\%); \textsc{Len}: mean rationale length (tokens); other metrics as in \autoref{tab:explanation_quality}.}
\label{tab:explanation_by_condition}
\end{table}

\clearpage
\newpage
Across the six NLI languages, SC ranges from 67.1\% (Spanish) to 72.7\% (English; see \autoref{tab:explanation_by_language}), falling 10-17 points below the corresponding human reference in every language. Support and coverage vary more: support is highest in Hindi (26.9) and Spanish (21.8), coverage lowest in French (20.6), and contradiction highest in Urdu (15.3). Yoruba's outlying NLI scores are consistent with its unreliable verdict recovery.

\begin{table}[!t]
\centering
\definecolor{negshade}{HTML}{2A78D6}\definecolor{posshade}{HTML}{E34948}
\small
\setlength{\tabcolsep}{1.5pt}
\begin{adjustbox}{max width=\columnwidth}
\begin{tabular}{@{}l *{9}{r} c r@{}}
\toprule
& \multicolumn{2}{c}{\textbf{Presence}} & \multicolumn{2}{c}{\textbf{Similarity}} & \multicolumn{2}{c}{\textbf{Verdict (NLI)}} & \multicolumn{3}{c}{\textbf{Reasoning (NLI)}} && \textbf{Human} \\
\cmidrule(lr){2-3}\cmidrule(lr){4-5}\cmidrule(lr){6-7}\cmidrule(lr){8-10}\cmidrule(lr){12-12}
\textbf{Language} & \textbf{Expl} & \textbf{Len} & \textbf{chrF++} & \textbf{LaBSE} & \textbf{SC} & \textbf{HA} & \textbf{Sup} & \textbf{Con} & \textbf{Cov} && \textbf{Ref} \\
\midrule
English & \cellcolor[HTML]{62B4C7}98.8 & 41.6 & \cellcolor[HTML]{85CABE}28.6 & \cellcolor[HTML]{FFFFE0}55.7 & \cellcolor[HTML]{62B4C7}72.7 & \cellcolor[HTML]{E6F4BD}48.5 & \cellcolor[HTML]{FDFEDD}15.8 & \cellcolor[HTML]{6EBBC4}7.6 & \cellcolor[HTML]{76C1C2}31.5 & & 82.8 \\
French & \cellcolor[HTML]{8FCFBC}97.2 & 54.8 & \cellcolor[HTML]{93D1BB}28.0 & \cellcolor[HTML]{C7E7B6}60.5 & \cellcolor[HTML]{D2EBB7}69.0 & \cellcolor[HTML]{ADDCB6}51.2 & \cellcolor[HTML]{FFFFE0}15.5 & \cellcolor[HTML]{73BFC3}7.9 & \cellcolor[HTML]{FFFFE0}20.6 & & 85.9 \\
Hindi & \cellcolor[HTML]{8CCDBD}97.2 & 52.3 & \cellcolor[HTML]{E6F4BD}24.1 & \cellcolor[HTML]{62B4C7}67.8 & \cellcolor[HTML]{CBE8B6}69.2 & \cellcolor[HTML]{62B4C7}55.0 & \cellcolor[HTML]{62B4C7}26.9 & \cellcolor[HTML]{B9E1B4}11.3 & \cellcolor[HTML]{85CABE}30.2 & & 81.3 \\
Norwegian & \cellcolor[HTML]{EFF8CA}94.0 & 46.7 & \cellcolor[HTML]{CFEAB6}25.2 & \cellcolor[HTML]{A6D9B7}62.6 & \cellcolor[HTML]{F6FBD3}67.6 & \cellcolor[HTML]{FFFFE0}46.7 & \cellcolor[HTML]{FBFDDA}15.9 & \cellcolor[HTML]{AADAB6}10.5 & \cellcolor[HTML]{C4E5B5}25.9 & & 79.1 \\
Spanish & \cellcolor[HTML]{93D1BB}97.0 & 48.6 & \cellcolor[HTML]{62B4C7}30.6 & \cellcolor[HTML]{88CCBE}64.8 & \cellcolor[HTML]{FFFFE0}67.1 & \cellcolor[HTML]{A6D9B7}51.5 & \cellcolor[HTML]{ADDCB6}21.8 & \cellcolor[HTML]{62B4C7}6.9 & \cellcolor[HTML]{62B4C7}33.2 & & 80.7 \\
Urdu & \cellcolor[HTML]{FFFFE0}93.3 & 57.2 & \cellcolor[HTML]{FFFFE0}22.3 & \cellcolor[HTML]{C4E5B5}60.8 & \cellcolor[HTML]{82C8BF}71.5 & \cellcolor[HTML]{76C1C2}53.9 & \cellcolor[HTML]{C4E5B5}20.2 & \cellcolor[HTML]{FFFFE0}15.3 & \cellcolor[HTML]{97D2BA}29.0 & & 86.1 \\
Yoruba$^\dagger$ & 83.0 & 86.4 & 16.4 & 43.4 & 38.4 & 34.4 & 33.9 & 32.1 & 37.5 & & 47.5 \\
\bottomrule
\end{tabular}
\end{adjustbox}
\caption{\textbf{Rationale quality by language}, averaged over judges and both rationale conditions. Metrics as in Tables~\ref{tab:explanation_quality} and~\ref{tab:explanation_by_condition}; \textsc{Ref}: class-balanced agreement of human rationales with their labels. $^\dagger$Yoruba NLI scores are shown for completeness only.}
\label{tab:explanation_by_language}
\end{table}

\begin{table}[!t]
\centering
\definecolor{negshade}{HTML}{2A78D6}\definecolor{posshade}{HTML}{E34948}
\small
\setlength{\tabcolsep}{1.5pt}
\begin{adjustbox}{max width=\columnwidth}
\begin{tabular}{@{}l rr r rrr rr r@{}}
\toprule
 & & & & \multicolumn{3}{c}{\textbf{All}} & \multicolumn{2}{c}{\textbf{Wrong}} & \\
\cmidrule(lr){5-7}\cmidrule(lr){8-9}
\textbf{Rationale text} & \textbf{Change} & \textbf{Empty} & \textbf{Prob.} & \textbf{SC} & \textbf{Ref} & \textbf{HA} & \textbf{SC} & \textbf{HA} & $\rho$ \\
\midrule
Original & -- & -- & \cellcolor[HTML]{C0E4B5}44.2 & \cellcolor[HTML]{8CCDBD}72.4 & \cellcolor[HTML]{7CC5C1}82.7 & \cellcolor[HTML]{B5DFB4}50.8 & \cellcolor[HTML]{97D2BA}65.8 & \cellcolor[HTML]{EBF6C3}17.8 & -- \\
$-$ verdict tokens & 2.6 & 0.0 & \cellcolor[HTML]{C0E4B5}44.2 & \cellcolor[HTML]{8CCDBD}72.3 & -- & \cellcolor[HTML]{B5DFB4}50.8 & \cellcolor[HTML]{97D2BA}65.8 & \cellcolor[HTML]{EBF6C3}17.8 & 1.00 \\
\addlinespace[2pt]
Original (retained) & -- & -- & \cellcolor[HTML]{C0E4B5}44.7 & \cellcolor[HTML]{8FCFBC}70.7 & -- & \cellcolor[HTML]{B5DFB4}49.8 & \cellcolor[HTML]{9BD4BA}64.4 & \cellcolor[HTML]{EBF6C3}18.7 & -- \\
$-$ preference wording & 28.3 & 6.5 & \cellcolor[HTML]{D2EBB7}34.0 & \cellcolor[HTML]{9BD4BA}64.3 & \cellcolor[HTML]{8CCDBD}71.1 & \cellcolor[HTML]{BCE2B5}46.8 & \cellcolor[HTML]{A2D7B8}59.0 & \cellcolor[HTML]{E6F4BD}22.5 & 0.84 \\
\bottomrule
\end{tabular}
\end{adjustbox}
\caption{\textbf{Verdict recovery after masking} (\%; Yoruba excluded). $-$ verdict tokens: verdict labels and lines removed; $-$ preference wording: also every sentence with a preference or equality term; Original (retained): unmasked rationales left non-empty by the second mask. \textsc{Change}/\textsc{Empty}: share of rationales changed/left with fewer than three tokens (excluded); \textsc{Prob.}: share with highest entailment $\geq 0.5$; \textsc{Ref}: class-balanced agreement of masked human rationales with their labels; \textsc{Wrong}: wrong verdicts; $\rho$: Spearman correlation of masked and unmasked SC judge rankings.}
\label{tab:explanation_masking}
\end{table}

\paragraph{Verdict masking:}
\label{app:masking}
We apply two masks to judge and human rationales. The first removes verdict tokens and lines, including labeled verdicts (``Evaluation: B''), standalone verdict lines, and verdict tokens after the final sentence. The reported results use an aggressive mask that additionally removes every sentence containing a preference or equality term, such as \emph{better}, \emph{preferred}, \emph{equally}, or \emph{tie}, using Google Translate for term translations. This can remove reasoning clues (``better sentence structure'') while retaining comparisons without flagged terms (``Response B is more natural''). Removing an opening verdict sentence can leave near-zero entailment for all hypotheses (see \autoref{fig:nli_example}).

\begin{figure}[!ht]
\centering
\fontsize{10}{11.5}\selectfont
\tcbset{colback=white, colframe=black!35, boxrule=0.4pt, arc=1pt, left=3pt, right=3pt, top=1pt, bottom=1pt, toptitle=0.5pt, bottomtitle=0.5pt,
fonttitle=\bfseries, coltitle=black, colbacktitle=black!6}
\begin{tcolorbox}[title={Human rationale (label: A)}]
Response A is the best overall. It provides a richer and more detailed overview of the USPs of a SaaS survey tool, while Response B is good but lacks some of the depth and variety found in Response A.
\end{tcolorbox}
\vspace{-4pt}
\begin{tcolorbox}[title={Judge rationale (\texttt{Gemini-2.5-Pro}, verdict: B)}]
\textcolor{black!45}{Response B is better because it more accurately addresses the concept of a ``Unique Selling Point'' (USP) by focusing on features that differentiate one SaaS survey tool from its competitors.} Response A conflates general benefits of the SaaS model (like accessibility and automatic updates) with actual USPs, making its list less precise.
\end{tcolorbox}
\vspace{-4pt}
\begin{tcolorbox}[title={NLI outputs}]
\setlength{\tabcolsep}{3pt}
\begin{tabular}{@{}l ccc l@{}}
& \multicolumn{3}{c}{Entailment of hypothesis} & \\
\cmidrule(lr){2-4}
Judge rationale & A & B & Tie & Implied \\
\midrule
Original & 0.00 & 0.99 & 0.00 & B \\
Masked & 0.00 & 0.00 & 0.00 & A \\
\end{tabular}

\vspace{1pt}
Original: the implied verdict matches the judge (SC), but not the human label (HA). Masked: A is selected by the argmax rule despite near-zero entailment probabilities for all hypotheses.
\end{tcolorbox}
\caption{\textbf{Verdict wording and NLI recovery.} One English comparison (overall quality, \ccritr). The mask removes the gray sentence. Probabilities are rounded to two decimals; the highest-scoring hypothesis determines the implied verdict even below 0.5.}
\label{fig:nli_example}
\end{figure}

Token removal barely changes scores. Across the six NLI languages, the aggressive mask changes 28.3\% of rationales and leaves 6.5\% shorter than three tokens, which we exclude. We compare original and masked scores on the retained rationales (SC 70.7\% vs.\ 64.3\%; see \autoref{tab:explanation_masking}). The masked human reference is 71.1\%; on wrong verdicts, masked SC and HA are 59.0\% and 22.5\%, respectively. The original and masked judge rankings by SC have a Spearman correlation of 0.84.

\subsection{Metrics and Statistical Analysis}
\label{app:metrics}

\paragraph{Agreement and judge ensembles:}
For each language-criterion combination, we compute Cohen's $\kappa$ between judges and between each judge and the human labels, using comparisons where both verdicts are valid. We average over pairs, then equally over languages, then over criteria. Family comparisons contrast within-family pairs with all other pairs after averaging each pair's $\kappa$ over languages and criteria. 

Majority-vote ensembles select the most frequent valid verdict per comparison, breaking ties with the best single judge, \texttt{Gemini-2.5-Pro}, and use the same macro-F1 scoring as single judges. Shared-error analyses use the five judges with the highest macro-F1; unanimous errors occur when all five give the same valid verdict and disagree with the human label.

\begin{table}[t]
\centering
\definecolor{negshade}{HTML}{2A78D6}\definecolor{posshade}{HTML}{E34948}
\small
\setlength{\tabcolsep}{4pt}
\begin{adjustbox}{max width=\columnwidth}
\begin{tabular}{@{}l r r r@{}}
\toprule
\textbf{Subset} & \textbf{Macro-F1} & \textbf{Accuracy} & \shortstack{\textbf{Accuracy}\\\textbf{(valid only)}} \\
\midrule
\multicolumn{4}{@{}l}{\cjointr $-$ \cjoint (+ rationale, five criteria per call)} \\
All judges & \cellcolor{posshade!10}+1.4 & \cellcolor{posshade!18}+2.4 & \cellcolor{posshade!16}+2.1 \\
Prompt-following cells & \cellcolor{posshade!18}+2.4 & \cellcolor{posshade!29}+3.9 & \cellcolor{posshade!30}+3.9 \\
Reliable judges & +0.0 & \cellcolor{negshade!1}$-$0.1 & \cellcolor{posshade!1}+0.1 \\
\midrule
\multicolumn{4}{@{}l}{\ccritr $-$ \ccrit (+ rationale, one criterion per call)} \\
All judges & \cellcolor{negshade!11}$-$1.5 & \cellcolor{negshade!17}$-$2.3 & \cellcolor{negshade!20}$-$2.7 \\
Prompt-following cells & \cellcolor{negshade!10}$-$1.4 & \cellcolor{negshade!18}$-$2.4 & \cellcolor{negshade!18}$-$2.4 \\
Reliable judges & \cellcolor{negshade!15}$-$2.0 & \cellcolor{negshade!30}$-$4.0 & \cellcolor{negshade!27}$-$3.6 \\
\midrule
\multicolumn{4}{@{}l}{\ccrit $-$ \cjoint (one vs five criteria, no rationale)} \\
All judges & \cellcolor{negshade!1}$-$0.2 & \cellcolor{negshade!2}$-$0.3 & \cellcolor{negshade!4}$-$0.6 \\
Prompt-following cells & \cellcolor{posshade!4}+0.5 & \cellcolor{negshade!2}$-$0.2 & +0.0 \\
Reliable judges & \cellcolor{negshade!2}$-$0.2 & \cellcolor{negshade!1}$-$0.2 & \cellcolor{negshade!3}$-$0.4 \\
\midrule
\multicolumn{4}{@{}l}{\ccritr $-$ \cjointr (one vs five criteria, with rationale)} \\
All judges & \cellcolor{negshade!22}$-$3.0 & \cellcolor{negshade!37}$-$4.9 & \cellcolor{negshade!41}$-$5.4 \\
Prompt-following cells & \cellcolor{negshade!26}$-$3.4 & \cellcolor{negshade!43}$-$5.7 & \cellcolor{negshade!45}$-$6.0 \\
Reliable judges & \cellcolor{negshade!17}$-$2.3 & \cellcolor{negshade!30}$-$4.0 & \cellcolor{negshade!31}$-$4.1 \\
\bottomrule
\end{tabular}
\end{adjustbox}
\caption{\textbf{Condition effects across metrics and judge subsets.} Changes in points. Prompt-following cells drop a judge in a language with more than 15\% invalid verdicts in either condition; reliable judges are the seven judges with a valid-verdict rate of at least 90\% in all four conditions.}
\label{tab:explanation_robustness}
\end{table}

\paragraph{Condition contrasts and intervals:}
Condition contrasts are paired by judge, language, criterion, and comparison. We average differences equally over languages, then over judges and criteria. The 95\% intervals use 1,000 bootstrap resamples of instructions within each language. We repeat the contrasts on two subsets (see \autoref{tab:explanation_robustness}). \emph{Prompt-following} excludes judge-language combinations with more than 15\% invalid verdicts in either contrasted condition. \emph{Reliable judges} retain seven judges with at least 90\% valid verdicts in all conditions, excluding \texttt{LLaMA-3.1-70B} and \texttt{Qwen2.5-72B} together with the three low-validity judges.
\FloatBarrier
\section{Results per Prompt Variant}
\label{app:prompt_variants}
\begin{table}[h]
\centering
\small
\setlength{\tabcolsep}{3.5pt}
\definecolor{negshade}{HTML}{2A78D6}\definecolor{posshade}{HTML}{E34948}
\begin{tabularx}{\columnwidth}{@{}l>{\centering\arraybackslash}X>{\centering\arraybackslash}X>{\raggedleft\arraybackslash}X@{}}
\toprule
& \shortstack{\textbf{Kendall}\\$\tau_b$}
& \shortstack{\textbf{Spearman}\\$\rho$}
& \shortstack{\textbf{Top-1}\\\textbf{stable}} \\
\midrule
Relevance & \cellcolor[HTML]{93D1BB}0.63 & \cellcolor[HTML]{8CCDBD}0.77 & 14\% \\
Naturalness & \cellcolor[HTML]{FFFFE0}0.46 & \cellcolor[HTML]{FFFFE0}0.59 & 0\% \\
Truthfulness & \cellcolor[HTML]{6EBBC4}0.69 & \cellcolor[HTML]{6EBBC4}0.82 & 14\% \\
Safety & \cellcolor[HTML]{85CABE}0.65 & \cellcolor[HTML]{79C3C1}0.80 & 29\% \\
Overall quality & \cellcolor[HTML]{DDF0B8}0.53 & \cellcolor[HTML]{C0E4B5}0.70 & 0\% \\
\midrule
English & \cellcolor[HTML]{FBFDDA}0.47 & \cellcolor[HTML]{FFFFE0}0.59 & 20\% \\
Urdu & \cellcolor[HTML]{8CCDBD}0.64 & \cellcolor[HTML]{7FC6C0}0.79 & 0\% \\
Norwegian & \cellcolor[HTML]{AADAB6}0.60 & \cellcolor[HTML]{8CCDBD}0.77 & 0\% \\
Hindi & \cellcolor[HTML]{85CABE}0.65 & \cellcolor[HTML]{7FC6C0}0.79 & 20\% \\
Yoruba & \cellcolor[HTML]{62B4C7}0.71 & \cellcolor[HTML]{62B4C7}0.84 & 40\% \\
Spanish & \cellcolor[HTML]{C7E7B6}0.56 & \cellcolor[HTML]{AADAB6}0.73 & 0\% \\
French & \cellcolor[HTML]{E9F5C0}0.51 & \cellcolor[HTML]{E9F5C0}0.64 & 0\% \\
\midrule
\textbf{Aggregate} & \cellcolor[HTML]{B1DDB5}\textbf{0.59} & \cellcolor[HTML]{A2D7B8}\textbf{0.74} & \textbf{11\%} \\
\bottomrule
\end{tabularx}
\caption{\textbf{Stability of judge rankings} across prompts P0-P2 by criterion and language (judges ranked by macro-F1). Top-1 stability: share of cases in which the same judge ranks first under all prompts.}
\label{tab:ranking_stability}
\end{table}
\begin{table*}[p]
\centering
\definecolor{negshade}{HTML}{2A78D6}\definecolor{posshade}{HTML}{E34948}
\resizebox{\textwidth}{!}{%
\begin{tabular}{l *{8}{r} c *{8}{r} c *{8}{r} c r}
\toprule
\multirow{2}{*}{\textbf{Judge}} &\multicolumn{8}{c}{\textbf{P0}} &&\multicolumn{8}{c}{\textbf{P1}} &&\multicolumn{8}{c}{\textbf{P2}} && \multirow{2}{*}{$\mathbf{\Delta}$} \\
\cmidrule(lr){2-9}\cmidrule(lr){11-18}\cmidrule(lr){20-27}
& \textbf{R} & \textbf{N} & \textbf{T} & \textbf{S} & \textbf{O} & \textbf{F1} & \textbf{Borda} & \textbf{Valid} && \textbf{R} & \textbf{N} & \textbf{T} & \textbf{S} & \textbf{O} & \textbf{F1} & \textbf{Borda} & \textbf{Valid} && \textbf{R} & \textbf{N} & \textbf{T} & \textbf{S} & \textbf{O} & \textbf{F1} & \textbf{Borda} & \textbf{Valid} && \\
\midrule
Gemini-2.5-Pro & \cellcolor[HTML]{97D2BA}39.9 & \cellcolor[HTML]{A2D7B8}35.5 & \cellcolor[HTML]{9BD4BA}39.0 & \cellcolor[HTML]{82C8BF}47.3 & \cellcolor[HTML]{97D2BA}40.1 & \cellcolor[HTML]{93D1BB}40.3 & 81.8 & 100.0 & & \cellcolor[HTML]{88CCBE}\textbf{44.3} & \cellcolor[HTML]{9BD4BA}\textbf{38.5} & \cellcolor[HTML]{97D2BA}\textbf{40.1} & \cellcolor[HTML]{7FC6C0}\textbf{47.8} & \cellcolor[HTML]{8FCFBC}\textbf{42.0} & \cellcolor[HTML]{8FCFBC}\textbf{42.5} & 80.0 & 100.0 & & \cellcolor[HTML]{8FCFBC}41.8 & \cellcolor[HTML]{9ED5B9}36.9 & \cellcolor[HTML]{9BD4BA}39.0 & \cellcolor[HTML]{85CABE}45.4 & \cellcolor[HTML]{97D2BA}39.6 & \cellcolor[HTML]{93D1BB}40.5 & 70.9 & 99.9 & & 2.2 \\
Qwen3-32B & \cellcolor[HTML]{8FCFBC}42.4 & \cellcolor[HTML]{A6D9B7}34.2 & \cellcolor[HTML]{A6D9B7}34.5 & \cellcolor[HTML]{9ED5B9}37.5 & \cellcolor[HTML]{9ED5B9}37.1 & \cellcolor[HTML]{9ED5B9}37.2 & 61.8 & 99.2 & & \cellcolor[HTML]{82C8BF}\textbf{46.7} & \cellcolor[HTML]{A2D7B8}35.9 & \cellcolor[HTML]{93D1BB}\textbf{40.9} & \cellcolor[HTML]{93D1BB}40.5 & \cellcolor[HTML]{8CCDBD}\textbf{43.3} & \cellcolor[HTML]{8FCFBC}\textbf{41.5} & 85.5 & 99.2 & & \cellcolor[HTML]{85CABE}45.9 & \cellcolor[HTML]{A2D7B8}\textbf{36.2} & \cellcolor[HTML]{9BD4BA}38.4 & \cellcolor[HTML]{93D1BB}\textbf{41.3} & \cellcolor[HTML]{8FCFBC}42.1 & \cellcolor[HTML]{93D1BB}40.8 & 83.6 & 98.9 & & 4.3 \\
\bottomrule
\end{tabular}%
}
\caption{\textbf{English} per-prompt \cjoint results (selected: P1).}
\label{tab:prompt_variants_en}
\end{table*}

\begin{table*}[p]
\centering
\definecolor{negshade}{HTML}{2A78D6}\definecolor{posshade}{HTML}{E34948}
\resizebox{\textwidth}{!}{%
\begin{tabular}{l *{8}{r} c *{8}{r} c *{8}{r} c r}
\toprule
\multirow{2}{*}{\textbf{Judge}} &\multicolumn{8}{c}{\textbf{P0}} &&\multicolumn{8}{c}{\textbf{P1}} &&\multicolumn{8}{c}{\textbf{P2}} && \multirow{2}{*}{$\mathbf{\Delta}$} \\
\cmidrule(lr){2-9}\cmidrule(lr){11-18}\cmidrule(lr){20-27}
& \textbf{R} & \textbf{N} & \textbf{T} & \textbf{S} & \textbf{O} & \textbf{F1} & \textbf{Borda} & \textbf{Valid} && \textbf{R} & \textbf{N} & \textbf{T} & \textbf{S} & \textbf{O} & \textbf{F1} & \textbf{Borda} & \textbf{Valid} && \textbf{R} & \textbf{N} & \textbf{T} & \textbf{S} & \textbf{O} & \textbf{F1} & \textbf{Borda} & \textbf{Valid} && \\
\midrule
Gemini-2.5-Pro & \cellcolor[HTML]{65B6C6}\textbf{58.9} & \cellcolor[HTML]{E9F5C0}12.0 & \cellcolor[HTML]{9ED5B9}36.8 & \cellcolor[HTML]{82C8BF}47.4 & \cellcolor[HTML]{73BFC3}52.6 & \cellcolor[HTML]{8FCFBC}41.6 & 78.2 & 100.0 & & \cellcolor[HTML]{7FC6C0}47.6 & \cellcolor[HTML]{B9E1B4}\textbf{28.4} & \cellcolor[HTML]{9ED5B9}\textbf{37.0} & \cellcolor[HTML]{79C3C1}\textbf{49.8} & \cellcolor[HTML]{6EBBC4}\textbf{55.7} & \cellcolor[HTML]{8CCDBD}\textbf{43.7} & 83.6 & 100.0 & & \cellcolor[HTML]{65B6C6}58.5 & \cellcolor[HTML]{E9F5C0}12.1 & \cellcolor[HTML]{A2D7B8}36.4 & \cellcolor[HTML]{82C8BF}47.4 & \cellcolor[HTML]{73BFC3}52.9 & \cellcolor[HTML]{8FCFBC}41.5 & 81.8 & 100.0 & & 2.2 \\
\bottomrule
\end{tabular}%
}
\caption{\textbf{French} per-prompt \cjoint results (selected: P0).}
\label{tab:prompt_variants_fra}
\end{table*}

\begin{table*}[p]
\centering
\definecolor{negshade}{HTML}{2A78D6}\definecolor{posshade}{HTML}{E34948}
\resizebox{\textwidth}{!}{%
\begin{tabular}{l *{8}{r} c *{8}{r} c *{8}{r} c r}
\toprule
\multirow{2}{*}{\textbf{Judge}} &\multicolumn{8}{c}{\textbf{P0}} &&\multicolumn{8}{c}{\textbf{P1}} &&\multicolumn{8}{c}{\textbf{P2}} && \multirow{2}{*}{$\mathbf{\Delta}$} \\
\cmidrule(lr){2-9}\cmidrule(lr){11-18}\cmidrule(lr){20-27}
& \textbf{R} & \textbf{N} & \textbf{T} & \textbf{S} & \textbf{O} & \textbf{F1} & \textbf{Borda} & \textbf{Valid} && \textbf{R} & \textbf{N} & \textbf{T} & \textbf{S} & \textbf{O} & \textbf{F1} & \textbf{Borda} & \textbf{Valid} && \textbf{R} & \textbf{N} & \textbf{T} & \textbf{S} & \textbf{O} & \textbf{F1} & \textbf{Borda} & \textbf{Valid} && \\
\midrule
Gemini-2.5-Pro & \cellcolor[HTML]{62B4C7}75.1 & \cellcolor[HTML]{65B6C6}\textbf{58.3} & \cellcolor[HTML]{62B4C7}68.9 & \cellcolor[HTML]{62B4C7}65.5 & \cellcolor[HTML]{62B4C7}59.5 & \cellcolor[HTML]{62B4C7}65.5 & 87.3 & 100.0 & & \cellcolor[HTML]{62B4C7}62.1 & \cellcolor[HTML]{6EBBC4}55.4 & \cellcolor[HTML]{62B4C7}69.9 & \cellcolor[HTML]{73BFC3}53.0 & \cellcolor[HTML]{62B4C7}\textbf{60.2} & \cellcolor[HTML]{62B4C7}60.1 & 65.5 & 100.0 & & \cellcolor[HTML]{62B4C7}\textbf{75.3} & \cellcolor[HTML]{68B8C6}57.8 & \cellcolor[HTML]{62B4C7}\textbf{73.9} & \cellcolor[HTML]{62B4C7}\textbf{65.9} & \cellcolor[HTML]{65B6C6}59.3 & \cellcolor[HTML]{62B4C7}\textbf{66.4} & 85.5 & 100.0 & & 6.3 \\
Qwen3-32B & \cellcolor[HTML]{62B4C7}\textbf{72.4} & \cellcolor[HTML]{6BBAC5}\textbf{56.6} & \cellcolor[HTML]{62B4C7}66.7 & \cellcolor[HTML]{6EBBC4}55.6 & \cellcolor[HTML]{62B4C7}\textbf{72.8} & \cellcolor[HTML]{62B4C7}\textbf{64.8} & 78.2 & 98.5 & & \cellcolor[HTML]{62B4C7}63.2 & \cellcolor[HTML]{70BDC3}53.5 & \cellcolor[HTML]{62B4C7}63.2 & \cellcolor[HTML]{68B8C6}57.4 & \cellcolor[HTML]{68B8C6}58.0 & \cellcolor[HTML]{65B6C6}59.0 & 52.7 & 98.9 & & \cellcolor[HTML]{62B4C7}64.7 & \cellcolor[HTML]{70BDC3}54.5 & \cellcolor[HTML]{62B4C7}\textbf{71.0} & \cellcolor[HTML]{62B4C7}\textbf{60.5} & \cellcolor[HTML]{68B8C6}57.6 & \cellcolor[HTML]{62B4C7}61.7 & 58.2 & 99.6 & & 5.8 \\
\bottomrule
\end{tabular}%
}
\caption{\textbf{Hindi} per-prompt \cjoint results (selected: P0).}
\label{tab:prompt_variants_hi}
\end{table*}

\begin{table*}[p]
\centering
\definecolor{negshade}{HTML}{2A78D6}\definecolor{posshade}{HTML}{E34948}
\resizebox{\textwidth}{!}{%
\begin{tabular}{l *{8}{r} c *{8}{r} c *{8}{r} c r}
\toprule
\multirow{2}{*}{\textbf{Judge}} &\multicolumn{8}{c}{\textbf{P0}} &&\multicolumn{8}{c}{\textbf{P1}} &&\multicolumn{8}{c}{\textbf{P2}} && \multirow{2}{*}{$\mathbf{\Delta}$} \\
\cmidrule(lr){2-9}\cmidrule(lr){11-18}\cmidrule(lr){20-27}
& \textbf{R} & \textbf{N} & \textbf{T} & \textbf{S} & \textbf{O} & \textbf{F1} & \textbf{Borda} & \textbf{Valid} && \textbf{R} & \textbf{N} & \textbf{T} & \textbf{S} & \textbf{O} & \textbf{F1} & \textbf{Borda} & \textbf{Valid} && \textbf{R} & \textbf{N} & \textbf{T} & \textbf{S} & \textbf{O} & \textbf{F1} & \textbf{Borda} & \textbf{Valid} && \\
\midrule
Gemini-2.5-Pro & \cellcolor[HTML]{79C3C1}50.5 & \cellcolor[HTML]{A6D9B7}34.6 & \cellcolor[HTML]{79C3C1}50.3 & \cellcolor[HTML]{A6D9B7}35.3 & \cellcolor[HTML]{88CCBE}44.3 & \cellcolor[HTML]{8CCDBD}43.0 & 80.0 & 100.0 & & \cellcolor[HTML]{70BDC3}\textbf{53.6} & \cellcolor[HTML]{93D1BB}\textbf{40.5} & \cellcolor[HTML]{73BFC3}\textbf{52.8} & \cellcolor[HTML]{B5DFB4}30.5 & \cellcolor[HTML]{88CCBE}\textbf{44.4} & \cellcolor[HTML]{88CCBE}\textbf{44.4} & 83.6 & 100.0 & & \cellcolor[HTML]{7CC5C1}49.1 & \cellcolor[HTML]{9BD4BA}37.9 & \cellcolor[HTML]{73BFC3}52.6 & \cellcolor[HTML]{9BD4BA}\textbf{37.8} & \cellcolor[HTML]{88CCBE}44.1 & \cellcolor[HTML]{88CCBE}44.3 & 92.7 & 100.0 & & 1.4 \\
Qwen3-32B & \cellcolor[HTML]{8FCFBC}42.0 & \cellcolor[HTML]{ADDCB6}32.8 & \cellcolor[HTML]{79C3C1}\textbf{51.0} & \cellcolor[HTML]{B5DFB4}29.7 & \cellcolor[HTML]{85CABE}45.1 & \cellcolor[HTML]{97D2BA}40.1 & 76.4 & 99.3 & & \cellcolor[HTML]{82C8BF}46.5 & \cellcolor[HTML]{97D2BA}\textbf{39.1} & \cellcolor[HTML]{7CC5C1}49.0 & \cellcolor[HTML]{A6D9B7}\textbf{34.9} & \cellcolor[HTML]{7FC6C0}\textbf{48.3} & \cellcolor[HTML]{8CCDBD}\textbf{43.6} & 81.8 & 99.8 & & \cellcolor[HTML]{7FC6C0}\textbf{47.9} & \cellcolor[HTML]{ADDCB6}32.7 & \cellcolor[HTML]{82C8BF}46.9 & \cellcolor[HTML]{ADDCB6}32.0 & \cellcolor[HTML]{8CCDBD}43.0 & \cellcolor[HTML]{93D1BB}40.5 & 74.5 & 98.7 & & 3.4 \\
\bottomrule
\end{tabular}%
}
\caption{\textbf{Norwegian} per-prompt \cjoint results (selected: P1).}
\label{tab:prompt_variants_nor}
\end{table*}

\begin{table*}[p]
\centering
\definecolor{negshade}{HTML}{2A78D6}\definecolor{posshade}{HTML}{E34948}
\resizebox{\textwidth}{!}{%
\begin{tabular}{l *{8}{r} c *{8}{r} c *{8}{r} c r}
\toprule
\multirow{2}{*}{\textbf{Judge}} &\multicolumn{8}{c}{\textbf{P0}} &&\multicolumn{8}{c}{\textbf{P1}} &&\multicolumn{8}{c}{\textbf{P2}} && \multirow{2}{*}{$\mathbf{\Delta}$} \\
\cmidrule(lr){2-9}\cmidrule(lr){11-18}\cmidrule(lr){20-27}
& \textbf{R} & \textbf{N} & \textbf{T} & \textbf{S} & \textbf{O} & \textbf{F1} & \textbf{Borda} & \textbf{Valid} && \textbf{R} & \textbf{N} & \textbf{T} & \textbf{S} & \textbf{O} & \textbf{F1} & \textbf{Borda} & \textbf{Valid} && \textbf{R} & \textbf{N} & \textbf{T} & \textbf{S} & \textbf{O} & \textbf{F1} & \textbf{Borda} & \textbf{Valid} && \\
\midrule
Gemini-2.5-Pro & \cellcolor[HTML]{62B4C7}\textbf{60.3} & \cellcolor[HTML]{88CCBE}\textbf{44.2} & \cellcolor[HTML]{7CC5C1}49.2 & \cellcolor[HTML]{85CABE}45.9 & \cellcolor[HTML]{82C8BF}\textbf{47.1} & \cellcolor[HTML]{7CC5C1}\textbf{49.3} & 94.5 & 100.0 & & \cellcolor[HTML]{7CC5C1}49.6 & \cellcolor[HTML]{8FCFBC}42.1 & \cellcolor[HTML]{73BFC3}\textbf{52.8} & \cellcolor[HTML]{7CC5C1}\textbf{49.1} & \cellcolor[HTML]{82C8BF}47.0 & \cellcolor[HTML]{7FC6C0}48.1 & 92.7 & 100.0 & & \cellcolor[HTML]{6BBAC5}56.4 & \cellcolor[HTML]{AADAB6}33.4 & \cellcolor[HTML]{76C1C2}51.7 & \cellcolor[HTML]{7FC6C0}47.6 & \cellcolor[HTML]{85CABE}45.3 & \cellcolor[HTML]{82C8BF}46.8 & 83.6 & 100.0 & & 2.5 \\
GPT-5 & \cellcolor[HTML]{82C8BF}47.3 & \cellcolor[HTML]{8CCDBD}\textbf{42.9} & \cellcolor[HTML]{9BD4BA}38.7 & \cellcolor[HTML]{85CABE}45.2 & \cellcolor[HTML]{88CCBE}43.9 & \cellcolor[HTML]{8CCDBD}43.6 & 76.4 & 100.0 & & \cellcolor[HTML]{82C8BF}47.4 & \cellcolor[HTML]{9BD4BA}38.9 & \cellcolor[HTML]{79C3C1}\textbf{50.5} & \cellcolor[HTML]{7FC6C0}\textbf{48.2} & \cellcolor[HTML]{70BDC3}53.7 & \cellcolor[HTML]{7FC6C0}\textbf{47.7} & 81.8 & 100.0 & & \cellcolor[HTML]{7CC5C1}\textbf{48.7} & \cellcolor[HTML]{97D2BA}39.4 & \cellcolor[HTML]{82C8BF}46.5 & \cellcolor[HTML]{85CABE}45.9 & \cellcolor[HTML]{6BBAC5}\textbf{56.1} & \cellcolor[HTML]{82C8BF}47.3 & 80.0 & 100.0 & & 4.1 \\
\bottomrule
\end{tabular}%
}
\caption{\textbf{Spanish} per-prompt \cjoint results (selected: P1).}
\label{tab:prompt_variants_spa}
\end{table*}

\begin{table*}[t]
\centering
\definecolor{negshade}{HTML}{2A78D6}\definecolor{posshade}{HTML}{E34948}
\resizebox{\textwidth}{!}{%
\begin{tabular}{l *{8}{r} c *{8}{r} c *{8}{r} c r}
\toprule
\multirow{2}{*}{\textbf{Judge}} &\multicolumn{8}{c}{\textbf{P0}} &&\multicolumn{8}{c}{\textbf{P1}} &&\multicolumn{8}{c}{\textbf{P2}} && \multirow{2}{*}{$\mathbf{\Delta}$} \\
\cmidrule(lr){2-9}\cmidrule(lr){11-18}\cmidrule(lr){20-27}
& \textbf{R} & \textbf{N} & \textbf{T} & \textbf{S} & \textbf{O} & \textbf{F1} & \textbf{Borda} & \textbf{Valid} && \textbf{R} & \textbf{N} & \textbf{T} & \textbf{S} & \textbf{O} & \textbf{F1} & \textbf{Borda} & \textbf{Valid} && \textbf{R} & \textbf{N} & \textbf{T} & \textbf{S} & \textbf{O} & \textbf{F1} & \textbf{Borda} & \textbf{Valid} && \\
\midrule
Qwen3-32B & \cellcolor[HTML]{73BFC3}53.2 & \cellcolor[HTML]{8CCDBD}43.8 & \cellcolor[HTML]{62B4C7}62.8 & \cellcolor[HTML]{B1DDB5}30.7 & \cellcolor[HTML]{7FC6C0}48.1 & \cellcolor[HTML]{7FC6C0}47.7 & 78.2 & 98.0 & & \cellcolor[HTML]{62B4C7}\textbf{66.6} & \cellcolor[HTML]{88CCBE}44.2 & \cellcolor[HTML]{62B4C7}\textbf{70.0} & \cellcolor[HTML]{88CCBE}44.3 & \cellcolor[HTML]{7CC5C1}49.2 & \cellcolor[HTML]{6EBBC4}\textbf{54.8} & 87.3 & 98.0 & & \cellcolor[HTML]{68B8C6}57.7 & \cellcolor[HTML]{7FC6C0}\textbf{48.0} & \cellcolor[HTML]{62B4C7}64.4 & \cellcolor[HTML]{79C3C1}\textbf{50.0} & \cellcolor[HTML]{7CC5C1}\textbf{49.6} & \cellcolor[HTML]{70BDC3}53.9 & 89.1 & 98.8 & & 7.1 \\
Gemini-2.5-Pro & \cellcolor[HTML]{79C3C1}50.2 & \cellcolor[HTML]{8CCDBD}43.3 & \cellcolor[HTML]{62B4C7}68.2 & \cellcolor[HTML]{8CCDBD}43.0 & \cellcolor[HTML]{7FC6C0}\textbf{48.2} & \cellcolor[HTML]{79C3C1}50.6 & 81.8 & 99.5 & & \cellcolor[HTML]{62B4C7}\textbf{63.4} & \cellcolor[HTML]{88CCBE}\textbf{44.6} & \cellcolor[HTML]{62B4C7}\textbf{69.7} & \cellcolor[HTML]{85CABE}45.4 & \cellcolor[HTML]{7FC6C0}47.8 & \cellcolor[HTML]{70BDC3}\textbf{54.2} & 81.8 & 100.0 & & \cellcolor[HTML]{70BDC3}53.9 & \cellcolor[HTML]{8CCDBD}43.2 & \cellcolor[HTML]{62B4C7}69.4 & \cellcolor[HTML]{82C8BF}\textbf{47.4} & \cellcolor[HTML]{82C8BF}46.7 & \cellcolor[HTML]{76C1C2}52.1 & 65.5 & 100.0 & & 3.6 \\
\bottomrule
\end{tabular}%
}
\caption{\textbf{Urdu} per-prompt \cjoint results (selected: P1).}
\label{tab:prompt_variants_ur}
\end{table*}

\begin{table*}[t]
\centering
\definecolor{negshade}{HTML}{2A78D6}\definecolor{posshade}{HTML}{E34948}
\resizebox{\textwidth}{!}{%
\begin{tabular}{l *{8}{r} c *{8}{r} c *{8}{r} c r}
\toprule
\multirow{2}{*}{\textbf{Judge}} &\multicolumn{8}{c}{\textbf{P0}} &&\multicolumn{8}{c}{\textbf{P1}} &&\multicolumn{8}{c}{\textbf{P2}} && \multirow{2}{*}{$\mathbf{\Delta}$} \\
\cmidrule(lr){2-9}\cmidrule(lr){11-18}\cmidrule(lr){20-27}
& \textbf{R} & \textbf{N} & \textbf{T} & \textbf{S} & \textbf{O} & \textbf{F1} & \textbf{Borda} & \textbf{Valid} && \textbf{R} & \textbf{N} & \textbf{T} & \textbf{S} & \textbf{O} & \textbf{F1} & \textbf{Borda} & \textbf{Valid} && \textbf{R} & \textbf{N} & \textbf{T} & \textbf{S} & \textbf{O} & \textbf{F1} & \textbf{Borda} & \textbf{Valid} && \\
\midrule
Gemini-2.5-Pro & \cellcolor[HTML]{65B6C6}59.2 & \cellcolor[HTML]{7CC5C1}\textbf{49.2} & \cellcolor[HTML]{73BFC3}\textbf{53.2} & \cellcolor[HTML]{AADAB6}33.3 & \cellcolor[HTML]{73BFC3}\textbf{52.3} & \cellcolor[HTML]{7CC5C1}49.4 & 96.4 & 99.6 & & \cellcolor[HTML]{62B4C7}\textbf{64.6} & \cellcolor[HTML]{85CABE}45.5 & \cellcolor[HTML]{79C3C1}51.0 & \cellcolor[HTML]{9ED5B9}36.7 & \cellcolor[HTML]{79C3C1}50.4 & \cellcolor[HTML]{7CC5C1}\textbf{49.7} & 94.5 & 100.0 & & \cellcolor[HTML]{62B4C7}59.8 & \cellcolor[HTML]{93D1BB}40.9 & \cellcolor[HTML]{7FC6C0}48.0 & \cellcolor[HTML]{93D1BB}\textbf{40.9} & \cellcolor[HTML]{7FC6C0}47.6 & \cellcolor[HTML]{82C8BF}47.4 & 85.5 & 99.9 & & 2.2 \\
\bottomrule
\end{tabular}%
}
\caption{\textbf{Yoruba} per-prompt \cjoint results (selected: P1).}
\label{tab:prompt_variants_yo}
\end{table*}

\autoref{tab:ranking_stability} shows how stable judge rankings are across prompts (see \S\ref{subsec:res_prediction}). Tables~\ref{tab:prompt_variants_en}-\ref{tab:prompt_variants_yo} report \cjoint under P0-P2. R, N, T, S, and O give macro-F1 (\%) for relevance, naturalness, truthfulness, safety, and overall quality; F1 is their average. 

Borda summarizes judge rankings over the five criteria on a 0--100 scale. Valid is the valid-verdict rate (valid / expected verdicts, \%). Bold marks the highest macro-F1 across prompts for each judge-criterion pair and for mean F1, while $\Delta$ shows the difference between the highest and lowest mean F1 across prompts. Judges are sorted by mean F1 under the respective selected prompt.

\paragraph{Prompt effects by language:}
Averaged over judges and criteria, the best and worst prompts differ by 1.0-1.5 macro-F1 points in English, Spanish, and Norwegian, by 1.8-3.0 points in Yoruba, Hindi, and French, and by 4.2 points in Urdu, where P0 trails P1 and P2 (40.1 vs.\ 44.3). Thus, sensitivity to prompt formulation varies across languages, with the largest gap in Urdu. Yoruba has the most invalid verdicts under every prompt (23-26\% of the expected verdicts), indicating that none of the formulations resolves its high rate of invalid outputs.

\paragraph{Prompt effects by judge:}
The strongest judges are the least sensitive to the prompt: the mean macro-F1 of \texttt{Gemini-2.5-Pro} and \texttt{Claude-Sonnet-4} varies by less than one point across P0-P2. The spread is largest for \texttt{Qwen2.5-7B} (8.4 points; 30.1 under P0 vs.\ 21.7 under P2), followed by \texttt{Phi-4-mini} and \texttt{LLaMA-3.1-70B} (5.7 each) and \texttt{LLaMA-3.3-70B} (5.4); both LLaMA-70B judges score lowest under P0.

\paragraph{Prompt effects by criterion:}
Averaged over judges and languages, the prompt shifts macro-F1 by at most 2.5 points per criterion, but for a single judge and language, the best and worst prompts differ by 5.5-9.6 points on average, and in 36 of these 84 cases, each prompt is best on some criterion.

\begin{table}[t]
\centering
\begin{minipage}[t]{\columnwidth}
\centering
\definecolor{negshade}{HTML}{2A78D6}\definecolor{posshade}{HTML}{E34948}
\small
\setlength{\tabcolsep}{2pt}
\begin{adjustbox}{max width=\columnwidth}
\begin{tabular}{@{}l *{5}{r} c *{4}{r}@{}}
\toprule
& \multicolumn{5}{c}{\textbf{Macro-F1 by criterion}} && \multicolumn{4}{c}{\textbf{Overall}} \\
\cmidrule(lr){2-6}\cmidrule(lr){8-11}
\textbf{Judge} & \textbf{R} & \textbf{N} & \textbf{T} & \textbf{S} & \textbf{O} && \textbf{F1} & \textbf{Acc} & \textbf{Borda} & \textbf{Valid} \\
\midrule
Gemini-2.5-Pro & \cellcolor[HTML]{88CCBE}44.3 & \cellcolor[HTML]{9BD4BA}38.5 & \cellcolor[HTML]{97D2BA}\underline{40.1} & \cellcolor[HTML]{7FC6C0}\textbf{47.8} & \cellcolor[HTML]{8FCFBC}42.0 & & \cellcolor[HTML]{8FCFBC}\textbf{42.5} & 59.3 & \underline{80.0} & \textbf{100.0} \\
Qwen3-32B & \cellcolor[HTML]{82C8BF}\textbf{46.7} & \cellcolor[HTML]{A2D7B8}35.9 & \cellcolor[HTML]{93D1BB}\textbf{40.9} & \cellcolor[HTML]{93D1BB}40.5 & \cellcolor[HTML]{8CCDBD}\underline{43.3} & & \cellcolor[HTML]{8FCFBC}\underline{41.5} & \underline{59.4} & \textbf{85.5} & 99.2 \\
DeepSeek-V3 & \cellcolor[HTML]{85CABE}\underline{46.1} & \cellcolor[HTML]{93D1BB}\underline{40.7} & \cellcolor[HTML]{A6D9B7}34.4 & \cellcolor[HTML]{A6D9B7}35.0 & \cellcolor[HTML]{85CABE}\textbf{46.1} & & \cellcolor[HTML]{93D1BB}40.5 & 57.8 & 76.4 & 99.1 \\
GPT-5 & \cellcolor[HTML]{9BD4BA}38.0 & \cellcolor[HTML]{93D1BB}\textbf{41.1} & \cellcolor[HTML]{9BD4BA}37.9 & \cellcolor[HTML]{97D2BA}39.0 & \cellcolor[HTML]{8CCDBD}43.2 & & \cellcolor[HTML]{97D2BA}39.9 & 58.9 & 70.9 & \textbf{100.0} \\
Claude-Sonnet-4 & \cellcolor[HTML]{85CABE}45.4 & \cellcolor[HTML]{A6D9B7}34.5 & \cellcolor[HTML]{B9E1B4}29.2 & \cellcolor[HTML]{8FCFBC}\underline{42.1} & \cellcolor[HTML]{93D1BB}40.7 & & \cellcolor[HTML]{9BD4BA}38.4 & 52.6 & 67.3 & 97.7 \\
LLaMA-3.3-70B & \cellcolor[HTML]{85CABE}45.3 & \cellcolor[HTML]{BCE2B5}28.1 & \cellcolor[HTML]{9BD4BA}38.4 & \cellcolor[HTML]{9BD4BA}38.7 & \cellcolor[HTML]{93D1BB}40.3 & & \cellcolor[HTML]{9BD4BA}38.2 & \textbf{59.6} & 60.0 & \textbf{100.0} \\
\bottomrule
\end{tabular}
\end{adjustbox}
\caption{\textbf{English} fine-grained \cjoint results (P1).}
\label{tab:leaderboard_en}
\end{minipage}
\end{table}

\begin{table*}[tp]
\begin{minipage}[t]{0.485\textwidth}
\centering
\definecolor{negshade}{HTML}{2A78D6}\definecolor{posshade}{HTML}{E34948}
\small
\setlength{\tabcolsep}{2pt}
\begin{adjustbox}{max width=\columnwidth}
\begin{tabular}{@{}l *{5}{r} c *{4}{r}@{}}
\toprule
& \multicolumn{5}{c}{\textbf{Macro-F1 by criterion}} && \multicolumn{4}{c}{\textbf{Overall}} \\
\cmidrule(lr){2-6}\cmidrule(lr){8-11}
\textbf{Judge} & \textbf{R} & \textbf{N} & \textbf{T} & \textbf{S} & \textbf{O} && \textbf{F1} & \textbf{Acc} & \textbf{Borda} & \textbf{Valid} \\
\midrule
Gemini-2.5-Pro & \cellcolor[HTML]{65B6C6}58.9 & \cellcolor[HTML]{E9F5C0}12.0 & \cellcolor[HTML]{9ED5B9}\textbf{36.8} & \cellcolor[HTML]{7FC6C0}\textbf{47.4} & \cellcolor[HTML]{73BFC3}52.6 & & \cellcolor[HTML]{8FCFBC}\textbf{41.6} & \textbf{63.7} & \textbf{78.2} & \textbf{100.0} \\
DeepSeek-V3 & \cellcolor[HTML]{76C1C2}51.7 & \cellcolor[HTML]{E1F1B8}15.7 & \cellcolor[HTML]{AADAB6}33.5 & \cellcolor[HTML]{7FC6C0}\textbf{47.4} & \cellcolor[HTML]{6BBAC5}\textbf{56.4} & & \cellcolor[HTML]{93D1BB}\underline{40.9} & \underline{63.3} & \underline{76.4} & \textbf{100.0} \\
GPT-5 & \cellcolor[HTML]{62B4C7}\textbf{63.7} & \cellcolor[HTML]{F6FBD3}4.3 & \cellcolor[HTML]{A6D9B7}\underline{34.2} & \cellcolor[HTML]{85CABE}45.8 & \cellcolor[HTML]{6EBBC4}55.1 & & \cellcolor[HTML]{93D1BB}40.6 & 59.2 & 72.7 & \textbf{100.0} \\
Qwen3-32B & \cellcolor[HTML]{76C1C2}52.0 & \cellcolor[HTML]{E1F1B8}15.9 & \cellcolor[HTML]{ADDCB6}32.7 & \cellcolor[HTML]{8FCFBC}41.8 & \cellcolor[HTML]{6EBBC4}\underline{55.2} & & \cellcolor[HTML]{97D2BA}39.5 & 57.1 & 69.1 & \textbf{100.0} \\
Claude-Sonnet-4 & \cellcolor[HTML]{62B4C7}\underline{62.5} & \cellcolor[HTML]{F8FCD6}3.6 & \cellcolor[HTML]{B9E1B4}28.8 & \cellcolor[HTML]{82C8BF}46.9 & \cellcolor[HTML]{6EBBC4}55.1 & & \cellcolor[HTML]{97D2BA}39.4 & 59.0 & 61.8 & \textbf{100.0} \\
Qwen2.5-7B & \cellcolor[HTML]{B1DDB5}30.9 & \cellcolor[HTML]{ADDCB6}\textbf{32.9} & \cellcolor[HTML]{ADDCB6}32.0 & \cellcolor[HTML]{88CCBE}44.3 & \cellcolor[HTML]{B5DFB4}30.0 & & \cellcolor[HTML]{AADAB6}34.0 & 56.3 & 45.5 & 99.8 \\
\bottomrule
\end{tabular}
\end{adjustbox}
\caption{\textbf{French} fine-grained \cjoint results (P0).}
\label{tab:leaderboard_fra}
\end{minipage}\hfill
\begin{minipage}[t]{0.485\textwidth}
\centering
\definecolor{negshade}{HTML}{2A78D6}\definecolor{posshade}{HTML}{E34948}
\small
\setlength{\tabcolsep}{2pt}
\begin{adjustbox}{max width=\columnwidth}
\begin{tabular}{@{}l *{5}{r} c *{4}{r}@{}}
\toprule
& \multicolumn{5}{c}{\textbf{Macro-F1 by criterion}} && \multicolumn{4}{c}{\textbf{Overall}} \\
\cmidrule(lr){2-6}\cmidrule(lr){8-11}
\textbf{Judge} & \textbf{R} & \textbf{N} & \textbf{T} & \textbf{S} & \textbf{O} && \textbf{F1} & \textbf{Acc} & \textbf{Borda} & \textbf{Valid} \\
\midrule
Gemini-2.5-Pro & \cellcolor[HTML]{62B4C7}\textbf{75.1} & \cellcolor[HTML]{65B6C6}58.3 & \cellcolor[HTML]{62B4C7}\textbf{68.9} & \cellcolor[HTML]{62B4C7}\textbf{65.5} & \cellcolor[HTML]{62B4C7}59.5 & & \cellcolor[HTML]{62B4C7}\textbf{65.5} & \textbf{83.1} & \textbf{87.3} & \textbf{100.0} \\
Qwen3-32B & \cellcolor[HTML]{62B4C7}\underline{72.4} & \cellcolor[HTML]{6BBAC5}56.6 & \cellcolor[HTML]{62B4C7}66.7 & \cellcolor[HTML]{6EBBC4}55.6 & \cellcolor[HTML]{62B4C7}\textbf{72.8} & & \cellcolor[HTML]{62B4C7}\underline{64.8} & 77.1 & 78.2 & 98.5 \\
Claude-Sonnet-4 & \cellcolor[HTML]{62B4C7}67.2 & \cellcolor[HTML]{65B6C6}58.9 & \cellcolor[HTML]{62B4C7}\underline{68.8} & \cellcolor[HTML]{62B4C7}\underline{61.8} & \cellcolor[HTML]{62B4C7}60.7 & & \cellcolor[HTML]{62B4C7}63.5 & \underline{81.3} & \underline{83.6} & 99.9 \\
DeepSeek-V3 & \cellcolor[HTML]{62B4C7}65.2 & \cellcolor[HTML]{62B4C7}\textbf{63.0} & \cellcolor[HTML]{62B4C7}65.0 & \cellcolor[HTML]{76C1C2}51.7 & \cellcolor[HTML]{65B6C6}59.3 & & \cellcolor[HTML]{62B4C7}60.8 & 79.3 & 72.7 & 98.9 \\
GPT-5 & \cellcolor[HTML]{62B4C7}68.4 & \cellcolor[HTML]{65B6C6}58.3 & \cellcolor[HTML]{62B4C7}61.3 & \cellcolor[HTML]{6BBAC5}56.1 & \cellcolor[HTML]{65B6C6}59.3 & & \cellcolor[HTML]{62B4C7}60.7 & 78.8 & 69.1 & 99.7 \\
Qwen2.5-72B & \cellcolor[HTML]{62B4C7}60.0 & \cellcolor[HTML]{62B4C7}\underline{60.9} & \cellcolor[HTML]{62B4C7}63.3 & \cellcolor[HTML]{8FCFBC}41.6 & \cellcolor[HTML]{6BBAC5}56.2 & & \cellcolor[HTML]{6BBAC5}56.4 & 67.0 & 60.0 & 89.9 \\
\bottomrule
\end{tabular}
\end{adjustbox}
\caption{\textbf{Hindi} fine-grained \cjoint results (P0).}
\label{tab:leaderboard_hi}
\end{minipage}
\end{table*}

\begin{table*}[tp]
\begin{minipage}[t]{0.485\textwidth}
\centering
\definecolor{negshade}{HTML}{2A78D6}\definecolor{posshade}{HTML}{E34948}
\small
\setlength{\tabcolsep}{2pt}
\begin{adjustbox}{max width=\columnwidth}
\begin{tabular}{@{}l *{5}{r} c *{4}{r}@{}}
\toprule
& \multicolumn{5}{c}{\textbf{Macro-F1 by criterion}} && \multicolumn{4}{c}{\textbf{Overall}} \\
\cmidrule(lr){2-6}\cmidrule(lr){8-11}
\textbf{Judge} & \textbf{R} & \textbf{N} & \textbf{T} & \textbf{S} & \textbf{O} && \textbf{F1} & \textbf{Acc} & \textbf{Borda} & \textbf{Valid} \\
\midrule
Gemini-2.5-Pro & \cellcolor[HTML]{70BDC3}\textbf{53.6} & \cellcolor[HTML]{93D1BB}\textbf{40.5} & \cellcolor[HTML]{73BFC3}\textbf{52.8} & \cellcolor[HTML]{B5DFB4}30.5 & \cellcolor[HTML]{88CCBE}\underline{44.4} & & \cellcolor[HTML]{88CCBE}\textbf{44.4} & \textbf{65.6} & \textbf{83.6} & \textbf{100.0} \\
Qwen3-32B & \cellcolor[HTML]{82C8BF}46.5 & \cellcolor[HTML]{97D2BA}\underline{39.1} & \cellcolor[HTML]{7CC5C1}49.0 & \cellcolor[HTML]{A6D9B7}34.9 & \cellcolor[HTML]{7FC6C0}\textbf{48.3} & & \cellcolor[HTML]{8CCDBD}\underline{43.6} & \underline{61.8} & 81.8 & 99.8 \\
GPT-5 & \cellcolor[HTML]{7FC6C0}47.7 & \cellcolor[HTML]{97D2BA}\underline{39.1} & \cellcolor[HTML]{79C3C1}\underline{50.1} & \cellcolor[HTML]{A6D9B7}34.3 & \cellcolor[HTML]{8CCDBD}43.6 & & \cellcolor[HTML]{8CCDBD}43.0 & 60.2 & 74.5 & \textbf{100.0} \\
Claude-Sonnet-4 & \cellcolor[HTML]{73BFC3}\underline{52.4} & \cellcolor[HTML]{AADAB6}33.7 & \cellcolor[HTML]{88CCBE}44.9 & \cellcolor[HTML]{9ED5B9}\textbf{36.7} & \cellcolor[HTML]{88CCBE}\underline{44.4} & & \cellcolor[HTML]{8FCFBC}42.4 & 58.9 & \textbf{83.6} & \textbf{100.0} \\
DeepSeek-V3 & \cellcolor[HTML]{88CCBE}44.5 & \cellcolor[HTML]{ADDCB6}32.0 & \cellcolor[HTML]{82C8BF}47.0 & \cellcolor[HTML]{A6D9B7}34.9 & \cellcolor[HTML]{93D1BB}41.0 & & \cellcolor[HTML]{97D2BA}39.9 & 59.2 & 60.0 & 99.5 \\
LLaMA-3.3-70B & \cellcolor[HTML]{7CC5C1}49.6 & \cellcolor[HTML]{B9E1B4}29.0 & \cellcolor[HTML]{8CCDBD}43.0 & \cellcolor[HTML]{A2D7B8}\underline{35.9} & \cellcolor[HTML]{93D1BB}40.7 & & \cellcolor[HTML]{97D2BA}39.7 & 58.0 & 58.2 & \textbf{100.0} \\
\bottomrule
\end{tabular}
\end{adjustbox}
\caption{\textbf{Norwegian} fine-grained \cjoint results (P1).}
\label{tab:leaderboard_nor}
\end{minipage}\hfill
\begin{minipage}[t]{0.485\textwidth}
\centering
\definecolor{negshade}{HTML}{2A78D6}\definecolor{posshade}{HTML}{E34948}
\small
\setlength{\tabcolsep}{2pt}
\begin{adjustbox}{max width=\columnwidth}
\begin{tabular}{@{}l *{5}{r} c *{4}{r}@{}}
\toprule
& \multicolumn{5}{c}{\textbf{Macro-F1 by criterion}} && \multicolumn{4}{c}{\textbf{Overall}} \\
\cmidrule(lr){2-6}\cmidrule(lr){8-11}
\textbf{Judge} & \textbf{R} & \textbf{N} & \textbf{T} & \textbf{S} & \textbf{O} && \textbf{F1} & \textbf{Acc} & \textbf{Borda} & \textbf{Valid} \\
\midrule
Gemini-2.5-Pro & \cellcolor[HTML]{7CC5C1}\underline{49.6} & \cellcolor[HTML]{8FCFBC}\underline{42.1} & \cellcolor[HTML]{73BFC3}\underline{52.8} & \cellcolor[HTML]{7CC5C1}\textbf{49.1} & \cellcolor[HTML]{82C8BF}\underline{47.0} & & \cellcolor[HTML]{7FC6C0}\textbf{48.1} & \textbf{69.1} & \textbf{92.7} & \textbf{100.0} \\
GPT-5 & \cellcolor[HTML]{7FC6C0}47.4 & \cellcolor[HTML]{9BD4BA}38.9 & \cellcolor[HTML]{79C3C1}50.5 & \cellcolor[HTML]{7FC6C0}48.2 & \cellcolor[HTML]{70BDC3}\textbf{53.7} & & \cellcolor[HTML]{7FC6C0}\underline{47.7} & \underline{65.8} & \underline{81.8} & \textbf{100.0} \\
Qwen3-32B & \cellcolor[HTML]{8FCFBC}42.6 & \cellcolor[HTML]{9BD4BA}38.4 & \cellcolor[HTML]{65B6C6}\textbf{58.7} & \cellcolor[HTML]{7FC6C0}\underline{48.5} & \cellcolor[HTML]{85CABE}46.1 & & \cellcolor[HTML]{82C8BF}46.9 & 64.2 & 76.4 & 97.2 \\
DeepSeek-V3 & \cellcolor[HTML]{7FC6C0}47.7 & \cellcolor[HTML]{AADAB6}33.9 & \cellcolor[HTML]{82C8BF}46.9 & \cellcolor[HTML]{7FC6C0}\underline{48.5} & \cellcolor[HTML]{8FCFBC}42.1 & & \cellcolor[HTML]{88CCBE}43.8 & 64.4 & 61.8 & 99.8 \\
LLaMA-3.3-70B & \cellcolor[HTML]{93D1BB}40.5 & \cellcolor[HTML]{A6D9B7}34.6 & \cellcolor[HTML]{7CC5C1}49.2 & \cellcolor[HTML]{82C8BF}47.2 & \cellcolor[HTML]{8FCFBC}41.8 & & \cellcolor[HTML]{8CCDBD}42.7 & 64.2 & 49.1 & \textbf{100.0} \\
Claude-Sonnet-4 & \cellcolor[HTML]{79C3C1}\textbf{50.2} & \cellcolor[HTML]{9BD4BA}38.3 & \cellcolor[HTML]{A2D7B8}36.6 & \cellcolor[HTML]{7FC6C0}47.9 & \cellcolor[HTML]{97D2BA}39.9 & & \cellcolor[HTML]{8FCFBC}42.6 & 61.9 & 60.0 & 96.5 \\
\bottomrule
\end{tabular}
\end{adjustbox}
\caption{\textbf{Spanish} fine-grained \cjoint results (P1).}
\label{tab:leaderboard_spa}
\end{minipage}
\end{table*}

\begin{table*}[tp]
\begin{minipage}[t]{0.485\textwidth}
\centering
\definecolor{negshade}{HTML}{2A78D6}\definecolor{posshade}{HTML}{E34948}
\small
\setlength{\tabcolsep}{2pt}
\begin{adjustbox}{max width=\columnwidth}
\begin{tabular}{@{}l *{5}{r} c *{4}{r}@{}}
\toprule
& \multicolumn{5}{c}{\textbf{Macro-F1 by criterion}} && \multicolumn{4}{c}{\textbf{Overall}} \\
\cmidrule(lr){2-6}\cmidrule(lr){8-11}
\textbf{Judge} & \textbf{R} & \textbf{N} & \textbf{T} & \textbf{S} & \textbf{O} && \textbf{F1} & \textbf{Acc} & \textbf{Borda} & \textbf{Valid} \\
\midrule
Qwen3-32B & \cellcolor[HTML]{62B4C7}\textbf{66.6} & \cellcolor[HTML]{88CCBE}44.2 & \cellcolor[HTML]{62B4C7}\underline{70.0} & \cellcolor[HTML]{88CCBE}\underline{44.3} & \cellcolor[HTML]{7CC5C1}49.2 & & \cellcolor[HTML]{6EBBC4}\textbf{54.8} & \underline{66.2} & \textbf{87.3} & 98.0 \\
Gemini-2.5-Pro & \cellcolor[HTML]{62B4C7}63.4 & \cellcolor[HTML]{88CCBE}\underline{44.6} & \cellcolor[HTML]{62B4C7}69.7 & \cellcolor[HTML]{85CABE}\textbf{45.4} & \cellcolor[HTML]{7FC6C0}47.8 & & \cellcolor[HTML]{70BDC3}\underline{54.2} & \textbf{67.7} & \underline{81.8} & \textbf{100.0} \\
DeepSeek-V3 & \cellcolor[HTML]{62B4C7}60.1 & \cellcolor[HTML]{8CCDBD}43.2 & \cellcolor[HTML]{62B4C7}\textbf{74.6} & \cellcolor[HTML]{A2D7B8}35.9 & \cellcolor[HTML]{7FC6C0}47.5 & & \cellcolor[HTML]{73BFC3}52.3 & 60.7 & 67.3 & 99.0 \\
LLaMA-3.3-70B & \cellcolor[HTML]{62B4C7}62.9 & \cellcolor[HTML]{8FCFBC}42.2 & \cellcolor[HTML]{62B4C7}65.9 & \cellcolor[HTML]{B5DFB4}29.6 & \cellcolor[HTML]{7FC6C0}48.2 & & \cellcolor[HTML]{7CC5C1}49.8 & 57.1 & 60.0 & 99.9 \\
GPT-5 & \cellcolor[HTML]{76C1C2}51.6 & \cellcolor[HTML]{88CCBE}44.4 & \cellcolor[HTML]{65B6C6}58.3 & \cellcolor[HTML]{88CCBE}44.1 & \cellcolor[HTML]{7CC5C1}48.9 & & \cellcolor[HTML]{7CC5C1}49.5 & 64.3 & 65.5 & 99.9 \\
Qwen2.5-72B & \cellcolor[HTML]{62B4C7}\underline{64.4} & \cellcolor[HTML]{7FC6C0}\textbf{48.5} & \cellcolor[HTML]{62B4C7}64.6 & \cellcolor[HTML]{E4F3BA}13.8 & \cellcolor[HTML]{70BDC3}\textbf{54.4} & & \cellcolor[HTML]{7CC5C1}49.1 & 51.3 & 74.5 & 93.3 \\
\bottomrule
\end{tabular}
\end{adjustbox}
\caption{\textbf{Urdu} fine-grained \cjoint results (P1).}
\label{tab:leaderboard_ur}
\end{minipage}\hfill
\begin{minipage}[t]{0.485\textwidth}
\centering
\definecolor{negshade}{HTML}{2A78D6}\definecolor{posshade}{HTML}{E34948}
\small
\setlength{\tabcolsep}{2pt}
\begin{adjustbox}{max width=\columnwidth}
\begin{tabular}{@{}l *{5}{r} c *{4}{r}@{}}
\toprule
& \multicolumn{5}{c}{\textbf{Macro-F1 by criterion}} && \multicolumn{4}{c}{\textbf{Overall}} \\
\cmidrule(lr){2-6}\cmidrule(lr){8-11}
\textbf{Judge} & \textbf{R} & \textbf{N} & \textbf{T} & \textbf{S} & \textbf{O} && \textbf{F1} & \textbf{Acc} & \textbf{Borda} & \textbf{Valid} \\
\midrule
Gemini-2.5-Pro & \cellcolor[HTML]{62B4C7}\textbf{64.6} & \cellcolor[HTML]{85CABE}\underline{45.5} & \cellcolor[HTML]{79C3C1}\underline{51.0} & \cellcolor[HTML]{9ED5B9}\underline{36.7} & \cellcolor[HTML]{79C3C1}\textbf{50.4} & & \cellcolor[HTML]{7CC5C1}\textbf{49.7} & \textbf{73.2} & \textbf{94.5} & \textbf{100.0} \\
GPT-5 & \cellcolor[HTML]{85CABE}45.3 & \cellcolor[HTML]{93D1BB}41.2 & \cellcolor[HTML]{76C1C2}\textbf{51.3} & \cellcolor[HTML]{7CC5C1}\textbf{49.2} & \cellcolor[HTML]{85CABE}45.1 & & \cellcolor[HTML]{82C8BF}\underline{46.4} & \underline{71.7} & \underline{76.4} & \textbf{100.0} \\
Claude-Sonnet-4 & \cellcolor[HTML]{85CABE}45.8 & \cellcolor[HTML]{85CABE}45.2 & \cellcolor[HTML]{8CCDBD}43.4 & \cellcolor[HTML]{D2EBB7}20.7 & \cellcolor[HTML]{85CABE}45.6 & & \cellcolor[HTML]{97D2BA}40.1 & 59.8 & 74.5 & \textbf{100.0} \\
DeepSeek-V3 & \cellcolor[HTML]{93D1BB}40.9 & \cellcolor[HTML]{97D2BA}39.7 & \cellcolor[HTML]{7CC5C1}49.0 & \cellcolor[HTML]{DDF0B8}17.0 & \cellcolor[HTML]{88CCBE}44.9 & & \cellcolor[HTML]{9BD4BA}38.3 & 51.2 & 56.4 & 87.4 \\
GPT-4o-mini & \cellcolor[HTML]{82C8BF}46.6 & \cellcolor[HTML]{82C8BF}\textbf{47.0} & \cellcolor[HTML]{8FCFBC}42.0 & \cellcolor[HTML]{EFF8CA}8.8 & \cellcolor[HTML]{85CABE}\underline{45.8} & & \cellcolor[HTML]{9BD4BA}38.1 & 56.7 & 74.5 & 99.5 \\
LLaMA-3.3-70B & \cellcolor[HTML]{7CC5C1}\underline{48.8} & \cellcolor[HTML]{97D2BA}40.0 & \cellcolor[HTML]{8CCDBD}43.2 & \cellcolor[HTML]{EFF8CA}8.5 & \cellcolor[HTML]{85CABE}45.0 & & \cellcolor[HTML]{9ED5B9}37.1 & 53.2 & 58.2 & 96.7 \\
\bottomrule
\end{tabular}
\end{adjustbox}
\caption{\textbf{Yoruba} fine-grained \cjoint results (P1).}
\label{tab:leaderboard_yo}
\end{minipage}
\end{table*}

\begin{table*}[t]
\centering
\definecolor{negshade}{HTML}{2A78D6}\definecolor{posshade}{HTML}{E34948}
\small
\setlength{\tabcolsep}{4.2pt}
\begin{adjustbox}{max width=\textwidth}
\begin{tabular}{@{}l *{8}{r} c *{8}{r}@{}}
\toprule
& \multicolumn{8}{c}{\textbf{Macro-F1}} && \multicolumn{8}{c}{\textbf{Valid verdicts (\%)}} \\
\cmidrule(lr){2-9}\cmidrule(lr){11-18}
\textbf{Condition} & \textbf{En} & \textbf{Fr} & \textbf{Hi} & \textbf{No} & \textbf{Es} & \textbf{Ur} & \textbf{Yo} & \textbf{All} && \textbf{En} & \textbf{Fr} & \textbf{Hi} & \textbf{No} & \textbf{Es} & \textbf{Ur} & \textbf{Yo} & \textbf{All} \\
\midrule
\cjoint & \cellcolor[HTML]{ADDCB6}32.2 & \cellcolor[HTML]{A2D7B8}35.6 & \cellcolor[HTML]{76C1C2}52.0 & \cellcolor[HTML]{A6D9B7}35.1 & \cellcolor[HTML]{9ED5B9}37.4 & \cellcolor[HTML]{88CCBE}44.3 & \cellcolor[HTML]{B1DDB5}31.4 & \cellcolor[HTML]{9BD4BA}38.3 & & 86.1 & 99.2 & 92.3 & 92.0 & 96.8 & 93.9 & 73.3 & 90.5 \\
\cjointr & \cellcolor[HTML]{A6D9B7}35.0 & \cellcolor[HTML]{AADAB6}33.9 & \cellcolor[HTML]{70BDC3}53.6 & \cellcolor[HTML]{9ED5B9}36.7 & \cellcolor[HTML]{8FCFBC}42.3 & \cellcolor[HTML]{88CCBE}44.6 & \cellcolor[HTML]{B1DDB5}31.3 & \cellcolor[HTML]{97D2BA}39.6 & & 94.9 & 97.4 & 98.5 & 94.0 & 96.8 & 91.9 & 71.7 & 92.2 \\
\ccrit & \cellcolor[HTML]{AADAB6}33.5 & \cellcolor[HTML]{AADAB6}33.4 & \cellcolor[HTML]{76C1C2}51.7 & \cellcolor[HTML]{9ED5B9}36.6 & \cellcolor[HTML]{9BD4BA}38.8 & \cellcolor[HTML]{8FCFBC}42.2 & \cellcolor[HTML]{B5DFB4}30.6 & \cellcolor[HTML]{9BD4BA}38.1 & & 95.5 & 95.1 & 93.8 & 95.6 & 95.9 & 91.9 & 74.1 & 91.7 \\
\ccritr & \cellcolor[HTML]{ADDCB6}32.8 & \cellcolor[HTML]{B5DFB4}29.7 & \cellcolor[HTML]{76C1C2}51.6 & \cellcolor[HTML]{A6D9B7}34.5 & \cellcolor[HTML]{9BD4BA}37.9 & \cellcolor[HTML]{97D2BA}39.7 & \cellcolor[HTML]{B5DFB4}30.3 & \cellcolor[HTML]{9ED5B9}36.6 & & 97.5 & 96.1 & 98.2 & 94.7 & 96.1 & 92.1 & 73.2 & 92.5 \\
\bottomrule
\end{tabular}
\end{adjustbox}
\caption{\textbf{Evaluation conditions by language.} Mean macro-F1 and valid-verdict rate across twelve judges and five criteria. \textsc{All}: unweighted language mean.}
\label{tab:conditions_by_language}
\end{table*}

\section{Detailed Results}
\label{app:results_by_language}

Below, we discuss additional detailed results on preference prediction, evaluation conditions, position and length bias, as well as self-preference.

\subsection{Preference Prediction}
\label{app:lang_prediction}
Tables~\ref{tab:leaderboard_en}-\ref{tab:leaderboard_yo} report \cjoint with the selected prompt for each language. Borda combines the five criterion rankings. \texttt{Gemini-2.5-Pro} leads in every language except Urdu, with mean macro-F1 from 41.6 in French to 65.5 in Hindi; in Urdu, \texttt{Qwen3-32B} (54.8) is just ahead of it (54.2). The second-ranked judge is within one point of the leader in every language except Yoruba, where \texttt{GPT-5} trails by 3.3 points. The breakdown also reveals validity failures: \texttt{Phi-4-mini} has a valid-verdict rate of only 1.9\% in Yoruba and 45.9\% in Norwegian, and \texttt{LLaMA-3.1-8B} of 12.9\% in Yoruba.

Restricting \cjoint to the 19 instructions shared by all seven languages leaves the judge ranking unchanged (Spearman $\rho = 1.00$; every judge's mean macro-F1 changes by at most 0.5 points) and English mean macro-F1 nearly unchanged (31.8 vs.\ 32.2), so the lower English scores do not stem from its additional instructions. On the 62 comparisons shared by the non-English languages, Hindi (52.7) and Urdu (46.1) remain the highest and Norwegian (35.9) and Yoruba (34.2) the lowest.

\subsection{Evaluation Conditions}
\label{app:lang_conditions}
\begin{figure*}[p]
\centering
\includegraphics[width=\textwidth,trim=6 6 6 6,clip]{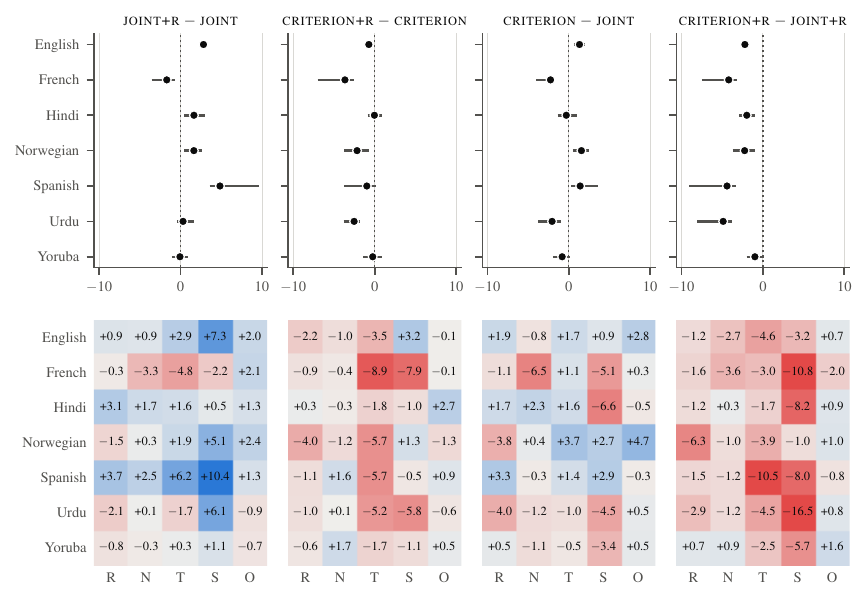}
\caption{\textbf{Condition effects by language.} Change in macro-F1 (in points) between the two conditions in each panel title (e.g., \cjointr $-$ \cjoint). Top: mean change per language across judges and criteria, with 95\% bootstrap intervals. Bottom: mean change per language and criterion across judges (R, N, T, S, and O as in \autoref{tab:leaderboard}).}
\label{fig:setup_contrasts_by_language}
\end{figure*}

\begin{figure*}[p]
\centering
\includegraphics[width=\textwidth,trim=6 6 6 6,clip]{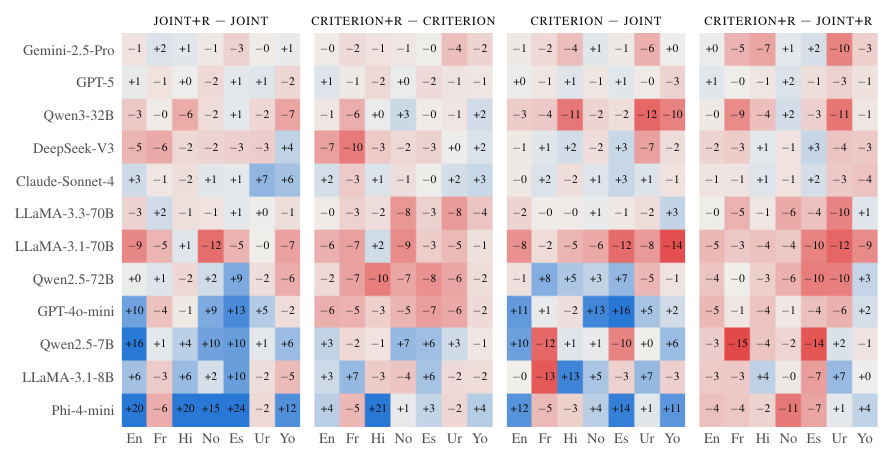}
\caption{\textbf{Condition effects by judge and language.} Change in macro-F1 (in points) per judge and language, averaged over the five criteria, between the two conditions in each panel title (e.g., \cjointr $-$ \cjoint). Judges are ordered by \cjoint macro-F1.}
\label{fig:setup_contrasts_judge_language}
\end{figure*}

\autoref{tab:conditions_by_language} reports macro-F1 and valid-verdict rates across conditions. Figures~\ref{fig:setup_contrasts_by_language} and~\ref{fig:setup_contrasts_judge_language} break down the contrasts in \autoref{fig:setup_contrasts}, and \autoref{tab:explanation_robustness} in \S\ref{app:metrics} repeats them for subsets of judges (see \S\ref{subsec:res_conditions}). Averaged over languages, \cjointr yields the highest mean macro-F1 (39.6) and \ccritr the lowest (36.6). Hindi has the highest scores (51.5-53.6) and Yoruba the lowest (30.3-31.4), with valid-verdict rates of only 71-74\%. The decline from separating criteria when rationales are requested (\ccritr $-$ \cjointr) appears in every language, from $-1.0$ in Yoruba to $-4.9$ in Urdu, and its 95\% interval excludes 0 in all languages except Yoruba (see \autoref{fig:setup_contrasts_by_language}). Requesting rationales in joint calls helps most in Spanish ($+4.8$) and English ($+2.8$), but hurts French ($-1.7$). In \autoref{tab:explanation_robustness}, the direction of each contrast is similar; for the reliable judges, however, the gain from rationales in joint calls disappears ($+0.0$).

\subsection{Position and Length Bias}
\label{app:lang_length}
\begin{table}[ht]
\centering
\begin{minipage}{\columnwidth}
\centering
\definecolor{negshade}{HTML}{2A78D6}\definecolor{posshade}{HTML}{E34948}
\small
\setlength{\tabcolsep}{3pt}
\begin{adjustbox}{max width=\columnwidth}
\begin{tabular}{@{}l r c r r r@{}}
\toprule
& \multicolumn{2}{c}{\textbf{Position score}} & \textbf{1st-slot} & \textbf{2nd-slot} & \textbf{Consis-} \\
\cmidrule(lr){2-3}
& \textbf{Mean} & \textbf{95\% CI} & \textbf{lock} & \textbf{lock} & \textbf{tency} \\
\midrule
\multicolumn{6}{@{}l}{By criterion} \\
Relevance & \cellcolor{posshade!18}+0.20 & [+0.19, +0.21] & 17.9 & 2.9 & 61.9 \\
Naturalness & \cellcolor{posshade!5}+0.06 & [+0.05, +0.08] & 13.2 & 8.6 & 65.5 \\
Truthfulness & \cellcolor{posshade!13}+0.14 & [+0.13, +0.16] & 13.1 & 2.4 & 61.9 \\
Safety & \cellcolor{posshade!15}+0.17 & [+0.16, +0.18] & 14.3 & 1.9 & 64.2 \\
Overall quality & \cellcolor{posshade!11}+0.12 & [+0.11, +0.14] & 16.4 & 5.4 & 70.3 \\
\midrule
\multicolumn{6}{@{}l}{By language} \\
English & \cellcolor{posshade!19}+0.21 & [+0.20, +0.23] & 18.9 & 2.2 & 60.3 \\
Urdu & \cellcolor{posshade!10}+0.11 & [+0.10, +0.13] & 11.5 & 2.2 & 73.4 \\
Norwegian & \cellcolor{posshade!18}+0.20 & [+0.17, +0.22] & 16.9 & 2.9 & 59.1 \\
Hindi & \cellcolor{posshade!6}+0.07 & [+0.05, +0.09] & 12.3 & 5.8 & 71.1 \\
Yoruba & \cellcolor{posshade!7}+0.08 & [+0.04, +0.13] & 13.0 & 7.4 & 65.6 \\
Spanish & \cellcolor{posshade!18}+0.20 & [+0.17, +0.23] & 17.4 & 2.9 & 59.4 \\
French & \cellcolor{posshade!9}+0.09 & [+0.08, +0.12] & 14.4 & 6.1 & 64.6 \\
\midrule
\textbf{Overall} & \cellcolor{posshade!13}+0.14 & -- & 15.0 & 4.2 & 64.8 \\
\bottomrule
\end{tabular}
\end{adjustbox}
\caption{\textbf{Position bias by criterion and language} averaged over judges, using comparisons with a valid verdict in both response orders. 95\% CI are over instructions within each language.}
\label{tab:position_bias_groups}
\end{minipage}
\end{table}

\autoref{tab:position_bias_groups} reports position bias by criterion and language, and Tables \ref{tab:length_bias} and \ref{tab:length_bias_by_language} report length bias.

\paragraph{Position bias:} Judges favor the first slot for every criterion and language, with all 95\% intervals above zero (see \autoref{tab:position_bias_groups}). First-slot locks (11.5-18.9\%) are far more common than second-slot locks (1.9-8.6\%). Consistency is highest in Urdu (73.4\%) and Hindi (71.1\%) and lowest in Norwegian (59.1\%) and Spanish (59.4\%).

\paragraph{Length bias:} \texttt{GPT-4o-mini} ($+11.3$) and \texttt{LLaMA-3.3-70B} ($+9.9$) favor longer responses most strongly relative to humans, whereas \texttt{Qwen2.5-7B} and \texttt{Phi-4-mini} favor them slightly less than humans ($-1.9$ and $-2.0$; see \autoref{tab:length_bias}). When annotators rate a comparison as a tie but the judge does not, the judge picks the longer response in 68.7\% of cases, from 53.6\% in Yoruba to 79.4\% in Spanish. Humans prefer longer responses most often in Hindi, e.g.,~85.6\% for relevance (see \autoref{tab:length_bias_by_language}). Overall, excluding the three low-validity judges raises the mean bias from $+4.7$ to $+6.5$ points.

\begin{table*}[p]
\centering
\small
\setlength{\tabcolsep}{3pt}
\definecolor{negshade}{HTML}{2A78D6}\definecolor{posshade}{HTML}{E34948}
\begin{adjustbox}{max width=\textwidth}
\begin{tabular}{@{}l *{5}{r} @{\hspace{10pt}} *{7}{r} @{\hspace{10pt}} r@{}}
\toprule
 & \multicolumn{5}{c}{\textbf{By criterion}} & \multicolumn{7}{c}{\textbf{By language}} & \\
\cmidrule(r{6pt}){2-6} \cmidrule(r{6pt}){7-13}
\textbf{Judge} & \textbf{R} & \textbf{N} & \textbf{T} & \textbf{S} & \textbf{O} & \textbf{En} & \textbf{Fr} & \textbf{Hi} & \textbf{No} & \textbf{Es} & \textbf{Ur} & \textbf{Yo} & \textbf{Mean} \\
\midrule
GPT-4o-mini & \cellcolor{posshade!28}+9.5 & \cellcolor{posshade!44}+14.6 & \cellcolor{posshade!37}+12.3 & \cellcolor{posshade!30}+10.0 & \cellcolor{posshade!31}+10.2 & \cellcolor{posshade!43}+14.5 & \cellcolor{posshade!27}+9.0 & \cellcolor{posshade!21}+7.0 & \cellcolor{posshade!45}+15.6 & \cellcolor{posshade!45}+16.7 & \cellcolor{posshade!39}+13.1 & \cellcolor{posshade!10}+3.4 & \cellcolor{posshade!34}+11.3 \\
LLaMA-3.3-70B & \cellcolor{posshade!13}+4.3 & \cellcolor{posshade!45}+15.0 & \cellcolor{posshade!29}+9.6 & \cellcolor{posshade!37}+12.4 & \cellcolor{posshade!25}+8.4 & \cellcolor{posshade!32}+10.8 & \cellcolor{posshade!17}+5.5 & \cellcolor{posshade!10}+3.5 & \cellcolor{posshade!45}+14.9 & \cellcolor{posshade!40}+13.2 & \cellcolor{posshade!43}+14.2 & \cellcolor{posshade!23}+7.6 & \cellcolor{posshade!30}+9.9 \\
Qwen2.5-72B & \cellcolor{posshade!7}+2.4 & \cellcolor{posshade!31}+10.3 & \cellcolor{posshade!33}+11.1 & \cellcolor{posshade!45}+15.6 & \cellcolor{posshade!15}+5.1 & \cellcolor{posshade!43}+14.2 & \cellcolor{posshade!31}+10.5 & \cellcolor{posshade!11}+3.8 & \cellcolor{posshade!28}+9.2 & \cellcolor{posshade!37}+12.3 & \cellcolor{posshade!40}+13.2 & \cellcolor{negshade!2}$-$0.8 & \cellcolor{posshade!27}+8.9 \\
DeepSeek-V3 & \cellcolor{posshade!14}+4.7 & \cellcolor{posshade!35}+11.5 & \cellcolor{posshade!21}+7.0 & \cellcolor{posshade!13}+4.4 & \cellcolor{posshade!19}+6.5 & \cellcolor{posshade!33}+11.1 & \cellcolor{posshade!19}+6.3 & \cellcolor{posshade!2}+0.6 & \cellcolor{posshade!30}+9.9 & \cellcolor{posshade!26}+8.6 & \cellcolor{posshade!34}+11.2 & +0.1 & \cellcolor{posshade!20}+6.8 \\
Claude-Sonnet-4 & \cellcolor{posshade!16}+5.2 & \cellcolor{posshade!22}+7.2 & \cellcolor{posshade!28}+9.4 & \cellcolor{posshade!17}+5.6 & \cellcolor{posshade!16}+5.4 & \cellcolor{posshade!28}+9.4 & \cellcolor{posshade!12}+4.1 & \cellcolor{posshade!7}+2.4 & \cellcolor{posshade!27}+8.9 & \cellcolor{posshade!33}+11.0 & \cellcolor{posshade!38}+12.7 & \cellcolor{negshade!8}$-$2.7 & \cellcolor{posshade!20}+6.6 \\
LLaMA-3.1-70B & \cellcolor{posshade!1}+0.2 & \cellcolor{posshade!29}+9.7 & \cellcolor{posshade!23}+7.7 & \cellcolor{posshade!30}+10.0 & \cellcolor{posshade!10}+3.4 & \cellcolor{posshade!31}+10.4 & \cellcolor{posshade!24}+8.1 & \cellcolor{negshade!29}$-$9.8 & \cellcolor{posshade!30}+10.2 & \cellcolor{posshade!30}+10.0 & \cellcolor{posshade!28}+9.2 & \cellcolor{posshade!16}+5.5 & \cellcolor{posshade!19}+6.2 \\
GPT-5 & \cellcolor{posshade!16}+5.3 & \cellcolor{posshade!15}+5.1 & \cellcolor{posshade!11}+3.7 & \cellcolor{posshade!6}+2.0 & \cellcolor{posshade!8}+2.5 & \cellcolor{negshade!13}$-$4.4 & \cellcolor{posshade!6}+2.1 & +0.1 & \cellcolor{posshade!26}+8.8 & \cellcolor{posshade!37}+12.4 & \cellcolor{posshade!28}+9.3 & \cellcolor{negshade!6}$-$2.1 & \cellcolor{posshade!11}+3.7 \\
Gemini-2.5-Pro & $-$0.1 & \cellcolor{posshade!21}+7.1 & \cellcolor{posshade!8}+2.6 & \cellcolor{posshade!4}+1.4 & \cellcolor{posshade!10}+3.2 & \cellcolor{posshade!14}+4.8 & \cellcolor{posshade!5}+1.5 & \cellcolor{negshade!7}$-$2.4 & \cellcolor{posshade!15}+5.0 & \cellcolor{posshade!25}+8.3 & \cellcolor{posshade!23}+7.8 & \cellcolor{negshade!16}$-$5.3 & \cellcolor{posshade!8}+2.8 \\
Qwen3-32B & $-$0.1 & \cellcolor{posshade!1}+0.2 & \cellcolor{posshade!10}+3.5 & \cellcolor{posshade!10}+3.5 & \cellcolor{posshade!11}+3.7 & \cellcolor{posshade!10}+3.3 & \cellcolor{negshade!1}$-$0.5 & \cellcolor{negshade!6}$-$1.9 & \cellcolor{posshade!15}+4.9 & \cellcolor{posshade!15}+4.9 & \cellcolor{posshade!17}+5.7 & \cellcolor{negshade!4}$-$1.4 & \cellcolor{posshade!7}+2.2 \\
LLaMA-3.1-8B & \cellcolor{negshade!2}$-$0.6 & \cellcolor{negshade!1}$-$0.5 & \cellcolor{posshade!19}+6.4 & \cellcolor{posshade!10}+3.2 & \cellcolor{posshade!2}+0.7 & \cellcolor{posshade!31}+10.3 & \cellcolor{negshade!8}$-$2.7 & \cellcolor{negshade!44}$-$14.7 & \cellcolor{posshade!21}+6.9 & \cellcolor{posshade!15}+5.0 & \cellcolor{posshade!5}+1.8 & \cellcolor{posshade!19}+6.2 & \cellcolor{posshade!6}+1.8 \\
Qwen2.5-7B & \cellcolor{negshade!22}$-$7.4 & \cellcolor{posshade!2}+0.6 & \cellcolor{negshade!1}$-$0.3 & \cellcolor{posshade!10}+3.5 & \cellcolor{negshade!18}$-$5.9 & \cellcolor{posshade!15}+5.0 & \cellcolor{negshade!11}$-$3.6 & \cellcolor{negshade!34}$-$11.2 & \cellcolor{posshade!11}+3.5 & \cellcolor{posshade!11}+3.8 & \cellcolor{negshade!3}$-$1.2 & \cellcolor{negshade!29}$-$9.7 & \cellcolor{negshade!6}$-$1.9 \\
Phi-4-mini & \cellcolor{negshade!21}$-$7.1 & \cellcolor{negshade!8}$-$2.6 & \cellcolor{posshade!6}+1.8 & \cellcolor{posshade!9}+3.1 & \cellcolor{negshade!16}$-$5.3 & \cellcolor{posshade!2}+0.7 & \cellcolor{negshade!9}$-$3.1 & \cellcolor{negshade!34}$-$11.5 & \cellcolor{posshade!27}+9.1 & \cellcolor{posshade!2}+0.5 & \cellcolor{posshade!2}+0.5 & \cellcolor{negshade!34}$-$11.5 & \cellcolor{negshade!6}$-$2.0 \\
\midrule
Judges: longer (\%) & 66.8 & 65.9 & 63.4 & 57.2 & 71.0 & 61.4 & 63.6 & 74.2 & 61.6 & 62.5 & 72.1 & 58.8 & 64.9 \\
Humans: longer (\%) & 65.4 & 59.4 & 57.2 & 51.0 & 67.8 & 53.9 & 60.5 & 77.0 & 52.7 & 53.6 & 63.9 & 59.5 & 60.2 \\
Judges $>$ humans & 7 & 10 & 11 & 12 & 10 & 11 & 8 & 6 & 12 & 12 & 11 & 5 & 10 \\
Longer when humans tie (\%) & 69.9 & 65.9 & 69.4 & 68.5 & 69.5 & 66.7 & 66.8 & 65.3 & 69.8 & 79.4 & 74.9 & 53.6 & 68.7 \\
\bottomrule
\end{tabular}
\end{adjustbox}
\caption{\textbf{Length bias}: judge-human difference (points) in preference for the longer response, on comparisons with valid verdicts in both orders (judge scores average the two orders). Criterion columns average over languages, language columns over criteria. Bottom rows: mean shares across judges, number of judges with a positive difference, and share favoring the longer response when humans label Tie but the judge does not.}
\label{tab:length_bias}
\end{table*}

\begin{table*}[p]
\centering
\small
\setlength{\tabcolsep}{4pt}
\definecolor{negshade}{HTML}{2A78D6}\definecolor{posshade}{HTML}{E34948}
\begin{tabularx}{\textwidth}{@{}l *{5}{>{\raggedleft\arraybackslash}X} @{\hspace{10pt}} *{5}{>{\raggedleft\arraybackslash}X} @{\hspace{10pt}} >{\raggedleft\arraybackslash}X>{\raggedleft\arraybackslash}X@{}}
\toprule
 & \multicolumn{5}{c}{\textbf{Humans: longer (\%)}} & \multicolumn{5}{c}{\textbf{Length bias (points)}} & \multicolumn{2}{c}{\textbf{Mean bias}} \\
\cmidrule(r{6pt}){2-6} \cmidrule(r{6pt}){7-11} \cmidrule{12-13}
\textbf{Language} & \textbf{R} & \textbf{N} & \textbf{T} & \textbf{S} & \textbf{O} & \textbf{R} & \textbf{N} & \textbf{T} & \textbf{S} & \textbf{O} & \textbf{All} & \textbf{Nine} \\
\midrule
English & 58.6 & 49.9 & 50.1 & 49.7 & 60.9 & \cellcolor{posshade!14}+4.8 & \cellcolor{posshade!43}+14.4 & \cellcolor{posshade!24}+7.9 & \cellcolor{posshade!11}+3.6 & \cellcolor{posshade!20}+6.7 & \cellcolor{posshade!23}+7.5 & \cellcolor{posshade!25}+8.2 \\
French & 78.8 & 48.6 & 47.7 & 49.5 & 78.0 & \cellcolor{negshade!31}$-$10.5 & \cellcolor{posshade!38}+12.6 & \cellcolor{posshade!44}+14.7 & \cellcolor{posshade!17}+5.6 & \cellcolor{negshade!21}$-$6.9 & \cellcolor{posshade!9}+3.1 & \cellcolor{posshade!16}+5.2 \\
Hindi & 85.6 & 84.2 & 72.3 & 56.3 & 86.7 & \cellcolor{negshade!25}$-$8.2 & \cellcolor{negshade!25}$-$8.3 & \cellcolor{posshade!6}+2.1 & \cellcolor{posshade!20}+6.7 & \cellcolor{negshade!19}$-$6.4 & \cellcolor{negshade!9}$-$2.8 & \cellcolor{posshade!1}+0.4 \\
Norwegian & 55.9 & 48.4 & 49.7 & 51.3 & 58.3 & \cellcolor{posshade!21}+7.0 & \cellcolor{posshade!45}+15.0 & \cellcolor{posshade!30}+10.1 & \cellcolor{posshade!9}+2.9 & \cellcolor{posshade!29}+9.5 & \cellcolor{posshade!27}+8.9 & \cellcolor{posshade!29}+9.7 \\
Spanish & 52.9 & 56.7 & 50.6 & 49.3 & 58.4 & \cellcolor{posshade!29}+9.8 & \cellcolor{posshade!26}+8.6 & \cellcolor{posshade!27}+8.9 & \cellcolor{posshade!11}+3.8 & \cellcolor{posshade!40}+13.4 & \cellcolor{posshade!27}+8.9 & \cellcolor{posshade!33}+10.8 \\
Urdu & 68.3 & 64.4 & 68.3 & 50.1 & 68.5 & \cellcolor{posshade!16}+5.4 & \cellcolor{posshade!28}+9.5 & \cellcolor{posshade!9}+2.9 & \cellcolor{posshade!45}+15.5 & \cellcolor{posshade!22}+7.3 & \cellcolor{posshade!24}+8.1 & \cellcolor{posshade!32}+10.7 \\
Yoruba & 57.6 & 63.6 & 61.6 & 50.9 & 63.9 & \cellcolor{posshade!4}+1.2 & \cellcolor{negshade!19}$-$6.2 & \cellcolor{negshade!9}$-$2.9 & \cellcolor{posshade!16}+5.4 & \cellcolor{negshade!3}$-$1.1 & \cellcolor{negshade!2}$-$0.7 & \cellcolor{posshade!1}+0.5 \\
\midrule
\textit{Mean} & 65.4 & 59.4 & 57.2 & 51.0 & 67.8 & \cellcolor{posshade!4}+1.4 & \cellcolor{posshade!20}+6.5 & \cellcolor{posshade!19}+6.2 & \cellcolor{posshade!19}+6.2 & \cellcolor{posshade!10}+3.2 & \cellcolor{posshade!14}+4.7 & \cellcolor{posshade!19}+6.5 \\
\bottomrule
\end{tabularx}
\caption{\textbf{Length bias by language and criterion}, as in \autoref{tab:length_bias}. \textsc{Humans}: tie-adjusted share of comparisons in which annotators prefer the longer response. \textsc{All}: mean over criteria for all twelve judges; \textsc{Nine}: without the low-validity judges.}
\label{tab:length_bias_by_language}
\end{table*}

\subsection{Self-Preference}
\label{app:self_preference}
\autoref{tab:self_preference} reports self-preference for every judge-generator pair from the same model or model family (see \S\ref{subsec:res_bias}). \texttt{GPT-4o-mini} and \texttt{LLaMA-3.3-70B} also judge responses of their own model. The control judges are \texttt{DeepSeek-V3}, \texttt{Gemini-2.5-Pro}, \texttt{Qwen2.5-72B}, \texttt{Qwen2.5-7B}, \texttt{Qwen3-32B}, and \texttt{Phi-4-mini}. Comparisons are pooled over languages, and judge verdicts come from \cjoint and \cpos with the selected prompts. Positive values for most pairs and criteria indicate the bias.

\begin{table*}[p]
\centering
\definecolor{negshade}{HTML}{2A78D6}\definecolor{posshade}{HTML}{E34948}
\small
\setlength{\tabcolsep}{4pt}
\begin{tabular*}{\textwidth}{@{\extracolsep{\fill}}ll *{5}{r} r r@{}}
\toprule
& & \multicolumn{5}{c}{\textbf{By criterion}} & \multicolumn{2}{c}{\textbf{All criteria}} \\
\cmidrule(lr){3-7}\cmidrule(lr){8-9}
\textbf{Judge} & \textbf{Generator} & \textbf{R} & \textbf{N} & \textbf{T} & \textbf{S} & \textbf{O} & \textbf{Mean} & \textbf{95\% CI} \\
\midrule
GPT-5 & GPT-4o-mini & \cellcolor{posshade!37}+12.4 & \cellcolor{posshade!24}+8.0 & \cellcolor{posshade!15}+5.1 & \cellcolor{negshade!19}$-$6.2 & \cellcolor{posshade!21}+6.9 & \cellcolor{posshade!16}+5.3 & [+3.9, +6.7] \\
 & GPT-4-0613$^{\mathrm{en}}$ & \cellcolor{posshade!18}+6.1 & \cellcolor{posshade!44}+14.8 & \cellcolor{posshade!45}+16.2 & \cellcolor{posshade!17}+5.5 & \cellcolor{posshade!37}+12.5 & \cellcolor{posshade!33}+11.0 & [+8.8, +13.4] \\
\addlinespace[2pt]
GPT-4o-mini & GPT-4o-mini$^{*}$ & \cellcolor{posshade!32}+10.7 & \cellcolor{posshade!28}+9.2 & \cellcolor{posshade!31}+10.5 & \cellcolor{posshade!33}+11.1 & \cellcolor{posshade!24}+8.1 & \cellcolor{posshade!30}+10.0 & [+8.4, +11.5] \\
 & GPT-4-0613$^{\mathrm{en}}$ & \cellcolor{negshade!34}$-$11.5 & \cellcolor{negshade!29}$-$9.5 & \cellcolor{negshade!16}$-$5.3 & \cellcolor{posshade!5}+1.8 & \cellcolor{negshade!32}$-$10.6 & \cellcolor{negshade!21}$-$7.0 & [$-$8.7, $-$5.2] \\
\addlinespace[2pt]
Claude-Sonnet-4 & Claude-3.7-Sonnet & \cellcolor{posshade!11}+3.7 & \cellcolor{posshade!35}+11.8 & \cellcolor{posshade!23}+7.6 & \cellcolor{posshade!28}+9.2 & \cellcolor{posshade!17}+5.8 & \cellcolor{posshade!23}+7.6 & [+5.3, +10.1] \\
 & Claude-3-Opus$^{\mathrm{en}}$ & \cellcolor{posshade!10}+3.5 & \cellcolor{posshade!33}+10.9 & \cellcolor{posshade!21}+6.9 & \cellcolor{negshade!6}$-$2.1 & \cellcolor{posshade!37}+12.3 & \cellcolor{posshade!19}+6.3 & [+5.0, +7.7] \\
\addlinespace[2pt]
LLaMA-3.3-70B & LLaMA-3.3-70B$^{*}$ & \cellcolor{posshade!1}+0.5 & \cellcolor{negshade!18}$-$6.1 & \cellcolor{negshade!14}$-$4.6 & \cellcolor{negshade!17}$-$5.6 & \cellcolor{negshade!1}$-$0.4 & \cellcolor{negshade!10}$-$3.2 & [$-$5.6, $-$1.1] \\
 & LLaMA-3-70B$^{\mathrm{en}}$ & \cellcolor{posshade!15}+5.0 & \cellcolor{posshade!38}+12.7 & \cellcolor{posshade!3}+0.8 & \cellcolor{negshade!11}$-$3.7 & \cellcolor{posshade!36}+11.9 & \cellcolor{posshade!16}+5.3 & [+3.7, +7.0] \\
\addlinespace[2pt]
LLaMA-3.1-70B & LLaMA-3.3-70B & \cellcolor{posshade!4}+1.2 & \cellcolor{negshade!11}$-$3.7 & \cellcolor{negshade!5}$-$1.5 & \cellcolor{negshade!11}$-$3.6 & \cellcolor{posshade!3}+1.1 & \cellcolor{negshade!4}$-$1.3 & [$-$4.1, +1.2] \\
 & LLaMA-3-70B$^{\mathrm{en}}$ & \cellcolor{posshade!14}+4.5 & \cellcolor{posshade!31}+10.2 & \cellcolor{posshade!15}+5.1 & \cellcolor{posshade!7}+2.2 & \cellcolor{posshade!27}+9.0 & \cellcolor{posshade!19}+6.2 & [+4.5, +8.0] \\
\addlinespace[2pt]
LLaMA-3.1-8B & LLaMA-3.3-70B & \cellcolor{posshade!18}+6.0 & \cellcolor{posshade!10}+3.4 & \cellcolor{posshade!16}+5.2 & \cellcolor{posshade!6}+2.0 & \cellcolor{posshade!30}+10.0 & \cellcolor{posshade!16}+5.3 & [+3.0, +8.0] \\
 & LLaMA-3-70B$^{\mathrm{en}}$ & \cellcolor{posshade!20}+6.6 & \cellcolor{posshade!6}+2.0 & \cellcolor{posshade!19}+6.4 & \cellcolor{posshade!29}+9.7 & \cellcolor{posshade!5}+1.7 & \cellcolor{posshade!16}+5.2 & [+2.6, +7.8] \\
\bottomrule
\end{tabular*}
\caption{\textbf{Self-preference.} Each row pairs a judge with a related generator. Cells give the judge's excess preference for that generator over the human labels, minus the same excess for the six control judges (95\% CI are over instructions). Judge scores average both response orders; positive values mean the judge favors the related generator more than humans and control judges do. $^{*}$Same model; $^{\mathrm{en}}$English generator (all others: the six other languages).}
\label{tab:self_preference}
\end{table*}

\clearpage
\twocolumn[{%
\section{Human Evaluation of NLI Verdict Recovery}
\label{app:human_evaluation}
\setlength{\multicolsep}{0pt}
\begin{multicols}{2}
Here, we present a shortened version of the guidelines for human evaluation of NLI verdict recovery.

We further show a snapshot of the interface we used (see \S\ref{app:rationale}).
\end{multicols}

\subsection{Guidelines}
\begin{center}
\begin{minipage}{\textwidth}
\fontsize{10}{11}\selectfont
\setlength{\parindent}{0pt}
\setlist[itemize]{
    leftmargin=1.5em,
    itemsep=2pt,
    topsep=3pt,
    parsep=0pt
}
\setlist[enumerate]{
    leftmargin=1.7em,
    itemsep=3pt,
    topsep=3pt,
    parsep=0pt
}

\smallskip
\hrule height 0.6pt
\medskip

\noindent
\begin{minipage}[t]{0.48\linewidth}
\setlength{\parskip}{5pt}
\textbf{Objective}

Read a short rationale that compares two AI-generated responses to the same instruction, Response A and Response B, on one of the five criteria. You see only the rationale, not the instruction or the responses. Identify the preference the rationale expresses; do not decide whether its assessment is correct.

\textbf{Choosing an answer}

\begin{itemize}
    \item \textbf{A:} The rationale prefers A, explicitly or through its reasoning, such as praising A or identifying a problem with B.
    \item \textbf{B:} The rationale prefers B in the same way.
    \item \textbf{Tie:} The rationale expresses equal quality or no meaningful difference; the responses can be equally good or equally bad.
    \item \textbf{``Can't tell'':} No preference or equality can be inferred: there is no comparison, the trade-offs are unresolved, or the text is too unclear or broken to interpret.
\end{itemize}

\textbf{Self-contradiction flag}

Tick the flag only when the rationale states a conclusion and its reasons clearly favor a different answer on the same criterion. Still select the stated conclusion as your answer. Leave the flag unchecked when the reasons are weak or incomplete, the preference is small, or the rationale weighs different strengths before reaching a conclusion.
\end{minipage}
\hfill
\begin{minipage}[t]{0.48\linewidth}
\setlength{\parskip}{5pt}
\textbf{General rules}

\begin{enumerate}
    \item \textbf{A stated conclusion takes priority.}
    If the rationale says ``Final verdict: B'', select B.

    \item \textbf{Without a stated conclusion, use the reasoning.}
    ``A contains a factual error; B has none'' implies B.

    \item \textbf{Small preferences count.}
    ``Both are fine, but B is slightly clearer'' means B, not Tie.

    \item \textbf{Different strengths do not mean Tie.}
    ``A is more detailed; B is more concise'' gives no preference or equality; choose ``Can't tell''.

    \item \textbf{Follow the rationale.}
    Do not replace its preference with your own, and do not fact-check its claims.
\end{enumerate}

\textbf{General steps}

\begin{enumerate}
    \item \textbf{Train.}
    Complete seven English training checks and compare each response with the suggested answer.

    \item \textbf{Annotate.}
    Work independently through two files of 50 rationales each.

    \item \textbf{Add a note.}
    Report issues such as cut-off text or mixed languages.

    \item \textbf{Skip sensitive content.}
    Choose ``Can't tell'' and add a note if you prefer not to read a rationale.
\end{enumerate}
\end{minipage}

\medskip
\hrule height 0.6pt

\end{minipage}
\end{center}

\subsection{Interface}
\begin{center}
\setfigurecaptype
\begin{minipage}{\textwidth}
\small
\setlength{\parindent}{0pt}
\setlength{\fboxsep}{6pt}
\setlength{\fboxrule}{0.4pt}

\smallskip
\hrule height 0.6pt
\medskip

\textbf{Which response is favored?}
\par\smallskip
\colorbox{black!6}{%
    \parbox[t]{\dimexpr\linewidth-2\fboxsep\relax}{%
        Response A is more factually accurate. Response B contains several factual errors that are absent from A. Final verdict: B.
    }%
}
\par\medskip

\radiobutton~A \qquad
\radiobutton*~\textbf{B} \qquad
\radiobutton~Tie \qquad
\radiobutton~Can't tell
\par\medskip

\noindent\fbox{%
    \parbox[t]{\dimexpr\linewidth-2\fboxsep-2\fboxrule\relax}{%
        {\setlength{\fboxsep}{1pt}\fbox{\scriptsize\ding{51}}}\ \textbf{The reasoning contradicts the stated conclusion}
        \par\smallskip
        Select the stated conclusion above; tick this box only if the reasons clearly favor a different answer.
    }%
}
\par\medskip

\noindent\fbox{%
    \parbox[t]{\dimexpr\linewidth-2\fboxsep-2\fboxrule\relax}{%
        \textit{Add a note (optional).}
        \par\vspace{5mm}
    }%
}
\par\medskip

\fbox{Back}\hfill\fbox{Next}
\par\medskip
\hrule height 0.6pt

\end{minipage}
\caption{\textbf{Schematic of the human evaluation interface.} Annotators see only the rationale, choose A, B, Tie, or ``Can't tell'' for the preference it expresses, can flag reasoning that contradicts the stated conclusion, add an optional note, and go back to revise earlier answers. The example is an English training check: the reasons favor A, but the stated conclusion is B, so B is selected and the flag is checked.}
\label{fig:human_evaluation_interface}
\end{center}
}]

\end{document}